\documentclass{report}

\usepackage{etoolbox}
\makeatletter
\newif\if@abstractmode

\renewenvironment{titlepage}
{
  \if@twocolumn
  \@restonecoltrue\onecolumn
  \else
  \@restonecolfalse\newpage
  \fi
  \if@abstractmode
  \thispagestyle{plain}%
  \stepcounter{page}%
  \else
  \thispagestyle{empty}%
  \setcounter{page}\@ne%
  \fi
}%
{\if@restonecol\twocolumn \else \newpage \fi
  \if@twoside\else
  \if@abstractmode
  \else
  \setcounter{page}\@ne%
  \fi
  \fi
}
\AtBeginEnvironment{abstract}{%
  \@abstractmodetrue%
}
\makeatother

\usepackage{geometry}

\usepackage{algorithm}
\usepackage{algpseudocode}
\usepackage{amsfonts}
\usepackage{amsmath}
\usepackage{amssymb}
\usepackage{bm}
\usepackage{booktabs}
\usepackage{caption}
\usepackage{CJKutf8}
\usepackage{csquotes}
\usepackage{comment}
\usepackage{graphicx} %
\usepackage{hyperref}
\usepackage{import}
\usepackage{latexsym}
\usepackage{lineno}
\usepackage{longtable}
\usepackage{makecell}
\usepackage{mathtools}
\usepackage{mdframed}
\usepackage{multirow}
\usepackage{natbib}
\usepackage{subcaption}
\usepackage{times}
\usepackage{titlesec}
\usepackage{usebib}
\usepackage{xcolor}

\newcommand{\davis}[1]{}%

\newcommand{\prob}{\mathcal{P}}
\newcommand{\seqs}{\mathcal{S}}
\newcommand{\dperfect}{\mathcal{D}_{\text{clean}}}
\newcommand{\dclean}{\mathcal{D}_{\text{clean}}}
\newcommand{\sperfect}{\mathcal{S}_{\text{clean}}}
\newcommand{\sclean}{\mathcal{S}_{\text{clean}}}
\newcommand{\stypical}{\seqs_{\text{typical}}}

\newcommand{\dtrain}{\mathcal{D}_{\text{train}}}
\newcommand{\ptrain}{\prob_{\text{train}}}
\newcommand{\dmodel}{\mathcal{D}_{\text{model}}}
\newcommand{\pmodel}{\mathcal{P}_{\text{model}}}

\newcommand{\pclf}{\mathcal{P}_{\text{clf}}}

\DeclareMathOperator*{\argmax}{argmax}

\newcommand\given[1][]{\:#1\vert\:}

\title{Making the Most of your Model: Methods for Finetuning and Applying Pretrained Transformers}
\author{Davis Yoshida}
\date{August 2023}

\begin{document}
\pagenumbering{roman}
\begin{titlepage}
    \begin{center}
        \vspace*{1cm}
        \Huge \textbf{Making the Most of your Model:\\Methods for Finetuning and Applying Pretrained Transformers}

        \vspace{1.5cm}

        \Large
        \vspace{0.5cm}

        \Large \textbf{Davis Yoshida}

        \vspace{1.5cm}

        \large
        A dissertation submitted in partial fulfillment of the\\
        requirements for the degree of\\
        Doctor of Philosophy in Computer Science\\
        at the Toyota Technological Institute at Chicago\\
        2023

        \vfill

        \large
        Thesis Committee:\\
        \vspace{0.3cm}
        Kevin Gimpel (Thesis advisor)\\
        Karen Livescu\\
        Hyokun Yun\\
        David McAllester
        \vspace{1cm}
    \end{center}
\end{titlepage}
\clearpage
\begin{abstract}
Recent progress in natural language processing (NLP) has been dominated by pretrained transformer language models (LMs).
While the rate of improvement has been astounding, we are far from knowing how to optimally use even the models we already have.
Improvements to our knowledge about how to make use of these models, because they increase the utility of all extant pretrained transformer models, even those trained for niche domains which are not well covered by even state of the art large language models (LLMs).

This thesis provides methods and analysis of models which make progress on this goal.
The techniques outlined are task agnostic, and should provide benefit when used with nearly any transformer LM.
We introduce two new finetuning methods which add new capabilities to the models they are used on.
The first adds a recurrence mechanism, which removes the fixed-window sized constraint and improves the efficiency of a transformer decoder.
The second allows masked language models (MLMs) to be used for initialization of both the encoder and decoder of a non-autoregressive sequence-to-sequence transformer, opening up generative applications of models which were previously only used for natural language understanding tasks.

We also introduce two new techniques for improving the quality of predictions of any transformer decoder without additional finetuning.
One, hidden state optimization, can be applied to any transformer decoder to improve the quality of predictions at inference time, especially for few-shot classification.
The other, conditional beam search, allows practitioners to search for natural language generation (NLG) model outputs with high likelihood while conditioning on the event that the output is not degenerate (e.g. empty, repetitive, etc.).

Finally, we provide theoretical and empirical insights on the divergence of model-likelihood and output quality which has widely been observed in prior work.
These insights apply to any model which represents a distribution over text, and apply to language models which are not transformers or even autoregressive.
We argue that the NLP community has, to some extent, misunderstood the implications of these findings, and encourage a point of view which has more nuance.

Taken together, the findings in this thesis should allow NLP practitioners to make much more effective use of pretrained models, either those that already exist or ones that will be created in the future.
\end{abstract}

\tableofcontents
\begin{CJK*}{UTF8}{gbsn}
\listoffigures
\listoftables
\end{CJK*}
\cleardoublepage

\chapter*{Acknowledgements}
Thank you to everyone who taught, mentored, and supported me throughout my PhD journey. Especially thank you to Kevin Gimpel for patiently advising me over my time at TTIC.

Thank you also to my friends and family, without whom I definitely would not have been able to complete this journey. Most importantly, thank you to Meg Vader, my girlfriend, then fiancée, then wife. Your unwavering support gives me a foundation I wouldn't know what to do without.

\clearpage

\pagenumbering{arabic}
\chapter{Introduction}
In recent years, natural language processing (NLP) practice has steadily become more and more based on the application of increasingly large pretrained language models (LMs).
By one estimate\footnote{https://venturebeat.com/ai/ai-machine-learning-openai-gpt-3-size-isnt-everything/}, training GPT-3~\citep{brown2020language} took \$12 million, and the scale of training runs has only continued to increase.
During this process, models have improved and grown at a blindingly fast pace, with each generation of models replacing the last.

Since we keep moving on to newer models so quickly, we're nowhere close to knowing how to optimally \emph{apply} the models we already have.
Many NLP models have been trained in specific domains such as scientific or medical text, code, or low-resource languages.
Improvements in modeling or the scale of training don't provide benefit to practitioners using these models, since the massive scale used for the latest models necessarily only produces generalist models.
However, new or improved ways of \emph{using existing models} can benefit all the users of the thousands of LMs that have been trained over the last several years.

While the pace of progress has been blindingly fast, NLP has been surprisingly stable in its usage of the transformer architecture~\citep{vaswani2017attention}.
As such, methods for making more effective use of transformers won't just help the previous generation of models, but can be applied to the latest (and hopefully future) generations as well.

This thesis discusses methods that open up new ways of using language models, especially those based on transformers~\citep{vaswani2017attention}.
The first two methods are new ways of finetuning pretrained transformers.
In Chapter~\ref{chapter:recurrence} we outline a method for adding recurrence to pretrained transformers at finetuning time.
This method leads to reduced execution cost for a desired performance level compared to the unmodified transformer it is applied to.
In Chapter~\ref{chapter:mt} we propose a method for converting a pretrained encoder-only transformer into a non-autoregressive (non-AR) text generator.
We demonstrate that it is an effective means of bringing pretrained models to bear on non-autoregressive natural language and code translation.
Similarly to the other work discussed here, the method makes few assumptions about the transformer being used, and could be deployed for any sequence-to-sequence generative task.

The remainder of the work focuses on inference time usage of generative models of text.
In Chapter~\ref{chapter:hso}, I give a gradient-based method for improving the inference time performance of autoregressive generation from any transformer, testing it on language modeling and few-shot classification.
Chapters~\ref{chapter:modes_theory} and \ref{chapter:modes_exact} focus on improving our understanding of natural language generation (NLG) by theoretical reframing of the ``bad mode'' problem, and empirical investigation of the properties of exact unconditional and conditional modes of NLG models.
Chapter~\ref{chapter:modes_beam} takes advantage of these insights to develop \emph{conditional beam search}, which finds outputs which are high likelihood while still being high-quality.

To provide context for this work, the next section will give a brief summary of the recent history of pretrained LMs, as well as discussing some of the methods which are essential for the use of NLG systems today.
Section~\ref{sec:contributions} then outlines the specific contributions of this thesis, and the benefits they can have for any NLP practitioner who wishes to make use of pretrained models.

\section{LMs and NLG: Background and related work}\label{section:related_work}
This section will give an overview of work in several areas which are relevant to this thesis.
Each chapter also has more specific discussion of work which is directly applicable to that that chapter.
The broadest two topics which the reader should be familiar with to read this thesis are language models (LMs) and specifically natural language generation (NLG).
Section~\ref{sec:relatedwork_llm} gives an extremely coarse overview of developments in language modeling over the last several years.
The remainder of this chapter looks at work from several related research areas, including various nonstandard methods for finetuning LMs, and decoding methods for NLG.

\subsection{Language modeling, transformers, and pretraining}\label{sec:relatedwork_llm}
In the last 5 years, there's been a shift from NLP models being trained from scratch, to almost all models being initialized from another model which was pretrained in some way.
These pretrained models are created with the goal of being general-purpose, as opposed to being trained for a single task.
This change has been a significant departure from the more traditional machine learning setting of training and testing on IID data for a single task.
We now broadly refer to these models as \emph{large language models} (LLMs), although that term wasn't in wide use several years ago.

In this section we'll give a brief overview of that transition, in order to give context for the work in this thesis.
This isn't meant to be a full history, but rather just to identify some key results which occurred along the way.
We'll refer to the task used to initially train the model as the \emph{pretraining task}, and the tasks on which it's later used as \emph{downstream tasks}.
Application of models to downstream tasks may or may not involve further training, depending on the model and task.
The key feature is that a single pretraining procedure is now used to produce a model which is used on \emph{many} downstream tasks.

\subsubsection{Three families of pretrained models}
There are essentially three kinds of pretrained LMs dominating the landscape of pretrained models today.
They are: decoder-only LMs, encoder-only masked language models (MLMs), and sequence-to-sequence (seq2seq) models.
All of these consist of different subsets of the transformer architecture introduced by \citet{vaswani2017attention}, although changes have been made over time.
Of the above, only the decoder models are ``true'' LMs, in the sense of estimating a distribution over text.

Arguably, the transition to using pretrained models started with the Generative Pretrained Transformer (GPT, which is now referred to as GPT-1, \citealp{radford2018improving}), which is a decoder-only LM.
\citet{radford2018improving}'s method had two parts.
The first was to train a transformer decoder on the pretraining task of language modeling.
Then, to apply the resulting LM to a downstream task, the final output linear layer is replaced with a task specific one.\footnote{This is necessary since for language modeling the output linear layer has an output dimension equal to the model's vocabulary size, but for a classification task the output dimension should be equal to the number of classes}
The model is then trained on the downstream task, with all of the model weights being updated.
For example, for recognizing textual entailment\footnote{This is a three-way classification task, where given a premise and a hypothesis, the model must predict whether the premise entails the hypothesis, contradicts is, or has a neural relationship to it.}, the premise and hypothesis would be concatenated into a single text, fed to the transformer, then the transformer's final layer hidden state would be passed into a linear layer with output dimension 3.

GPT-1 was just a few months after ELMo~\citep{elmo} was published, so when we say that GPT-1 is the start of the trend, we're not saying that they came up with all the ideas out of nowhere.
The reason to single out GPT-1 is because of three ways it differs from ELMo.
\begin{enumerate}
    \item ELMo was used by leaving the ELMo model itself static, and training a full model on top of it, whereas the entire GPT-1 model was intended to be finetuned.
    \item ELMo is a biLSTM model, and GPT-1 is a transformer.
    \item ELMo embedded words using a character RNN, but GPT-1 used a fixed subword tokenization called byte-pair encoding (BPE, \citealp{bpe}).
\end{enumerate}
The first point, that application to downstream tasks essentially only requires finetuning the pretrained model, not adding a large number of new parameters, was the key change.
ELMo (and the earlier CoVe~\citep{cove}) can be seen as the culmination of the earlier trend in which models were generally trained \emph{mostly} from scratch, but would use pretrained \emph{embeddings} such as GloVe~\citep{glove}.
The latter two points were not novel, but those particular choices turned out to be prescient (or influential), since essentially all later models have made the same choices.
Subword tokenization, in particular, allows the model to take any text as input without the need for replacing some tokens with an ``\texttt{<unk>}'' (unknown) token.

The next major landmark was BERT~\citep{bert}, which combined the ``finetune the entire model'' and transformer aspects of GPT-1 with the bidirectionality of ELMo.
One of their important findings was that ``deep bidirectionality'' was important, meaning that simply combining left-to-right and right-to-left LMs was less powerful than a transformer that could ``see'' the entire sentence at once while processing any token.
The method that allowed this to happen is called masked language modeling, so BERT is a \emph{masked language model}.
MLM training consists of taking an input text, then replacing a subset of the tokens\footnote{As with GPT-1, the tokens come from a fixed vocabulary of about 50,000 subwords.} with a special \texttt{[MASK]} token.
Then, for each training example, the model must predict what the original values of the masked tokens were.
This allowed BERT to use a transformer \emph{encoder} architecture instead of a transformer \emph{decoder} architecture.
Changing to an encoder-only architecture meant that the transformer's representation of every token depended on \emph{all} tokens, not just those preceding it.

The switch from language model training to masked language modeling means that an MLM is not a true language model, because it does not represent a distribution over text.
As such, these models are generally used for natural language understanding (NLU) tasks rather than NLG tasks.
However, both pretraining tasks are similar in that they use the entire input text itself as the target that must be predicted.
Intuitively, this should lead to models that extract as much usable information as possible from the input text.\footnote{Something like a binary classification task, on the other hand, would encourage the model to throw away all information that wasn't relevant to the task.}

The third major paradigm started with T5~\citep{t5}, which tries to get the best of both worlds by using a seq2seq architecture (i.e., an entire transformer as defined by \citealp{vaswani2017attention}).
The pretraining task was \emph{masked span prediction}, which is similar to MLM training, but leads to models more suited for text generation.
For each training example, several contiguous spans of tokens are dropped, and replaced with a single mask token.
The encoder processes the corrupted text, and then the decoder must predict the tokens which were removed in each span.
This is done using the same cross-entropy loss used for language modeling and machine translation.
The benefit of T5-style models is that the encoder can take advantage of bidirectional attention which \citet{bert} identified as important, while remaining useful for NLG, thanks to their transformer decoder component.

Many more models were trained during this time period, but GPT-1, BERT, and T5 were the models that launched their respective paradigms.
The differences between these three models can be summarized in terms of both the pretraining task and the architecture:
\begin{itemize}
    \item Architecture: GPT-1 is a decoder-only model, BERT is an encoder-only model, and T5 is an encoder-decoder model.\footnote{When we use the phrase ``encoder-only'' model in this thesis, we will mean a transformer that does not use causal attention masking. That is, the outputs for a token at position 3 can depend not only on positions 1 and 2, but also 4, 5, etc.}
    \item Pretraining: Decoder-only models are usually trained as LMs, encoder-only models are usually trained as MLMs, and T5-like models are usually trained to repair span corruption
\end{itemize}
Of course later models don't fall neatly into this categorization, as researchers have been trying to improve every aspect.
Despite that, we'd argue that this is presently still a good mental framework for classifying the current generation of models.
Specifically, decoder-only models dominate current usage for text generation, while MLMs are widely used for classification or embedding of text.
Encoder-decoder models are most likely to be used as initialization for conditional generation, although they can also be used for the same things as encoder-only or decoder-only models.
This classification is highly oversimplified.
For example, see \citet{ul2} for an attempt to train a single model type which is effective for all downstream tasks.

The large scale pretrained models\footnote{We do make use of a pretrained seq2seq model for machine translation, but it was trained on that task specifically as opposed to being a general purpose pretrained model..} in this work fall into the decoder-only and encoder-only categories.
Chapters~\ref{chapter:recurrence}, \ref{chapter:hso}, \ref{chapter:modes_exact}, and \ref{chapter:modes_beam} make use of decoder-only LMs, specifically GPT-2~\citep{gpt2} and LLaMA~\citep{llama}.
Chapter~\ref{chapter:mt} discusses a new method for the application of (encoder-only) MLMs to text generation, specifically non-autoregressive generation.

\subsubsection{Coherent text generation}
The rest of this section will focus exclusively on the LM/decoder-only family of models.
About one year after GPT-1's release, \citet{gpt2} introduced GPT-2.
GPT-2 was essentially the same architecture as GPT-1, but scaled up to 1.5 billion parameters (GPT-1 had around 100M).
GPT-2 demonstrated coherent long form text generation for the first time.
While previous language models could only generate small amounts of text, GPT-2 was able to generate entire news articles that were at least superficially high-quality.

Following this, there was a large amount of interest in how to achieve better text generation quality from models like GPT-2.
A language model represents a distribution over strings of tokens.
Specifically nearly all language models represent that distribution in an \emph{autoregressive} form, as $\prob (x_t | x_{<t})$ for a candidate token $x_t$ and the preceding tokens $x_{<t}$.
These conditional probabilities can be combined into a single probability $\prob(x_{1:T})$ for a string of $T$ tokens.
The ``default'' way of producing text from such a model is to sample $x_1$, $x_2$ and so on from the model sequentially until the model outputs an ``\texttt{<|endoftext|>}'' token, indicating that the output is complete.
However, drawing samples directly from the distribution in this way leads to very low quality samples. (This method is called \emph{ancestral sampling}).
\citet{gpt2} instead limited this sampling to only the 40 most likely tokens.\footnote{For each token to be generated, the model still predicts the probability for every token in vocabulary, but only the 40 tokens it ranks highest can be output.}

One follow-up work that should be mentioned is \citet{nucleus_sampling}, which found that text generated from these models was very different from human text in terms of the likelihood of the produced tokens.
They proposed \emph{nucleus sampling} (also called top-$p$ sampling), in which only the top $p$ fraction of the model's token distribution is used for generation, preventing generation of extremely unlikely tokens.
It can be seen as a more adaptive version of the top-40 approach used by \citet{gpt2} and is currently still in wide use.
For further discussion of generation methods, see Section~\ref{sec:relatedwork_generation}, and Chapters~\ref{chapter:modes_theory} and \ref{chapter:modes_beam}.

\subsubsection{Few and Zero-shot learning/Prompting}
Another substantial recent development is the ability to use pretrained LMs without any finetuning of the model weights.
Instead, the downstream task is expressed in some way via text, and the model generates a response (also in textual form).
This method was introduced by \citet{brown2020language} with their GPT-3 model.
Their instantiation of it was to encode in text a series of input/output pairs for the target, then finally an input to which the model should give a response.
For example, to use the model for sentiment classification, one might pass as input (as plain text):
\begin{displayquote}
Classify the following texts as Positive or Negative - ``I love it'': Positive, ``I hate it'': Negative, ``I thought it was neat'':
\end{displayquote}
GPT-3 (and later models) will output ``Positive'' consistently enough to make this a viable classification method.

This is referred to as few-shot learning, since the model must infer the desired behavior from a small number of examples.
If only a textual description is given and no examples, it is instead called zero-shot learning.
The method of eliciting a desired behavior out of a language model using purely textual input is now generally referred to as ``prompting'', with the input/instructions being referred to as the ``prompt''.

While prompting is extremely ad hoc compared to the finetuning method that started with GPT-1, training LLMs is extremely costly in terms of resources.
Many of the recently released open source models have tens of billions of parameters, meaning that even finetuning them on a small amount of data is out of reach using consumer hardware.
While a user might get better results using finetuning in principle, in practice it may be too expensive, and prompting turns out to work well enough in some cases.
However, there is also an ongoing research area that aims to reduce the cost of finetuning (See Section~\ref{sec:relatedwork_efficient} and Chapter~\ref{chapter:recurrence}).
In Chapter~\ref{chapter:hso}, we discuss the application of our \emph{Hidden-State Optimization} (HSO) technique to prompt-based few-shot classification, finding that it can improve performance in the setting we tested.

\subsubsection{The current generation of LMs: Training for instruction following}
Anything written here risks being badly out of date within a year or two, but we'll now give a snapshot of the current phase of the development of LLMs.
The current focus is on LLMs which aren't really trained as LMs, because they are not trying to model any particular text distribution.
Instead, models such as GPT-4~\citep{gpt4} are now trained using reinforcement learning from human feedback (RLHF, \citealp{rlhf}).
This constitutes an additional phase in training: First a model is pretrained on language modeling, then the model is finetuned using RLHF, and then the model is used for downstream tasks.

While GPT-4 was trained with RLHF, RLHF is extremely costly which makes it so that individual practitioners (or smaller groups) can't take advantage of it.
Instead, many open source models just take a pretrained language model\footnote{Currently the most popular is LLama2~\citep{llama2}}, then finetune it with something similar to instruction finetuning~\citep{flan}. (See for example Alpaca~\citealp{alpaca}).
Some other models trained with RLHF include LLaMA2-chat~\citep{llama2} and StableVicuna\footnote{\url{https://stability.ai/blog/stablevicuna-open-source-rlhf-chatbot}}.
In this thesis, we do not use any RLHF trained models for our experiments, although we discuss RLHF briefly in Chapter~\ref{chapter:modes_theory}.

Similar to how MLMs are still called language models despite not really representing a distribution over text, RLHF-trained models are still referred to as LMs despite their training focusing on maximizing reward rather than fitting a training data distribution.
However, unlike MLMs, the trained model still does \emph{represent} a distribution over text.
This distribution is used via variants of ancestral sampling such as nucleus sampling or greedy decoding (i.e., simply generating text by selecting the token the model thinks is most likely at each step).

This is where we are today: the most impressive models are all trained with RLHF, and there is a great deal of interest in how to improve their performance for NLG.
\citet{rlhf_problems} provides a good overview of the current thinking about RLHF and the challenges it faces.

\subsection{Improving the efficiency of transformers}\label{sec:relatedwork_efficient}
Now that we've established the overall context this thesis exists in, we can look at relevant related work in some specific subareas.
This section will look specifically at methods that attempt to improve the efficiency of transformers, as opposed to improving their raw power.

Due to the high cost of training and deploying transformer models, there has been significant interest in how they can be made more efficient.
The majority of these proposals involve training from scratch using architectures that are more efficient, a topic which is surveyed in \citet{tay2022efficient}.

\subsubsection{Efficient finetuning}
A particularly important line of work is methods for reducing the cost of finetuning transformers.
Examples of this include training adapter modules~\citep{adapters}, prefix tuning~\citep{li2021prefix}, or low-rank adaptation (LoRA \citealp{lora}).
These methods are generally described as ``parameter-efficient finetuning'' because they reduce the number of parameters to be trained.
While none of these reduce the cost of storing the model weights or activations, they still reduce the amount of GPU memory needed for training.
This is because a substantial amount of memory is needed to store optimizer \emph{moment estimates} for the nearly universally used Adam~\citep{adam} optimizer.
Since this is so costly, methods which reduce the number of parameters to be trained can still lead to much cheaper training, even if they don't reduce the cost of executing the model.
We take advantage of this fact in Chapter~\ref{chapter:hso}, by optimizing the values of the transformer activations themselves, rather than the model weights.

\subsubsection{Reducing memory requirements of pretrained models}
While the previous section discussed methods which reduce the memory required to finetune an LM, there are also several important methods for reducing the cost of using a transformer LM overall without making any modifications to the model's architecture.
Most of these methods only make inference time usage of LMs cheaper, but some reduce the resources needed for both training and test time.

\emph{Quantization} (See e.g., ~\citealp{gptq,llm_int8,qlora}) reduces the memory needed to store the model weights themselves by representing them in a smaller width floating point number, with the current most common size being 4 bits.
However, this is typically done \emph{after} training, as the minimum step size for these narrow floats is far too large.
\citet{qlora} showed that combining 4-bit quantization with LoRA~\citep{lora} (referring to this combination as QLoRA) could be effective.
This means the pretrained model's weights being stored in 4-bit precision, and only a small number of newly added weights need to be stored in full precision.
This method is in extremely widespread use today, as it enables partial finetuning of models with several billion parameters on a single GPU.

Another important technique is FlashAttention~\citep{flashattention}, which is simply a more efficient implementation of the standard transformer self-attention introduced by \citet{vaswani2017attention}.
This was followed by FlashAttention2~\citep{flashattention2}, which makes further improvements.
Previously, it was widely assumed that computing self-attention required quadratic memory, but this turns out to not actually be the case.

Another option is \emph{knowledge distillation} (see for example \citealp{distilbert}), in which a small model is trained to emulate a larger model.
This takes advantage of the fact that one of the main benefits of training large models seems to be (paradoxically) making training cheaper.
A small model can often represent the same function learned by a larger model, but finding the setting of the weights which does so is more expensive for the small model.
For discussion on this point, see \citet{scaling_laws,chinchilla,llama}.

These methods can all be combined with the methods discussed in this thesis, although the tradeoff between computational cost and performance is an open question for future work.
For example, the finetuning methods in Chapter~\ref{chapter:recurrence} and Chapter~\ref{chapter:hso} could be implemented using LoRA or even QLoRA, dramatically reducing the resources needed.
Similarly, none of our methods were implemented using FlashAttention, but they could be applied using fewer resources via the application of FlashAttention.

\subsection{Modifying and controlling generation}\label{sec:relatedwork_generation}
Chapter~\ref{chapter:modes_beam} discusses a particular type of generation, in which we are interested in generating from a model in a way which respects its implied conditional distribution with respect to some attribute.
Chapter~\ref{chapter:hso} also modifies the generative story of an autoregressive LM by updating the hidden states from tokens which have already been emitted.
Those chapters discuss other methods with similar goals, but there are many other methods of modifying generation from LMs.

\subsubsection{Decoding strategies}
Language models define distributions over text, but there are many different ways to go from that distribution to a specific output.
The process of doing so is called \emph{decoding}, and is usually done conditional on some input.
That input is embedded by a transformer encoder for seq2seq models, or is simply included as a prefix to the output for decoder-only models.

As we mentioned in Section~\ref{sec:relatedwork_llm}, the most obvious method (for autoregressive LMs) is ancestral sampling, which yields a sample from the learned distribution.
However, these samples are generally low-quality~\citep{nucleus_sampling,typical_sampling,wiher2022ondecoding}, so modified strategies are generally used.
Several simple such methods are top-$k$ sampling (only the $k$ highest scoring tokens are considered at each time step), greedy-decoding (equivalent to top-1 sampling), and temperature sampling.
Standard autoregressive LMs output unnormalized log-probabilities (logits) of each possible token at each timestep, and temperature decoding simply divides each logit by a scalar temperature value.
That is, for logits $\ell_i$, the probability of emitting token $i$ in a vocabulary $V$ is:
\begin{equation*}
   p_i = \frac{\mathrm{exp}(\ell_i / T)}{\sum\limits_{i\in V} \mathrm{exp}(\ell_i / T)}
\end{equation*}
In the $T=0$ limit, this is equivalent to greedy-decoding, while in the $T=\infty$ limit tokens are emitted from the uniform distribution.
Typically temperatures lower than 1 are used, to reduce the entropy of the model's token distribution.

\citet{typical_sampling} proposed a more complex sampling method, \emph{locally} typical sampling, in which tokens are only considered for generation if their log-probability is close to the average log-probability of the token distribution (i.e. the negative entropy).
They show that this leads to higher quality outputs than many other decoding methods.

The above methods are all \emph{sampling} methods, i.e. they produce outputs from some distribution.
Sampling methods are generally preferred for tasks where more open-ended output is required.
Another class of decoding methods are \emph{search-based} methods, which are often deterministic, and generally try to maximize some quantity.

The most common search-based method is \emph{beam search}~\citep{beamsearch}, which is a heuristic search method which tries to find an output which the NLG assigns a high probability to.
Beam search is rarely used with language models, but is frequently used in MT.
Other search methods which are less commonly used involve approximate minimum Bayes risk (MBR) inference ~\citep{inadequacyofmode}, and Monte-Carlo tree search (MCTS, \citealp{beyondbeamsearch}).
Approximate MBR tries to find an output which is high-quality with respect to some automatically computed metric.
MCTS may be used either to find high-likelihood sequences, or to try to maximize an external score as MBR does.

Another alternative is to abandon left-to-right generation altogether, which leads to \emph{non-autoregressive} (non-AR) generation. We discuss non-AR generation in more depth in Chapter~\ref{chapter:mt}.

\subsubsection{Specializing models via finetuning}
The simplest method of modifying a language model so that its output will display some desired behavior is to finetune it on some data which exemplifies that behavior.
The most prominent example of this today is reinforcement learning from human feedback (RLHF, \citealp{rlhf}), which was discussed briefly in Section~\ref{sec:relatedwork_llm}.

In RLHF, samples are drawn from the model, which are then judged by human raters with respect to how well the outputs satisfy certain desiderata.
These judgements are then used to train a \emph{reward model}, which is then used to apply proximal policy optimization~\citep{ppo} to the model.
The goal of this is to find a model which maximizes the expected reward with respect to the reward model which was trained on the human judgements.
Unlike simple language model training, this no longer has the goal of training the model to match a certain distribution.
RLHF has been used for specific NLG tasks such as summarization~\citep{learningtosummarize}, but is now used for training models to follow instructions in a very general setting~\citep{instructgpt}.

A simpler method is \emph{instruction finetuning}, which simply consists of ordinary LM training on a dataset of instruction/response pairs (see e.g. \citealp{flan,flanpalm}).
This is often used as an initial step prior to RLHF, to get the model to produce outputs which are at least of decent quality, as the LM needs to be at least producing some good outputs before human judgements can help it improve.

\subsection{Controllable generation}
Other methods try to give more finegrained control over the generation process.
In Chapter~\ref{chapter:modes_beam} we will discuss several which are the most relevant to this thesis in more detail, but we'll give a brief overview here.
The goal of controllable generation is to produce outputs which satisfy some contraint.
Methods may guarantee that this constraint is satisfied, or merely increase the probability a model output satsifies it.

For example, the CTRL LM~\citep{ctrl} was finetuned to respond to a large number of \emph{control codes}, which are conditioning information used to control style and content.
Another method, Plug-and-play Language Models (PPLMs, \citealp{Dathathri2020Plug}), trains a classifier to map from a language model's hidden states to the attribute which is to be controlled (e.g. the topic).
For each token, the classifier is applied, then the gradient of its prediction is used to make the language model's hidden states more likely (in the classifier's opinion) to produce an output which matches the constraint.

Recently, \citet{cfglm} generalized the idea of classifier-free guidance (CFG, \citealp{classifierfree}) to generation from LLMs.
Their method computes the difference between unconditional LM predictions and those conditional on a prompt, then uses that to induce the output to be even more related to the prompt.
CFG was proposed very recently, so while we do not evaluate its interactions with the methods in this thesis, we believe that they should roughly be complementary, and hope to verify that in future work.

\section{Contributions}\label{sec:contributions}
This section will outline and provide context for the techniques for modifying and applying pretrained language models that this thesis introduces.

\subsubsection{Adding Recurrence to Pretrained Transformers}
Due to the quadratic cost of dense self-attention, and the increasing size of modern transformers, finding improvements to the cost of attention is a very active research area.
There have been a large number of proposals for methods which aim to make self-attention more efficient. (For a survey of methods, see~\citet{tay2022efficient}).
Most architectural changes, however, require training a model from scratch which is extremely costly in the case of large pretrained models.

In this work we present a method for applying pretrained transformer language models which lowers the necessary memory footprint requirement both at training and inference time.
The method consists of applying the transformer block-wise to adjacent or overlapping windows of texts, and training a small new recurrence module which allows information to flow in between these blocks.
This method was devised prior to FlashAttention~\citep{flashattention}, which reduces the memory footprint of computing attention.
However, attention is still fundamentally quadratic in terms of computational cost, so factoring the attention as we do should still lead to inference time speedups.
We leave the combination of this method and efficient GPU attention kernels to future work.

An additional benefit is that our method removes the fixed context size constraint that most transformer models have, allowing for more flexible use. 
This was true at the time it was initially devised, and is still true for the current generation of LLMs such as LLaMA~\citep{llama}.
When applied to the GPT-2 language model, we find that our method attains better perplexity than an unmodified GPT-2 model on the PG-19 and WikiText-103 corpora, for a given amount of computation or memory.
This work is covered in detail in Chapter~\ref{chapter:mt}.

\subsubsection{Reconsidering the Past: Optimizing Hidden States in Language Models}
A standard autoregressive transformer decoder uses causal masking, meaning that hidden states at a given position only attend to that position or those which precede it, not future positions. This is necessary for language model training, but \citet{bert} showed that bidirectionally encoded representations lead to higher performance on downstream tasks than the unidirectional representations present in causally masked transformers.

As a toy example that demonstrates why bidirectionality is important, consider the famous garden path sentence: ``The old man the boat.''
While a language model is processing this sentence, it must produce an embedding of ``old'', using only ``The old.''
The most reasonable inference from this limited information is that ``old'' is an adjective, while in the full sentence it is clear that it is actually a noun.
An autoregressive LM will therefore likely have a representation which is in some way ``incorrect'' compared to an embedding produced by an encoder-only model (i.e., one without an attention mask) such as BERT~\citep{bert}.
While this example is extreme, in general the representation of a token should be higher quality if more information is used to compute it.

Some methods do allow for text generation conditioned on a bidirectionally encoded prefix or source sentence. Encoder-decoder models such as those in the T5 model family~\citep{t5} use bidirectional attention in the encoder, but still use unidirectional attention in the decoder.
Another is the PrefixLM task used as part of the training of UL2~\citep{ul2}, in which a transformer decoder has a bidirectional attention mask when embedding a prefix, but still uses a causal mask for decoding.
These methods both suffer from the fact that if one wants to include newly generated tokens in the bidirectional attention flow, the original forward pass must be run again.
They also are specifications of architectures, and so if one wants to apply a standard decoder-only transformer language model, they are not applicable.

Chapter~\ref{chapter:hso} presents Hidden-State Optimization (HSO), a gradient-based method for improving the performance of transformer language models at inference time by adding bidirectional information flow.
Similar to dynamic evaluation \citep{krause2018dynamic}, HSO computes the gradient of the log-probability the language model assigns to an evaluation text, but uses it to update the cached hidden states rather than the model parameters.
This allows models that were trained using purely right-to-left attention to still incorporate information from future tokens into their embeddings of past tokens.
We test HSO with pretrained Transformer-XL and GPT-2 language models, finding improvement on the WikiText-103 and PG-19 datasets in terms of perplexity, especially when evaluating a model outside of its training distribution. We also demonstrate downstream applicability by showing gains in the prompt-based few-shot evaluation setting, again with no extra parameters or training data.
More details are discussed in Chapter~\ref{chapter:hso}.

\subsubsection{Converting Masked Language Models into Non-Autoregressive Encoder-Decoders}
The dominant paradigm in machine translation is to use a transformer encoder-decoder model (with an autoregressive decoder), as introduced by \citet{vaswani2017attention}.
While these models achieve very high performance, autoregressive decoding leads to high latency as a forward pass is required for every single token.
Due to this, there has been a surge of interest in \emph{non-autoregressive} Machine Translation (non-AR MT) (see \citet{nonar_survey} for a survey of methods).
Non-autoregressive decoding methods can offer lower decoding latency, although often at the cost of some performance relative to autoregressive methods.
Unfortunately, most multilingual pretrained models either have auto-regressive decoders, such as in mBART~\citep{mbart}, or consist only of a standalone encoder, such as in XLM-R \citep{xlmr}.

In Chapter~\ref{chapter:mt}, we investigate how to apply pretrained MLMs to non-autoregressive machine translation, by initializing both the encoder and decoder from the same model.
The MLM pretraining objective is well-suited for application to non-AR generation, but we find that parameter transfer must be approached carefully, as MLMs are only trained as encoders, so the addition of cross-attention adds complications.
We identify methods for overcoming this barrier, and test the resulting method on the WMT'14 De-En and WMT'16 Ro-En datasets using the pretrained XLM-R MLM, and the SUNDAE non-autoregressive translation method.
This leads to an improvement of 2.3 and 7.5 BLEU on the the De-En and Ro-En datasets respectively.
We also apply our method to code translation on the CodeXGLUE Java-C\# benchmark using the pretrained CodeBERT model, leading to an improvement of 35-38 CodeBLEU compared to training from scratch.

\subsubsection{Exploration of the ``Bad mode'' problem, and proposed mitigations}
Chapters~\ref{chapter:modes_theory}--\ref{chapter:modes_beam} look at different aspects of one issue: the ``bad mode'' problem.
\citet{catgotyourtongue} showed that neural machine translation (NMT) systems often predict that the single most likely translation of a given input is the empty sequence.
This is clearly not what we want from our systems, and this mismatch is what we will refer to as the bad mode problem.
(For discussion on much more prior work on this issue, see Section~\ref{sec:modestheory_related_work}.)

We will summarize the main thrust of the argument here, but see Chapter~\ref{chapter:modes_theory} for much more detailed exposition.
The key point is that we believe the bad mode problem has too quickly been interpreted as direct evidence of model error, as opposed to being a mix of model error and properties of the distribution the data is trained on.
To support that point of view, we discuss several distributions that have the property that any model which perfectly fits them will necessarily also display the bad mode problem.
This is closely related to the idea of the \emph{typicality} of outputs, which (informally) is the extent to which they are similar to an average output.

Several prior works (most notably \citet{typical_sampling} and \citet{inadequacyofmode}) have emphasized the fact that high scoring model outputs are atypical, and have proposed methods that avoid producing such outputs.
We differ in that we describe cases atypical outputs can be \emph{desirable}.
The fact that these \emph{good} outputs are atypical means they can't be found by methods which focus only on typical outputs or samples.

Based on our reasoning in Chapter~\ref{chapter:modes_theory}, we investigate the possibility of finding \emph{conditional} modes rather than the unconditional mode for a given input.
We experiment with finding unconditional and conditional exact modes for an NMT and cloze completion model.
For these models, length-conditional modal outputs (i.e., the single output of a given length to which the model assigns the highest probability) are usually high quality.
This comes as a surprise, as prior work suggested that model likelihood and quality diverge in general at some point.
For these models, it seems that fixing \emph{only} the length degeneracy problem is sufficient to make the global mode fluent and high-quality.
We also search for exact unconditional modes of several LMs from the LLaMA family~\citep{llama}, and find that they show a mix of degeneracies, meaning that simple length conditioning won't fix the problem.

In Chapter~\ref{chapter:modes_beam}, we attempt to bridge the gap from expensive exact search, to approximate search which might be useful in practice.
We derive a variant of beam search which allows for approximate mode search using conditioning information predicted by a classifier.
The goal isn't to condition on quality, which would be difficult to define, but purely to condition on ``the output is not degenerate'', for whichever particular degeneracies a model's modal outputs show.
This method is related to other methods for conditional generation from prior work, which we discuss in detail in Section~\ref{sec:modesbeam_alternatives}.

Using length-conditional beam search, we are able to find outputs of a given length which score higher and are higher quality than outputs of that length found by beam search.
Again, this conflicts with the intuition that likelihood and quality are not well-related once we enter the atypically-high likelihood regime.
As a more potentially useful demonstration, we also demonstrate that we can block LLaMA-7B from displaying several types of degenerate outputs (primarily emptiness and repeating the prompt), on an instruction following task.
This is done using two classifiers each trained using only 500 binary labeled examples.
Despite this tiny amount of data, we are able to demonstrate fluent outputs from beam search on a ``raw'' (pretrained only) language model.
In general, beam search (or other search methods) are not used for open ended tasks (see for example \citealp{wiher2022ondecoding}), so we consider this a significant step towards making search-based strategies more viable for NLG.

Chapters~\ref{chapter:modes_theory}--\ref{chapter:modes_beam} constitute the first steps of a research program with the goal of improving the outputs of NLG systems \emph{without} needing to resort to finetuning.
The motivation of this program is that finetuning methods such as RLHF seem to necessarily require damaging the model's coverage of its pretraining data distribution. This is elaborated on in Chapter~\ref{chapter:modes_theory}, but see for example \citet{huse} for evidence that human quality judgements are usually making a tradeoff against measures of diversity in model outputs.
If we can get high-quality outputs without needing to damage the underlying model, we will \emph{also} be able to get high-diversity outputs when that's what we want.
We imagine that in the future, search-based methods such as the conditional beam search proposed in Chapter~\ref{chapter:modes_beam} will be used when precision is called for, and sampling methods will be used for creative tasks where variability is more desired than precision.

\subsection{Summary of contributions}
This section has outlined the contributions of this thesis that will let practitioners make more effective use of pretrained language models.
Chapter~\ref{chapter:recurrence} and Chapter~\ref{chapter:hso} discuss new techniques for finetuning transformers which add utility to the large amount of already extant pretrained models, especially those for which new replacements aren't being trained several times a year.

Chapters~\ref{chapter:modes_theory} and \ref{chapter:modes_exact} provide a more nuanced understanding of widely observed problems in generation from language models, using both theoretical and empirical arguments.
The hope is that this will lead to NLP practitioners being able to better understand how to improve the output of their NLG systems, or to the development of new techniques for NLG.
In Chapter~\ref{chapter:modes_beam} we propose one such method, and show that we are able to produce high-quality outputs using search-based methods.

\chapter{Adding Recurrence to Pretrained Transformers}\label{chapter:recurrence}
As discussed in Section~\ref{sec:relatedwork_llm}, cecent progress in NLP has been dominated by large pretrained transformer neural networks~\citep{vaswani2017attention}, such as BERT~\citep{bert}, and GPT-2~\citep{gpt2}.
However, using these models is extremely resource intensive.
Although architectural innovations such as those of \citet{reformer} and \citet{pg19} mitigate this and the issue of a predetermined maximum context size, large pretrained models applying these techniques are not available at this time.
Even if large pretrained models of this kind are released in the future, they will likely not cover the wide range of domains that BERT-family models have been published for. For example, there have been BERT-based models trained for other languages such as French~\citep{flaubert,camembert}, Italian~\citep{alberto}, and many other languages (see~\citet{bertlangoverview} for an overview) as well as specific domains such as scientific papers~\citep{scibert}, biomedical papers~\citep{biobert}, and health records~\citep{medbert}.
Individuals working with these models may not have the resources to train new models from scratch using the latest tricks, as the computation requirements for pretraining are extremely high. As such, identifying ways that already existing models can be improved could be widely impactful.

Another drawback of this family of models is that they have an a priori fixed maximum context size (typically 512 or 1024 tokens for the currently available pretrained models). A typical application of pretrained language models is producing contextual embeddings for a document.
If the document is simply chunked into disjoint segments of 512 tokens, tokens at the boundary of a window will have less contextual information than tokens in the center of a window.
This can be mitigated by striding the evaluation of the model, and only keeping the embedding for a token which has the largest context---but this adds quite a bit of wasted computation.

In this work, we propose a method for augmenting and fine-tuning pretrained transformer language models to use context without directly attending to it.
Our method simultaneously allows for increasing the context size a transformer processes, while allowing a controllable trade-off between computation and perplexity.
We accomplish this by adding a small recurrence module that computes a fixed size representation from the transformer hidden states in a window of text.
Then, the representation for that window is used during processing of the next window.
Shrinking the window size is then a way to reduce the memory footprint of the model, with less loss of performance than would occur with a standard transformer.
Our experiments add recurrence GPT-2 language models, and fine-tune them on the PG-19~\citep{pg19} and WikiText-103 corpora~\citep{wikitext103}, and require only the same amount of memory used for standard fine-tuning of a pretrained language model.
We demonstrate improvements in perplexity compared to a baseline model using the same amount of computation.
Qualitative analysis shows that our recurrent module propagates certain information from previous windows of text, which can facilitate handling of long-distance dependencies with fixed-size input windows.

\section{Related Work}
Many methods have been proposed to lower the memory footprint or computation time of transformer language models, or allow them to be used on larger contexts.
The Transformer-XL~\citep{dai-etal-2019-transformer} allows a position within an attention window to attend to tokens from the previous windows by introducing relative position embeddings.
While that mechanism, like ours, allows information to flow between windows, existing BERT and GPT-2 models do not use relative position embeddings, so training from scratch would be necessary to take advantage of this architecture.

We list here some other modifications of the transformer architecture, somewhat imprecisely grouping them for brevity. For a more detailed discussion, see \citet{transformer_survey}.
\citet{child2019generating}, \citet{qiu2019blockwise}, \citet{reformer}, \citet{sukhbaatar-etal-2019-adaptive}, and \citet{routing_transformers} introduce sparsity to self-attention in various forms, reducing its memory cost.
\citet{pg19} and \citet{longformer}---dynamically and statically respectively---add extra tokens to attend to which allow for global passing of information.
\citet{Tay2020SynthesizerRS} and \citet{Wu2019PayLA} replace dynamically computed self-attention with cheaper alternatives. While the above methods all allow for a reduction in computation, 
they also all require 
training from scratch. Our goal is to allow more efficient and powerful use of the wide array of existing pre-trained models that cover many domains.

\citet{Cao2020DeFormerDP} propose the DeFormer, which also modifies the execution of a pretrained transformer. However, unlike our method, they decompose a single window into multiple windows by removing the attention interactions between these windows. This is largely orthogonal to our method, as one could both decompose windows of text, and additionally use our method to allow information to be passed between neighboring windows. Similarly, distilled versions of pre-trained models such as DistilBERT~\citep{distilbert} provide more computational efficiency, but could be combined with our method to apply them to longer contexts, or reduce the quadratic cost of self-attention.

\citet{modeling_recurrence} apply pre-trained transformers recurrently for machine translation, but do so by using an attention network to embed the document, applying a recurrent encoder to those embeddings, and using the recurrent encoder alongside a typical transformer encoder. This differs from our method as we are fine-tuning language models, which are transformer decoders, and directly modifying the transformer's computation with a recurrent connection, rather than running an RNN on top of embeddings produced by a transformer.

More recently, \citet{flashattention2} introduced an efficient method for computing softmax attention which does not require memory quadratic in sequence length.
Our proposed method can take advantage of this by using the efficient attention implementation within each attention block, and still leads to a reduced cost of attention due to the fact that attention is only used inside blocks, not between them.
However, we leave the interaction between the method proposed here and modern attention kernels/models to future work.

\section{Method}\label{method}
\begin{figure}[h]
    \centering
    \begin{minipage}[b]{0.45\textwidth}
        \centering
        \includegraphics[width=\textwidth]{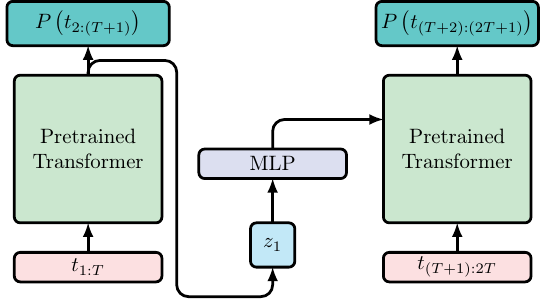}
        \captionof{figure}{Augmenting a pretrained transformer with a recurrence module, allowing reduction of attention computation as well as simpler processing of longer contexts.}\label{fig:rec_model}
    \end{minipage}\qquad%
    \begin{minipage}[b]{0.45\textwidth}
        \centering
        \includegraphics[width=0.9\textwidth]{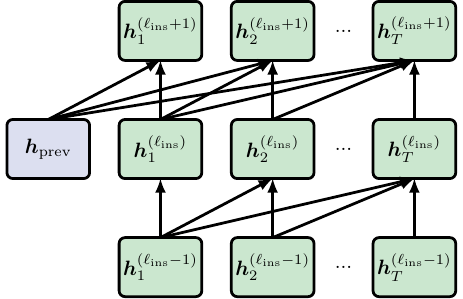}
        \captionof{figure}{$\bm{h}_{\text{prev}}$ is added as an additional key and value to one self-attention layer. Arrows show which positions can pass information to which other positions.}\label{fig:rec_token_insert}
    \end{minipage}
\end{figure}

The main idea of our method is to take a transformer that was pretrained in a fixed context size setting and add recurrence at the level of $T$-token windows of text.
For example, instead of executing the model on one 1000 token window of text, we could instead execute our model with 10 windows of 100 tokens.
The first window is processed by the transformer model as normal, but for subsequent windows we add a supplementary embedding, which is generated using the hidden states from the preceding window (see Figure~\ref{fig:rec_model}).
The recurrence module is extremely small compared to the size of transformer language model, so the additional computation required is negligible.

\subsection{Adding recurrence to pretrained transformers}
Starting by defining terms, we will consider a pretrained transformer with $L$ layers, a hidden state size of $k$, and a maximum context size of $T$ tokens. Let $\bm{h}_i^{(\ell)} \in \mathbb{R}^{k}$ be the output of the $\ell$-th layer of the pretrained model, at position $t$. To produce a fixed-size representation of tokens $t_1, t_2, \dots, t_T$, the embeddings produced by the pretrained transformer are mean-pooled as follows:
\begin{equation}\label{eq:pool}
    \bm{z}_1 = \sum\limits_{i=1}^T \sum\limits_{\ell=1}^L w_\ell \bm{h}_i^{(\ell)}
\end{equation}
where $w_\ell$ are weights softmax-normalized from learned parameters $\alpha_\ell$:
\begin{equation*}
    w_\ell = \frac{\mathrm{e}^{\alpha_\ell}}{\sum\limits_{j=1}^L \mathrm{e}^{\alpha_j}}
\end{equation*}

The fixed-size representation, $\bm{z}_1$, is passed through a feedforward network to produce an embedding $\bm{h}_{\text{prev},1}$ which represents the tokens processed so far, $t_{1:T}$.
Next, instead of evaluating the pretrained transformer without modification on positions $T+1$ through $2T$, $\bm{h}_{\text{prev},1}$ is inserted at a single layer (denoted $\ell_{\text{ins}}$) of the pretrained model, as an additional embedding that may be used in the computation of attention, as shown in Figure~\ref{fig:rec_token_insert}.
To keep the number of embeddings per layer fixed, this embedding is only used as a key and a value, but not a query, in the self-attention layer.
That is, for a window size of 300 tokens, there are 301 inputs to layer $\ell_{\text{ins}}$, but still only 300 outputs.
The embeddings for positions $T+1$ to $2T$ are then pooled in the same way as Equation~\ref{eq:pool} to produce $\bm{z_2}$ and passed through the feedforward network, outputting $\bm{h}_{\text{prev},2}$. $\bm{h}_{\text{prev},2}$ is used to modify the execution of the pretrained language model on tokens $2T +1$ through $3T$, and so on. Because the model is now being applied recurrently, it is trained end-to-end with backpropagation through time.

One could consider more complex recurrence modules, other methods for pooling the previous window's embeddings, or for inserting $\bm{h}_{\text{prev}}$ into the computation for the next window. We experimented with modifications such as max pooling instead of mean pooling, inserting multiple embeddings into the next window, inserting an embedding at all layers of the transformer for the next window, and using fixed key attention as the pooling function. However the performance after each of these changes was not significantly better than the model presented above, so we do not include those modifications here.

\subsection{Gradient checkpointing in networks with bottlenecks}
While our method can reduce the quadratic cost of attention by splitting the input into windows, we can also easily apply it to much longer contexts by use of gradient checkpointing~\citep{chen_checkpointing}.

Gradient checkpointing is a method for lowering the peak memory requirement of training large neural networks. This is accomplished by storing only a subset of activations during the forward pass, and recomputing forward from those cached states during the backwards pass. For example, in a 100 layer feedforward network with uniformly wide layers, one could store the output of only every 10th layer. Then, during the backward pass, in order to compute the gradients for the 95th layer, one would re-compute layers 91 through 99 using the stored 90th layer activations. The overall memory cost is reduced to $\sqrt{L}$ at the cost of a single additional forward pass.

In a network with variable width, the memory reduction can be even larger. When gradient checkpointing is applied to transformers, the outputs of each layer are usually stored ($k\times L \times T$ values), so that at most one set of self-attention activations is in memory at once. In the case of our recurrent models, we have an even narrower bottleneck: the $\bm{z}_i$'s and $\bm{h}_{\text{prev},i}$'s. Storing only these values means that the maximum number of activations present in memory while training on sequences $N$ tokens in length is $M + 2k\lceil{\frac{N}{T}}\rceil$, where $M$ is the number of activations stored when training the transformer on an individual window of length $T$. Because $k$ is extremely small compared to $M$, our model can be applied to very long contexts on any GPU on which the pretrained model can be fine-tuned.

\section{A note on the evaluation of transformer language models}\label{transformer_eval}
Before describing the empirical evaluation of our method, we discuss how transformer language models are evaluated in related work. The standard way of measuring perplexity uses extra computation in order to make as much context available for each token prediction. This yields low perplexities, but does not reflect how practitioners use transformer language models in applications. In this section, we describe the situation in detail and propose practical solutions that achieve relatively low perplexities while being closer to how transformers are used in practice.

\subsection{Potential misalignment between LM evaluation and application}
Transformers are often described as having quadratic time complexity in comparison to RNNs which have linear time complexity. However, this can be somewhat misleading when it comes to evaluation of perplexity. 
Given a test set of length $N$, an RNN requires $O(N)$ time to evaluate---but reaching the best perplexity for a transformer requires $O(NT^2)$, where $T$ is its maximum context size. (These time complexities exclude hidden state size, number of layers, and batch size.)
    This much higher time complexity is due to the fact that a transformer may be run with its full context size once for each token in the test set, so that the maximum context is available for each prediction.
    Re-execution of the whole model for each token is required for models with absolute position embeddings, since hidden state reuse is only possible up to the maximum context size of the network.
    Note that it is possible to achieve smaller wall-clock time by splitting evaluation of a test set over multiple GPUs, but this is not applicable to the generation setting where outputs depend on prior ones.

\begin{figure*}
    \centering
    \begin{subfigure}{0.3\textwidth}
        \centering
        \includegraphics[width=0.8\linewidth]{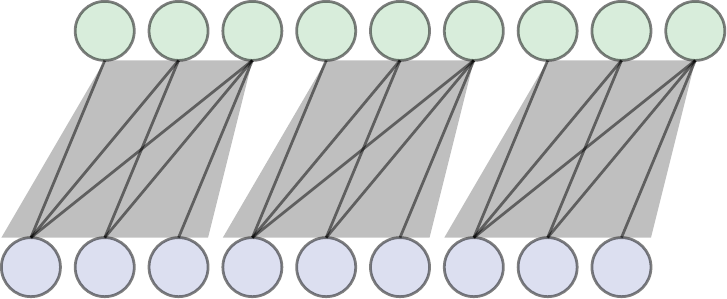}
        \caption{Disjoint execution. Predictions have context ranging between 1 and 3 tokens.}\label{fig:disjoint}
    \end{subfigure}
    \hfill
    \begin{subfigure}{0.3\textwidth}
        \centering
        \includegraphics[width=0.8\linewidth]{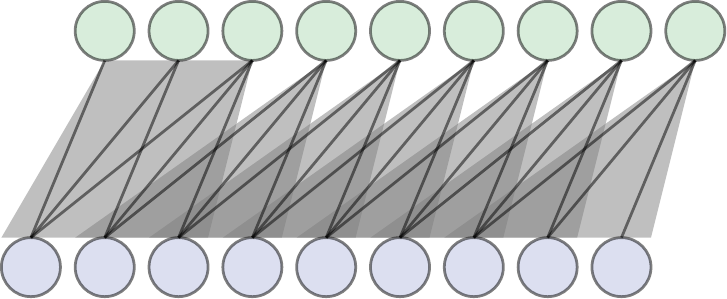}
        \caption{Maximum overlap. All predictions except the first two have maximal context. }\label{fig:max_overlap}
    \end{subfigure}
    \hfill
    \begin{subfigure}{0.3\textwidth}
        \centering
        \includegraphics[width=0.8\linewidth]{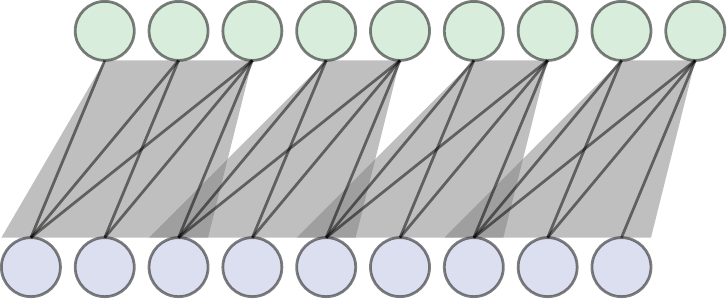}
        \caption{Intermediate degree of overlap. Except the first prediction, all predictions attend to at least 2 tokens of context.}\label{fig:partial_overlap}
    \end{subfigure}
    \caption{Varying degree of overlap while evaluating a transformer with a window size of 3. The green (top) circles are outputs, and the blue (bottom) circles are inputs.\label{fig:overlap}}
\end{figure*}

    To illustrate why re-computation is necessary, consider executing GPT-2 (which has 1024 position embeddings) on a test set.
    Each of the first 1024 tokens of a test set will have been passed into the network using a distinct position embedding.
    Having exhausted the position embeddings, one option is to start again with the 1025th token being treated as position 1---we will refer to this as \emph{disjoint execution}, illustrated in Figure~\ref{fig:disjoint}. The issue with disjoint execution is that it requires predicting the tokens at the beginning of a window from a very small amount of context.

    The alternative, which is used for standard test set evaluation, is \emph{overlapped execution}, as shown in Figure~\ref{fig:max_overlap}.
    The position embeddings are advanced by one position for each prediction, meaning that $T-1$ tokens are repeated between consecutive evaluations of the transformer, requiring much more computation.
    The benefit of this method is that it allows a model with $T$ position embeddings to have $T$ tokens of context for each prediction, as opposed to a variable amount between 1 and $T$.

    Stepping a transformer decoder forward one token at a time measures the best that such a model could perform, but it reflects a generative story that does not align with how the models may be used in practice.
    A perplexity that only measures the ability of GPT-2 to generate the 1024th token given a context of 1023 tokens is not necessarily indicative of the model's performance when generating from a smaller context.
    For example, the popular website Talk To Transformer\footnote{https://talktotransformer.com/} generates samples from GPT-2, but only provides 150 tokens of output.
    The evaluation of GPT-2 by stepping forward one token at a time provides little information about the quality of such generations.

    An example where the discrepancy is length instead of brevity is the GPT backed text adventure game AI Dungeon.\footnote{https://aidungeon.io/. Note that AIDungeon now uses the OpenAI GPT-3 API, but a similar project without OpenAI API access would still have to use GPT-2.} In this setting, the number of tokens can easily reach and exceed the full context size GPT-2 was trained on.
    Using overlapped execution as described above, generating each token would be 1024 times slower than with disjoint execution, so perplexity calculated by overlapped execution does not match this use case either.

While lower perplexity seems to correspond to better generation with shorter contexts in practice (perhaps due to parameter sharing between all sequence positions), there is no reason that this need be the case in principle. To demonstrate an extreme case of the concern being discussed, let $F$ be a transformer model with vocabulary $V$, which uses the previous 1023 tokens as context, and consider the following generative story for generating token $t_i$:
\begin{equation*}
    t_i \sim \begin{cases}
        \text{Uniform}(V) & \text{if $i \le 1023$}\\
        F(t_{(i-1023):(i-1)}) & \text{otherwise}
    \end{cases}
\end{equation*}
Clearly the above generative model would not be of any practical use for generation or otherwise. However, because perplexity is calculated per token, increasing the size of the test set will lead to a measured perplexity that approaches that of a standard evaluation of the model $F$.
This example is not representative of the models that are trained in practice, as even generations much shorter than the maximum context size from a GPT-2 model are quite impressive.
However, it does demonstrate that the criteria that we use to compare models, or to select the best model during early stopping, place very high weight on the ability of the model to produce text given a full context, and a potentially vanishingly small amount on its ability to generate text using shorter contexts.

\subsection{Varying overlap for evaluation}\label{overlap}
As we are interested in increasing computational efficiency at evaluation time for pretrained models, we investigate their performance using overlapped execution, but with a reduced degree of overlap between windows.
Varying the overlap lets us investigate the connection between degree of overlap and perplexity.
The overlap used in evaluation will be defined to be the number of tokens from each input window that are repeated in the next window (see Figure~\ref{fig:overlap}).
For example, consider a window size $T = 10$ and an overlap of 3.
The windows that the transformer will be executed are then $t_{1:10}$, $t_{8:17}$, $t_{15:24}$, \textellipsis, $t_{1 + 7n:10 + 7n}$where $n$ indexes the window.
These input windows are used to predict the spans of tokens $t_{2:11}$, $t_{12:18}$, $t_{19:25}$, \textellipsis, $t_{5 + 7n:11+ 7n}$.
Figure~\ref{fig:partial_overlap} illustrates an intermediate overlap setting with $T=3$ and an overlap of 1.
The perplexity-minimizing evaluation setting 
is then the extreme with an overlap $T-1$, and an overlap of 0 corresponds to disjoint execution.

While a 
standard 
transformer can be evaluated with any degree of overlap, our augmentation method produces the embedding $\bm{h}_{\text{prev}}$, which is used during training to help predict the first token of a window. 
If we change the overlap at test time, the alignment of the text represented by $\bm{h}_{\text{prev}}$ and the current window 
will be different than the model was trained for, and so performance will degrade. To address this, we use the same overlap that will be used at test time during training for the recurrent models.\footnote{Evaluating recurrent models trained with no overlap between adjacent windows on a different level of overlap is possible by changing which positions are pooled. We found that it led to a slight increase in perplexity, so we report results with training and evaluation matching.}

\section{Experiments}
We now present experiments comparing our proposed technique to the default usage of transformer language models. We describe experiments on the WikiText-103 corpus and a subset of the PG-19 corpus, using the GPT-2-small language model as the pretrained transformer in our models. 

WikiText-103 is a standard language modeling corpus composed of approximately 29,000 documents from English Wikipedia, containing 103 million words. We use the WikiText-103 ``raw'' corpus, which does not have rare words replaced by ``UNK''. While GPT-2 uses BPE tokenization, we compute perplexity using the number of words rather than the number of BPE tokens for clarity.

Although WikiText-103 does test long term dependencies, many of the documents are still shorter than the context size of the models we test. Therefore, we also use PG-19, which consists of books from the Project Gutenberg corpus. The average length of a WikiText-103 document is 3.6K words, while PG-19 documents (i.e.\ books) average 69K words, which far exceeds the context size of the models we test. However, the full PG-19 dataset is over 20 times larger than WikiText-103, so we use only a subset of it for training due to computational constraints. Specifically, we use only the first (alphabetically by filename) 1250 books of the PG-19 corpus, and use only the first 15000 tokens of each of the books in the validation set for early stopping. We make no modifications to the test set.

In all our experiments we use the HuggingFace implementation of the pretrained GPT-2 small model (12 layers, 768-dimensional hidden state). For both the recurrent and baseline models, the GPT-2 model was fine-tuned, not left frozen. We selected learning rates for both our models and the baseline separately, by evaluating on WikiText-103 for the same set of candidate learning rates. We used the same learning rates for the PG-19 experiments without further hyperparameter search. We fine-tune all models for 2 epochs, measuring the validation loss every 2 million tokens. All models were trained with Adam~\citep{adam}, warming the learning rate up linearly from 0 to its final value over 100 steps. The feedforward network used to produce $\bm{h}_{\text{prev}, i}$ from window $i-1$ consisted of 3 hidden layers with dimension 200. We fixed $\ell_{\text{ins}}$ to be 2.

Recall from Section~\ref{transformer_eval} that we are interested in evaluating the models in a setting similar to how they would be used in practice. To that end, we report separate perplexities for different degrees of overlap between adjacent windows of text, as described in Section~\ref{overlap}. For our models, we train with the same overlap that we test with, as unlike the baseline models, they cannot be trained with no overlap between adjacent windows and then tested with an overlap. This is because the embedding of the previous window of text is expected to represent all tokens up until the first token of the current window, but with an overlap of 30 for example, that embedding would be representing all tokens up until the 30th token of the current window.

\subsection{Results}
We first show that with the same amount of fine-tuning, our method achieves lower perplexity than a baseline GPT-2 model when evaluated using the same window size and degree of overlap between adjacent windows of text.

It is important to emphasize that the perplexities we report are based on pretrained models, and so should not be compared to models trained from scratch on these datasets. The GPT-2 models were trained on text from a web crawl from which all Wikipedia documents are removed, but this still leaves open the possibility of quotes from Wikipedia having been encountered, or text from PG-19.

\begin{table*}
\centering
\caption{Results on WikiText-103}\label{tab:rec_wikitext}
\begin{tabular}{p{3cm}cp{2cm}p{3cm}c}
    Model & Overlap & Validation Perplexity & Test Perplexity & FLOPs/token\\
    \midrule
    \multirow{5}{3cm}{GPT-2 (small), 300 token window} & 0 & 29.00 & 30.47 & $1.75\times 10^{8}$\\
    & 5 & 27.99 & 29.36 & $1.78\times 10^{8}$\\
    & 10 & 27.58 & 28.88 & $1.81\times 10^{8}$\\
    & 30 & 26.72 & 27.96 & $1.94\times 10^{8}$\\
    & 50 & 26.17 & 27.31 & $2.10\times 10^{8}$\\
    \midrule
    \multirow{5}{3cm}{Recurrent, 20 windows of 300 tokens (Ours)} & 0 & 27.70 & 29.01 & $1.75\times 10^{8}$\\
    & 5 & 26.88 & 28.12 & $1.78\times 10^{8}$\\
    & 10 & 26.51 & 27.77 & $1.81\times 10^{8}$\\
    & 30 & 25.90 & 27.12 & $1.94\times 10^{8}$\\
    & 50 & 25.53 & 26.73 & $2.10\times 10^{8}$\\
\end{tabular}
\normalsize
\end{table*}

Table~\ref{tab:rec_wikitext} shows the perplexity of our models and the non-recurrent GPT-2 models on the WikiText-103 dataset.
The models compared here all use windows of 300 tokens, with varying degrees of overlap.
The baseline models can only access information from the previous window of text through the overlapping tokens, while the recurrent models have a fixed size representation of the longer context.
Our addition of recurrence increases the performance of the GPT-2 models in this setting, but by a relatively small amount.
Increasing the overlap between each window of text decreases the perplexities of the baseline model as expected, but also decreases the perplexity of the recurrent models.\footnote{We did not attempt to train recurrent models with extremely high overlaps, as that would greatly increase the required training time.}
This indicates that there is room to increase the capacity of the recurrence mechanism, as if it passed all relevant information about the previous window forward, the overlapping tokens would be redundant.
On the other hand, some useful information beyond what is contained in the local context is being propagated, as otherwise the baseline model should catch up in perplexity at higher overlaps.
To investigate this further, we also experiment with the PG-19 dataset.

The results for the PG-19 experiments are shown in Table~\ref{tab:rec_pg19}.
While we find only small increases in performances on the WikiText-103 dataset, we see larger improvements on PG-19, confirming our prediction that the gains would be larger on a dataset that has a larger context available for each prediction on average.
We find that adding our recurrence module leads to a model that gives as good a perplexity with no overlap between adjacent windows as an unmodified model does when evaluated with an overlap of 30 out of 300 tokens in each window.
Training the recurrent model with a 5 token overlap gives perplexity lower than the baseline perplexity with an overlap of 50 or even 75. In terms of FLOPs, adding our recurrence module and overlapping adjacent windows of tokens by 50 is less than half as costly as using a non-recurrent model with an overlap of 200.

\begin{table*}[t]
\centering
    \caption{Results on PG-19}\label{tab:rec_pg19}
    \begin{tabular}{p{3cm}cp{2cm}p{3cm}c}
    Model & Overlap & Validation Perplexity & Test Perplexity & FLOPs/token\\
    \midrule
        \multirow{9}{3cm}{GPT-2 (small), 300 token window} & 0 & 172.25 & 147.71 & $1.75\times\mathrm{10}^8$\\
        & 5 & 165.93 & 142.30 & $1.78\times 10^8$\\
        & 10 & 162.66 & 139.49 & $1.81\times 10^8$\\
        & 30 & 156.21 & 134.30 & $1.94\times 10^8$\\
        & 50 & 152.64 & 131.25 & $2.10\times 10^8$\\
        & 75 & 149.54 & 128.46 & $2.33\times 10^8$\\
        & 100 & 147.05 & 126.51 & $2.62\times 10^8$\\
        & 150 & 143.62 & 123.53 & $3.50\times 10^8$\\
        & 200 & 141.14 & 121.40 & $5.25\times 10^8$\\
    \midrule
        \multirow{5}{3cm}{Recurrent, 20 windows of 300 tokens (Ours)} & 0 & 155.27 & 133.02 & $1.75\times 10^8$\\
        & 5 & 150.00 & 128.78 & $1.78\times 10^8$\\
        & 10 & 147.53 & 127.05 & $1.81\times 10^8$\\
        & 30 & 142.35 & 122.22 & $1.94\times 10^8$\\
        & 50 & 140.10 & 119.93 & $2.10\times 10^8$\\
\end{tabular}
\end{table*}

\subsection{Effect of window size}
\begin{figure}
\centering
    \begin{minipage}[b]{0.45\textwidth}
        \centering
        \includegraphics[width=\textwidth]{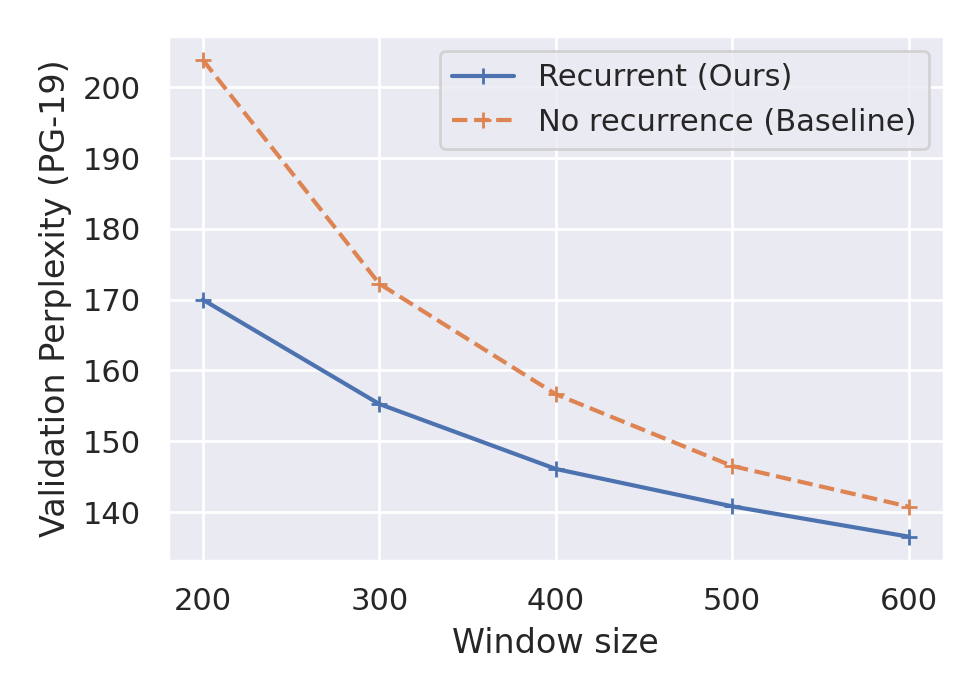}
        \captionof{figure}{Effect of window size on performance on PG-19 validation set.\\\mbox{}\\\mbox{}}\label{fig:rec_window_size}
    \end{minipage}\qquad%
    \begin{minipage}[b]{0.45\textwidth}
        \centering
        \includegraphics[width=\textwidth]{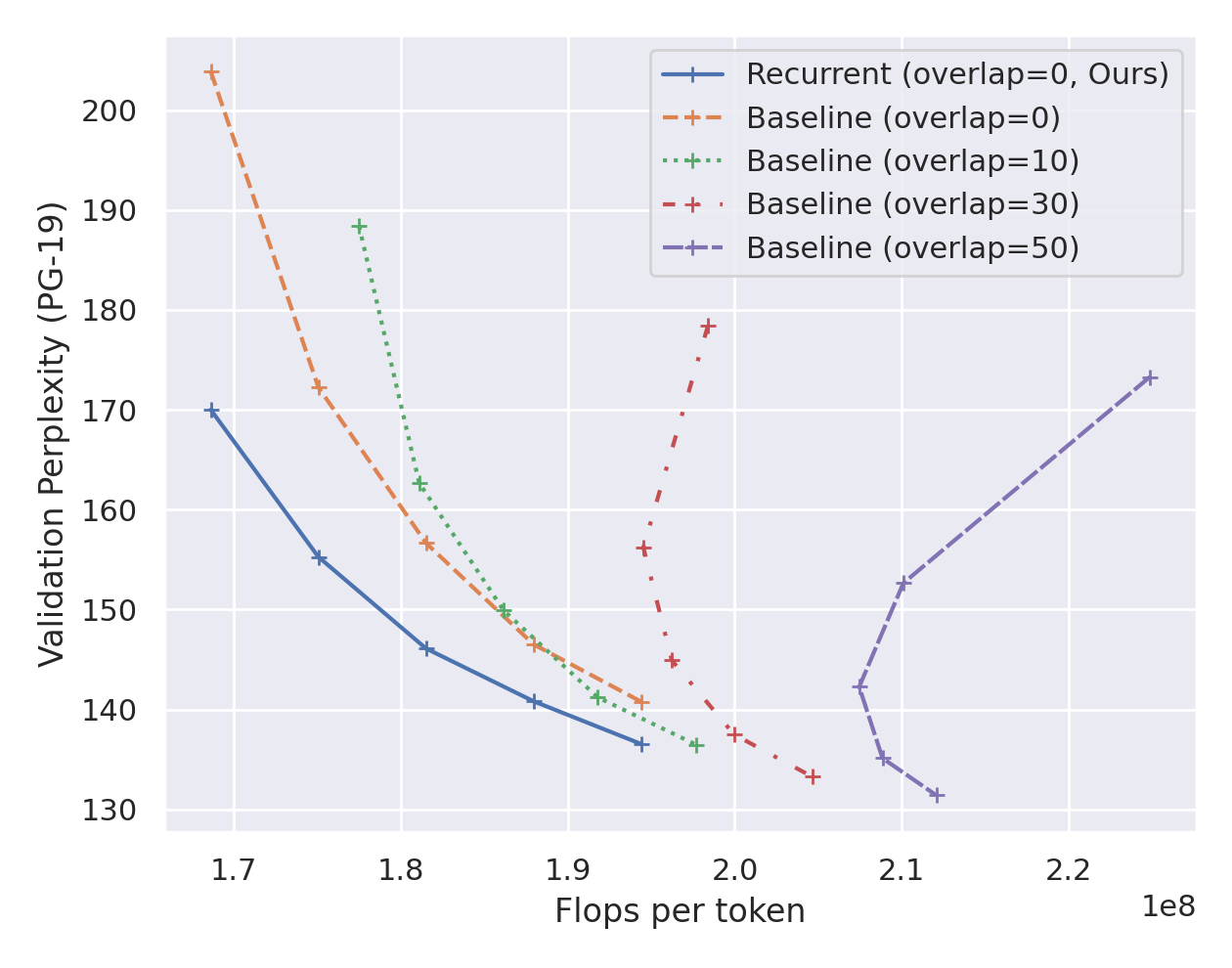}
        \captionof{figure}{Relationship between FLOPs and perplexity for recurrent and non-recurrent models. Curves range over window sizes from 200 to 600.}\label{fig:rec_flops}
    \end{minipage}
\end{figure}

As one of our motivations is to retain performance while decreasing compute requirements, we experiment with varying the window size used by our model and an unmodified GPT-2 model. At smaller window sizes the recurrent model has access to much more information than 
GPT-2, which can only attend to the current window. 
Because of this, we expect our augmentation to cause the performance to fall off less rapidly with decreasing window size. The results, shown in Figure~\ref{fig:rec_window_size}, confirm this prediction, as the performance gap widens with smaller windows. 
Figure~\ref{fig:rec_flops} contains the same points (and additional baseline curves for various overlaps), but in terms of FLOPs rather than window size. 
All of the results of the recurrent models lie on the Pareto frontier, meaning that to improve perplexity or computational cost, one must worsen the other. The non-monotonicity of the overlap 30 and 50 curves is due to the fact that at smaller window sizes, an overlap represents a higher fraction of the computation being used for positions that predictions were already produced for. Also note that while the baseline with overlap 50 curve has the lowest absolute perplexity in Figure~\ref{fig:rec_flops}, the recurrent models trained with overlaps shown in Table~\ref{tab:rec_pg19} still perform better.

\subsection{What information is being propagated between windows?}
\begin{table*}[!h]
    \scriptsize
    \setlength{\tabcolsep}{4pt}
    \begin{tabular}{p{0.48\textwidth}p{0.37\textwidth}p{0.09\textwidth}}
    Context & Continuation & Topic\\
    \midrule
        \textellipsis{} It drains a territory of 300,000 square miles, and, traced back to the rise of its principal source, is 2,000 miles long. At two points, Needles and Yuma on the California boundary, it is crossed by a railroad. Elsewhere its course lies far from Caucasian settlements and far from the routes of common travel, in the heart of a vast region fenced on the one hand by arid plains or deep forests and on the other by formidable mountains & .  The first of the two great rivers, the Mississippi, was the first to be navigated by the Americans, and the first to be navigated by the Indians. The other was the first to be navigated by the Americans, and the first to be navigated by the Indians. The first of the two great streams, the Mississippi, was navigable by the Indians, and the first by the English\textellipsis{} & Rivers \\
    \midrule
        \textellipsis{} ``Mr. Hammerby will have to wait for his money.'' ``Will he?'' ``I don't see what else he can do.  But, aside from that, three dollars and a half won't keep us a month.  I'll have to &  get the money, and he'll be glad to see you again.'' ``I'm glad to see you again, Tom. I'm glad to see you again. I'm glad to see you again. I'm glad to see you again. I'm glad to see you again & Money\\
    \midrule
        \textellipsis{} Long's his sister, Thankful, lived, we couldn't say nothing, of course. If she wanted to slave and work so's her brother could live in idleness and sloth, why, that was her business. There ain't any law against a body's making a fool of herself, more's the pity. But she's been dead a year, and he's done nothing since but live on those that'll trust him, and ask help from the town. He ain't sick--except sick of work. [\textellipsis{}] to-mor & rer and his wife, and the whole family.  He was a good man, and he was a good man.  He was a good man, and he was a good man.  He was a good man, and he was a good man.  He was a good man, and he was a good man.  He & Family, Description of a man\\
    \midrule
        \textellipsis{} and any form of thought or any institution tending to suppress education or destroy intelligence strikes at the very essence of the government, and constitutes a treason which no law can meet, and for which no punishment is adequate. Education, then, as universally diffused as the elements of God, is the life-blood of our body politic. The intelligence of the people is the one great fact of our civilization and our prosperity,--it & is the only way to preserve the peace of the world. The only way to preserve the peace of the world is to preserve the liberty of the people.  The first step to be taken is to abolish the power of the government. The first step to be taken is to abolish the power of the people. The first step to \textellipsis{} & Government\\
    \midrule
\end{tabular}
\normalsize
\caption{Continuations of PG-19 validation set passages using recurrent model ($T=300$, disjoint) which display approximate continuation of topic. Contexts are shortened for space, 300 tokens of context were used for each.}\label{tab:samples}
\end{table*}

\begin{table*}[!h]
    \scriptsize
    \setlength{\tabcolsep}{4pt}
    \begin{tabular}{p{0.57\linewidth}p{0.39\linewidth}}
    Context & Continuation\\
    \midrule
        \textellipsis{} Elaine, however, did not gratify their hopes; she referred to the ball with the detachment she would have shown in describing a drawing-room show of cottage industries.  It was not difficult to discern in her description of the affair the confession that she had been slightly bored.  From Courtenay, later in the day, the aunts received a much livelier impression of& the house, and the two women were standing in the doorway. ``I am afraid I have not been very well,'' she said, as she looked at the two men.  ``I have been very much worried about you, and I have been very much worried about you.  I have been very much worried about you, and I have been very much worried about you.  I have been very much pleased with you, and I have been very much\\
    \midrule
        \textellipsis{} ``Oh, he's not at all a bookworm,'' said Suzette quickly, ``though he's tremendously well-read.  He's quite the man of action.'' ``Does he hunt?'' asked Elaine. ``No, he doesn't get much time or opportunity for riding.'' ``What a pity,'' commented Elaine; ``I don't think I could marry a man who wasn & '' ``I don't know,'' said the Sergeant, ``but I have heard that you are a very good friend.'' ``I am not,'' said the girl, ``I am not.'' ``I am not,'' said the Sergeant, ``I am not.'' ``I am not,'' said the girl, ``I am not.\\
    \midrule
        \textellipsis{} Gregg was still the soft-stepping scoundrel who invited the toe of honesty, and Mornway, as he entered, was conscious of a sharp revulsion of feeling. But it was impossible to evade the interview, and he sat silent while the man stated his case.  Mrs. Mornway had represented the stenographer as being in desperate straits, and ready to accept any job that could be found, but though his appearance might have seemed to corroborate her account, he evidently took a less hopeless view of his case, and the Governor found with surprise that he had fixed his eye on a clerkship & that he had been in the habit of doing, and that he had been in the habit of doing.  ``I'm not going to be a fool,'' he said, ``and I'm not going to be a fool,\textellipsis{}\\
    \midrule
        \textellipsis{} Many transactions too, from a desire to avoid worrying me, were carried through without my knowledge, although formerly, as a matter of course, they would have been submitted to me.  Strangers, when they called, asked to see Johnson or Marsh.  I directed the messenger that they were to be shown into my room if I was disengaged.  This was a failure, for, when they came, I was obliged to ask for help, which was not given very generously.  Sometimes I sent for the papers, but it took a long time to read them, and my visitors became impatient.  During one of these interviews, I remember that I was sorely perplexed, but I had managed to & get the money, and I was glad to get it.  The next day I went to the Court of the Emperor, and found him standing in the courtyard of the Palace, with a large number of soldiers.  He was dressed in a very dirty uniform, and wore a very dirty hat.  He was dressed in a very dirty uniform, and wore a very dirty hat.  He was dressed in a very dirty hat, and wore a very dirty hat\\
    \midrule
        \textellipsis{} I know at which crossings to look out--I know what I'm going to see in the shop-windows. It saves a lot of wear and tear to know what's coming. For a good many years I never did know, from one minute to another, and now I like to think that everything's cut-and-dried, and nothing unexpected can jump out at me like a tramp from a ditch.'' He paused calmly to knock the ashes from his cigar, and Garnett said with a smile: ``Doesn't such a plan of life cut off nearly all the possibilities?'' The old gentleman made a contemptuous motion. ``Possibilities of what? Of being multifariously miserable? There are lots of ways of being & ''I'm afraid I'm not going to be able to do that,'' he said. ``I'm going to have to go to the station. I'm going to have to go to the station. I want to see the station. I want to see the station. I want to see the station. I want to see the station. I want to see the station. I want to see the station. I want to see the station. I want to see\\
    \midrule
\end{tabular}
    \caption{Continuations of PG-19 validation set passages using recurrent model ($T=300$, disjoint) which display continuation of correct pronouns or references to characters from context. Contexts are left-truncated, 300 tokens of context were given for each continuation.}\label{tab:pronouns}
\end{table*}

We now discuss some features that our models display when we compute continuations (using greedy decoding) from contexts in the PG-19 validation set, which illustrate types of information that the recurrent module passes (or fails to pass) forward.

The most common phenomenon we identify in these samples is successful propagation of topical information between adjacent windows. For instance, we see in Table~\ref{tab:samples} a context discussing geography and rivers, followed by a continuation maintaining the same topic, and we see a context discussing the topic of payment, leading to a mention of money in the continuation. Beyond passing of topical information, another success case in the generations is passing of certain information about characters between windows---in Table~\ref{tab:pronouns} we see that pronouns in the continuations often reflect characters mentioned in the context, and we see an example in which the continuation includes ``the two women'', after a context mentioning ``the aunts''. This behavior was likely learned due to the fact that PG-19 consists of narratives, so correctly passing character information between windows is quite beneficial.

However, these examples also contain discontinuities between the context and the continuation, in terms of local syntax or facts of the narrative. We see that some sentences are not completed in the expected form (for instance, ``There are lots of ways of being'' is continued with a new quote rather than completion of the thought), and new characters are sometimes invented rather than continuing to reference those described in the context.  %
One sample has a closing quotation mark, predicted from the previous window, being interpreted as an opening quotation mark.
These are the types of issues that an overlap between adjacent windows easily addresses---a fact that likely accounts in part for the gap between the recurrent model with disjoint and overlapped execution in Table~\ref{tab:rec_pg19}.
A higher capacity recurrent module might fix these issues in exchange for additional computation.

\subsubsection{Quantitative evaluation of topic propagation}\label{sec:recurrence_lda}
To verify the trend we identified of topic propagation in continuations generated by our recurrent models, we fit an LDA topic model~\citep{lda} with 20 topics to 5000 books from the PG-19 training set. Given a bag of words, this topic model will assign a distribution over topics, so we can use a statistical distance as a metric for the similarity between the topics of two segments of text.

We sampled 8000 contexts of 300 tokens from the PG-19 validation set, and computed argmax decoded continuations of 30 tokens from the same models used to generate Table~\ref{tab:samples}.\footnote{The baseline receives one token of context to begin generating from.} We then computed the Jensen-Shannon divergence (JSD) between the topic distribution of each context and the corresponding continuations. This procedure finds that continuations from the recurrent model have an average topic JSD of 0.5331, while those from the baseline model have an average topic JSD of 0.5951. For a given context, the continuation given by the recurrent model is likely to have a lower JSD at least 60\% of the time ($p < 0.00001$).

\section{Conclusion}
We showed that augmenting a pretrained language model with a recurrence module during fine-tuning can allow increased performance given a fixed computational budget.
Our method can be similarly applied to improve the computational efficiency of pretrained models that already exist for many languages and domains, as well as for future models that will be developed.
It can also allow their application to longer contexts than they were trained for, increasing their flexibility.

\chapter[Optimizing Hidden States in Language Models]{Optimizing Hidden States in Language Models\footnote{This work was published as \citet{hsoacl}}}\label{chapter:hso}
\newcommand{\hstate}[2]{\bm{h}_{#1:#2}}
\newcommand{\modstate}[2]{\tilde{\bm{h}}_{#1:#2}}

While Chapter~\ref{chapter:recurrence} focused on a new method for finetuning transformer LMs, this section will focus on improving the performance of pretrained transformers at evaluation time, using no extra finetuning.
One direction of language modeling research particular has attempt to find methods which leave the LM parameters fixed, but somehow update the model's hidden states (e.g., \citealp{Dathathri2020Plug} and \citealp{qin-etal-2020-back}).
In this work, we introduce Hidden-State Optimization (HSO), a method which allows for eliciting higher-quality predictions from any transformer language model.
HSO contributes to the goal of this thesis by generically increasing the utility of pretrained LMs.

HSO %
first computes the language modeling loss as usual, then modifies the LM hidden states using the gradient of the loss (using the original prediction for evaluation, to avoid using future information to inform predictions).
This process is repeated for each window of 10-25 tokens, updating the cached hidden states each time.
Attending to these modified hidden states creates higher quality predictions for future tokens.

We demonstrate HSO in the setting of language model evaluation on the WikiText-103 \citep{wt103} and PG-19~\citep{pg19} corpora, and find  %
improvements in measured perplexity. In order to demonstrate that this translates into value for downstream applications we apply HSO %
to few-shot classification with the 1.5B parameter GPT-2, and find improvement in that setting as well.

\section{Related Work}
\textbf{Learning during inference.} 
HSO is related to methods that perform learning on the test set. 
One such method is dynamic evaluation (DE) \citep{krause2018dynamic,krause2019dynamic}, which was the inspiration for HSO. DE consists of using test inputs for learning after evaluating on them, which means a larger test set will result in a larger gain from its use.
This is not reflective of the small amount of text present in a setting such as conditional generation or few-shot classification, while using HSO for LM evaluation is. HSO is also cheaper than DE because it differentiates with respect to hidden states rather than the model parameters.

\paragraph{Gradient-Based Optimization of Hidden States.}
\citet{qin-etal-2020-back} proposed Delorean, a method that incorporates future tokens into LM predictions by using backpropagation into earlier intermediate vectors. However, their goal is to produce better generations for intermediate timesteps, using sampled intermediate tokens and ground truth future tokens. We instead use the LM loss to tune past hidden states to allow better prediction of unseen future tokens. They also only perform gradient updates to logits while we update hidden states. 

Plug-and-Play language models (PPLM; \citealp{Dathathri2020Plug}) modify the behavior of pretrained LMs %
by updating hidden states at inference time, but with the goal of controllable generation (e.g., controlling sentiment) rather than improved fidelity. Unlike HSO, PPLMs require an attribute classifier which must be trained with labeled data. Several methods have been developed to more efficiently achieve the same goal as PPLM \citep{madotto-etal-2020-plug,krause2020gedi}, and these ideas could potentially be applied in analogous ways to speed up HSO. %

\paragraph{Alternatives to finetuning.} %
Our method is 
related to those that 
reduce the computational cost of finetuning by updating a smaller number of parameters or avoid finetuning altogether. 
\citet{pmlr-v97-houlsby19a} introduce adapter modules which are finetuned in lieu of the full model. 
\citet{DBLP:journals/corr/abs-2101-00190} introduce prefix-tuning, which adds a fixed set of learnable vectors to the beginning of the input sequence. The latter is related to 
using prompts for contextual generation, which has gained popularity both 
to extract 
information from %
language models (e.g., \citealp{radford2019language}, \citealp{jiang-etal-2020-know}) 
and perform tasks directly without updating any model parameters 
\citep{brown2020language}. Follow-up work has sought to understand the effectiveness of prompting \citep{le-scao-rush-2021-many} and automatically find or learn better prompts \citep{shin-etal-2020-autoprompt,DBLP:journals/corr/abs-2101-06804,qin-eisner-2021-learning}.

\section{Method}
Let $f$ be a transformer language model computing the distribution for token $x_{t}$ given tokens $x_{1:t-1}$:

\begin{equation*}
p_{t} = f(x_{1:t-1})
\end{equation*}
In practice, one may cache the hidden states, $\bm{h}_{t} \in \mathbb{R}^{\ell \times d}$, where $\ell$ is the number of layers and $d$ is the embedding size.
We represent this 
by factoring $f$ into
$f_h$ which computes hidden states (possibly depending on past hidden states) and $f_p$ which computes output probabilities from the hidden states:
\begin{align}
\bm{h}_t &= f_h\left(x_t, \bm{h}_{1:t - 1}\right) \label{eq:hidden}\\
p_{t} &= f_p\left(\bm{h}_t\right)\nonumber
\end{align}
Given a loss function $L$ which takes as arguments the ground truth next word and a distribution over word types, one can then compute its gradient with respect to both the present hidden states $\bm{h}_t$, and with respect to the cached hidden states $\bm{h}_{1:t - 1}$:
\begin{align*}
\bm{g}_{\text{present}} &= \nabla_{\bm{h}_t} L\left(x_{t+1}, f_p(\bm{h}_{t})\right)\\
\bm{g}_{\text{cached}} &= \nabla_{\bm{h}_{1:t-1}} L\left(x_{t+1}, f_p(f_h(x_t, \bm{h}_{1:t - 1})\right)
\end{align*}
Denoting the concatenation of these two quantities along the time axis as $\bm{g}_t = \left[\bm{g}_\text{cached};\bm{g}_\text{present}\right]$, we can make a gradient update to the hidden states:
\begin{equation}\label{eq:update}
\tilde{\bm{h}}_{1:t} = \bm{h}_{1:t} - \eta\bm{g}_t
\end{equation}

where $\eta$ is the step size. We apply Adam \citep{kingma-15} to this update, but with modifications described in Section~\ref{adam}.

In practice, we use standard cross entropy as our loss function $L$. So, intuitively, we are updating the hidden states to make the actual word at position $t+1$ more likely under the language model's distribution $p_t$ by altering only the previously computed hidden states. Note that when we update the hidden states with gradient-based updates, it will no longer be the case that the set of hidden states follow the feedforward procedure defined by the architecture of the transformer language model. 

While computing the hidden state for $x_{t+1}$, we then substitute $\tilde{\bm{h}}_{1:t}$ into Eq.~\ref{eq:hidden} in place of  $\bm{h}_{1:t-1}$:
\begin{equation*}
    \bm{h}_{t+1} = f_h\left(x_{t+1}, \tilde{\bm{h}}_{1:t}\right)
\end{equation*}

Provided that the loss for timestep $t$ is computed with the unmodified hidden state $\bm{h}_t$ rather than $\tilde{\bm{h}}_t$, this may be done at test time 
without the loss being improved by ``looking into the future.''
We continue to update all hidden states at each step.\footnote{$\modstate{1}{t}$ is then a concatenation of hidden states which have been updated between 1 and $t$ times.}

In practice taking a gradient step after %
each token is too costly, so we can process blocks of $k$ tokens (which we will refer to as a \emph{window size} of $k$):
\begin{align*}
\bm{h}_{t+1} &= f_h\left(x_{t+1}, \modstate{1}{t}\right)\\
p_{t+1} &= f_p\left(\bm{h}_{t+1}\right)\\
\bm{h}_{t+2} &= f_h\left(x_{t+2}, [\bm{h}_{t+1:t+1}; \modstate{1}{t}]\right)\\
&\vdots \\
\bm{h}_{t + k} &= f_h\left(x_{t + k}, [\hstate{t+1}{t+k-1}; \modstate{1}{t}]\right)\\
p_{t + k} &= f_p(\bm{h}_{t+k})
\end{align*}
This sequence of computations is done in a single forward pass, but we have broken it up by token to make clear how a mix of unmodified and modified hidden states is used to embed each token in the window. Once the loss function, $L$, is applied to $x_{t+2:t+k+1}$ and $p_{t+1:t+k}$, a backwards pass is done to compute the gradient of the sum of the losses with respect to the hidden states, at which point the modified hidden states $\modstate{1}{t+k}$ are computed. %

$k$ has a twofold effect on computational cost, as it controls both the number of gradient steps and the number of tokens processed at a time. A very small $k$ will require many more forward passes and will not take advantage of GPU parallelism. %

\subsection{Modifications to Adam}\label{adam}
One way of applying Adam to the HSO gradient update would be to view the past hidden states as a single $T \times \ell \times d$ tensor, where $T$ is the maximum context size. This would allow use of just two moment estimate tensors $\bm{m}, \bm{v} \in \mathbb{R}^{T \times \ell \times d}$. This version of Adam performs very poorly, as a given value in the hidden state cache will not be consistently associated with the same moment estimate.

Instead, we keep first and second moment estimates $\bm{m}_i$ and $\bm{v}_i$ for each hidden state, discarding them once the corresponding hidden states are further in the past than the maximum attention length. This also requires maintaining a different optimizer step value for each block of $k$ hidden states, as Adam's bias correction depends on how many updates have been made to a moment estimate. In terms of implementation, we do not actually keep a separate vector for each hidden state, but pack them into a tensor which is translated along with the cached hidden state tensor.

\section{Experiments}
We demonstrate HSO with the Transformer-XL (TXL)~\citep{dai-etal-2019-transformer} and GPT-2\footnote{For GPT-2, we backpropagate into the key and value vectors rather than the full embeddings at each layer for ease of implementation. They differ by only a linear transformation, so we do not expect this to be a critical difference.} \citep{radford2019language} models implemented using FLAX~\citep{flax} and Haiku~\citep{haiku}, on top of JAX \citep{jax}. The TXL model is initialized from the HuggingFace Transformers~\citep{hftransformers} model trained on WikiText-103 (WT-103). The GPT-2 models are initialized from the OpenAI checkpoints.

\subsection{Language modeling}
We test HSO with the %
TXL and 345M parameter GPT-2 models on the pre-tokenized WikiText-103~\citep{wt103} and PG-19~\citep{pg19} datasets. As the TXL was trained on WT-103, this covers both an in-distribution and out-of-distribution (OOD) evaluation for it. We found that TXL was not stable in the OOD setting, but that  
resetting its hidden states to zeros upon reaching its maximum context size reduced the baseline perplexity significantly. We do not do this for HSO as it does not appear to need this stabilization. 
We evaluate GPT-2 with non-overlapping contexts for efficiency. The perplexities reported are per token, which differs between GPT-2 and the word based TXL. Out of vocabulary words are UNK-ed for TXL, but GPT-2 has an open vocabulary.

We used a window size of $k=25$, a learning rate of 0.003, and 0.65/0.9 for Adam's $\beta_1$ and $\beta_2$ parameters. We found that some HSO hyperparameter settings gave better performance, especially for GPT-2, but for the sake of parsimony report our main results with consistent hyperparameters.

\begin{table}[t]
\small
\centering
    \begin{tabular}{cll}
        Method & WT-103 & PG-19\\
        \midrule
        Baseline & 21.3/22.4 & 166.4/164.2\\
        HSO & \textbf{20.7}/\textbf{21.7} & \textbf{140.0}/\textbf{145.7}\
    \end{tabular}
    \caption{Language modeling validation/test perplexity with Transformer-XL (pretrained on WT-103). Importantly, PG-19 is out of distribution for this model.}
    \label{tab:transfoxl}
\end{table}

\begin{table}[t]
\small
\centering
    \begin{tabular}{cll}
        Method & WT-103 & PG-19\\
        \midrule
        Baseline & 21.5/20.7 & 26.7/\textbf{26.5} \\
        HSO & \textbf{21.0}/\textbf{20.3} & \textbf{25.1}/26.5\\
    \end{tabular}
    \caption{Language modeling validation/test perplexity with GPT-2 (345M parameters). \label{tab:gpt2}}
\end{table}

Our LM results are shown in Tables~\ref{tab:transfoxl} and \ref{tab:gpt2}. HSO yields about a half a point improvement in perplexity on WT-103 with both architectures. While this is not a large improvement, recall that GPT-2's hidden states are reset every 1024 tokens, so this represents improvement in prediction within the context of one attention window, rather than cumulative training on the test set as in DE.

On PG-19, the perplexity improvements are larger for the most part: 1.6 points for GPT-2 on the validation set and over 10 points for TXL %
(but a <0.1 point increase for GPT-2 on the test set). As we used the same hyperparameters for all LM evaluations, HSO seems to be fairly robust to the choice of architecture and dataset. 

\subsubsection{Modifying HSO}\label{hparams}
Table~\ref{tab:hparams} shows the effect of various modifications to HSO on GPT-2's perplexity on the PG-19 validation set. Tuning Adam's parameters decreases perplexity by another point. Surprisingly, only updating the most recent window's hidden states (``present-only'') improves perplexity on PG-19 (initial experiments on WT-103 did not find this to be the case). This also requires significantly less computation. Since Adam tries to estimate moments over many steps this might seem to imply it is not necessary. To investigate this, we tested stochastic gradient descent (SGD) with several learning rates but it
performed worse than Adam for both full and ``present-only'' updates.\footnote{On the first step, Adam updates in the $L_\infty$ steepest descent direction so it differs from SGD even for only one step.}

\begin{table}[t]
\centering
\small
\begin{tabular}{lc}
Modifications & Perplexity\\
\midrule
None & 25.1\\
$\eta=3 \times 10^{-4}$, $\beta_1 = 0.8$ & 23.8\\
present-only & 23.6\\
$k=10$ & 24.4\\
$k=10$, present-only & 22.1\\
SGD, $\eta=0.01$, & 24.7\\
SGD, $\eta=0.01$, present-only & 25.1\\
\end{tabular}
\caption{GPT-2 (345M) perplexity on the PG-19 validation set. $\eta$ is learning rate, $k$ is window size, ``present-only'' means only the last $k$ hidden states are updated. 
}\label{tab:hparams}
\end{table}

\begin{table}[t]
\setlength{\tabcolsep}{5pt}
\centering
\small
\begin{tabular}{llcccc}
Dataset & $n$& \multicolumn{4}{c}{Method}\\
\cmidrule{3-6}
&& Baseline & DE\footnotemark & HSO & HSO-2\\
\midrule
\multirow{4}{*}{SST-2} & 2 & 53.9 & 52.2 (55.1) & 59.5 & \textbf{64.0}\\
& 4 & 58.3 & 55.6 (58.8) & 63.1 & \textbf{66.5}\\
& 6 & 57.9 & 56.2 (59.4) & 68.0 & \textbf{69.2}\\
& 8 & 58.4 & 59.9 (61.8) & \textbf{70.2} & 70.2\\
\midrule
\multirow{4}{*}{AGNews} & 2 & 53.1 & 32.2 (35.0) & 52.6 & \textbf{54.3}\\
& 4 & \textbf{77.8} & 52.2 (55.2) & 77.2 & 77.6\\
& 6 & 64.8 & \textemdash & 65.8 & \textbf{66.2}\\
& 8 & 63.3 & \textemdash & 68.5 & \textbf{69.3}\\
\end{tabular}
\caption{Effect of updating hidden states on few-shot classification accuracy of GPT-2-XL on SST-2 and AGNews, where $n$ is the number of examples per prompt. Neither hidden states or weights are updated for the baseline. HSO-2 is HSO with two gradient steps per window of text.}\label{tab:fewshot}
\end{table}
\footnotetext{Due to the much higher running time for using dynamic evaluation, these are partial results from running on a random subset of the test set. The accuracy in parentheses is a hypergeometric 95\% upper confidence bound. We exclude $n=6,8$ for AGNews due to the input exceeding the available GPU memory for those input sizes.}

\subsection{Few-shot classification}
While %
HSO can give gains in perplexity, 
we would like to see whether it benefits other tasks as well. 
So, we consider few-shot learning from examples %
in the LM's %
context, as in GPT-3~\cite{brown2020language}.
Lacking GPT-3 access, we demonstrate our method with the 1.5B parameter GPT-2-XL model. %

We use the binary SST-2~\citep{socher2013recursive} and 4-way  AGNews~\citep{zhang2015character} classification datasets. 
We follow choices made by \citet{zhao2021calibrate}, including their prompt formats, but we made several changes to their procedure %
to reduce computational requirements and variance. 
Most importantly, we resampled a class-balanced prompt for every test example (but kept the prompt fixed between the baseline and HSO) rather than using a fixed prompt.\footnote{\citet{zhao2021calibrate} reported %
high variance %
based on prompt choice, %
so we made this choice in order to only need to run each evaluation once. The other two changes %
were to sample 1200 examples from the AGNews test set to expedite the evaluation, and to only use examples with  %
$\leq$35 tokens in our prompts to reduce the required memory.} 
We used a learning rate of 0.01 and a window size of 10 tokens. Our experiments used a 24GB NVIDIA Quadro RTX 6000 GPU.

\davis{Can you take a look at the stuff I've added about DE here?} We also test DE, as in contrast to the LM setting, the amount of fine-tuning data will be the same between DE and HSO. We found that the learning rate of 0.01 led to the model collapsing to constant predictions, so we use a learning rate of $10^{-4}$ instead. We update the model every 10 tokens as with HSO, and recompute the hidden states after each update since the weights which produced them are no longer the model weights.

There are a few options to pick between when deciding what it meant to apply DE to this setting. One could choose to make a single gradient step based on the entire prompt, update the weights every 10 tokens but not recompute the hidden states, or perform multiple updates on the whole prompt. We chose what we believed was the closest comparison between HSO and DE, but did not experiment with these other variations. 

\subsubsection{Results}

Table~\ref{tab:fewshot} shows our results. HSO with a single gradient step leads to consistent improvements in accuracy across prompt sizes, and larger improvement with more prompt examples. The exceptions are AGNews with 2 and 4 example prompts, for which there is a slight decrease in accuracy. DE has similar performance to the baseline on SST-2, and degrades significantly on AGNews.

A longer prompt means both more examples to learn from and more gradient steps, so to disentangle the effect, we also tried two gradient steps per window (last column). 
This yields further improvement in 7 out of 8 cases. 
Surprisingly, for the cases where one gradient step was harmful, a second gradient step increases accuracy 
rather than causing further degradation.
Also, a second gradient step generally causes a larger increase in accuracy for shorter prompts (e.g., for SST-2, 
two steps with two examples 
beats one step with four examples). 

\subsubsection{Compute costs for HSO and DE}\label{hso_vs_de}
As we noted earlier, DE is not intended to be applied to a very small amount of text, so this is not an apples-to-apples comparison of methods, but can still help emphasize the differences between the two. In this setting, DE uses a much smaller amount of data (less than a single full GPT-2 window) to make updates to the entire transformer's weights. As such, it is not surprising it does not improve greatly over the baseline.

In terms of memory, the parameters and Adam moment estimates for DE of GPT-2-XL require more than 18GB in total. As the parameters are updated separately for each example, batching multiplies this overhead by the batch size, making DE infeasible for use on prompts coming from different distributions. HSO's extra overhead is the moment estimates for the hidden states, which cost \textasciitilde 1.2MB per token of input, for a total of \textasciitilde 1.3GB on a maximum size input. Furthermore, DE requires storing an additional copy of the model parameters, as they must be reset after each example. To avoid storing this extra copy on the GPU, we transferred it from RAM to GPU memory each time.

While the primary performance advantage over DE is reduced overhead and batching, we examine runtimes for each method in Table~\ref{tab:speed}. We additionally benchmark the 345M parameter GPT-2 for a speed comparison without the extra parameter transfer to the GPU. It is important to note that taking a single step per example instead of once per $k$ tokens would be much faster than either method, as both DE and HSO require $\lceil \frac{N}{k}\rceil$ backward passes for a length $N$ input.

\begin{table}[th]
\centering
\begin{tabular}{lccc}
     Method               & $n$ & \multicolumn{2}{c}{GPT-2 parameters}\\
     \cmidrule{3-4}
                          &     & 345M & 1558M\\
     \midrule
     \multirow{2}{*}{DE}  & 2   & 1.1  & 11.7\\
                          & 8   & 3.3  & 30.6\\
     \midrule
     \multirow{2}{*}{HSO} & 2   & 0.4  & 2.2\\ 
                          & 8   & 1.0  & 6.6\\
\end{tabular}
\caption{Seconds per example for few-shot evaluation using HSO and DE on SST-2. Because DE with GPT-2-XL requires copying the parameters from RAM to GPU memory every step, we also include speeds for GPT-2-medium which does not have that additional overhead.}\label{tab:speed}
\end{table}

\section{Conclusion}
In this chapter we presented a method that optimizes transformer language model hidden states, which improves LM perplexity and prompt-based few-shot classification, without additional parameters or data.
Future work will explore improving the cost of HSO by further investigation into updating only a subset of hidden weights, and approximation of the exact gradient update. Other directions we will explore are its application to conditional generation by improving the representation of the context, and its interaction with other methods for improving prompt-based few-shot classification.

\chapter{RUM-SUNDAE: Converting Masked Language Models into Non-Autoregressive Encoder-Decoders}\label{chapter:mt}

\section{Introduction}
\label{sec:intro}
In this chapter, we propose a method for using MLMs to bootstrap non-autoregressive (non-AR) generation.
This opens up new ways of using the large number of pretrained MLMs available, which previously could not be effectively be used for non-AR conditional generation.

The standard technique used for machine translation (MT) is autoregressive (AR) generation, meaning that each token is produced based on the preceding tokens.
Producing $N$ tokens this way requires $N$ forward passes, which is problematic for latency-sensitive applications.

In contrast to AR decoding, non-AR decoding methods generate text in more flexible orders, typically with all tokens being generated simultaneously.
There are many proposed techniques for non-AR generation, among which SUNDAE~\citep{sundae} is the leading MT method.
Due to the ability to decode tokens in parallel, \citet{sundae} reported a speedup of between $1.4\times$ and $4.7\times$ over an AR baseline.

\begin{comment}In contrast to AR generation, non-AR generation methods have been proposed to generate text in orders other than left-to-right, often with all tokens generated simultaneously. Non-AR methods promise to reduce latency by making far fewer than one forward pass per token to generate the whole sequence. %(although each forward pass can also be more expensive).
While there are a variety of different proposed methods for non-AR generation, SUNDAE~\citep{sundae} is the leading one among them, which reported a speedup of between $1.4\times$ and $4.7\times$, compared with a non-AR generative model.%, depending on the settings.\footnote{These speedups were for greedy decoding, whereas MT typically uses beam search. SUNDAE's model rollouts are embarrassingly parallelizable across hypotheses, while beam search is not, so it may be possible to realize faster speedups.}
% \textcolor{red}{(Need more explanation of how SUNDAE works)}
SUNDAE uses the encoder hidden state to guide the decoder to generate all the tokens in the target sequence simultaneously.
\end{comment}

While it is typical to use pretrained models to initialize MT training, this has mostly been explored in the AR case.
For instance, mBART~\citep{mbart} is a multilingual pretrained AR encoder-decoder, which was shown to improve translation results on low-to-medium resource language pairs.
Such models cannot be applied to non-AR MT since pretrained AR models employ causal masking in the decoder, which renders them incompatible with non-AR decoder architectures.

\begin{comment}
Pretrained language models have shown to be beneficial when used as an initialization point for AR models in generative NLP tasks such as MT (especially on lower-resource language pairs). For instance, mBART~\citep{mbart} is a multilingual pretrained AR (encoder-decoder) model, which has proved to improve translation performance on low-to-medium resource language pairs. Nevertheless, using pretrained language models for non-AR models is under-explored, as most non-AR methods need a decoder that does not have causal masking, implying the direct application of pretrained models to non-AR decoders inappropriate.
\end{comment}

% While it is typical to use pretrained models for NLP tasks such as MT (especially on lower-resource language pairs), finetuning of pretrained models for non-AR generation is under-explored.
% For MT, mBART~\citep{mbart} is a multilingual pretrained (AR-)encoder-decoder, which was shown to improve translation results on low-to-medium resource language pairs.

% However, most non-AR methods need a decoder that does not have causal masking, meaning that pretrained models with autoregressive decoders are not appropriate.

Another type of pretrained model, the masked language model (MLM), is trained without causal masking, so it is more similar to non-AR decoders.
Furthermore, the MLM objective and SUNDAE objective are extremely similar (See Appendix~\ref{sec:background} for an overview and comparison), suggesting that MLMs might be leveraged for SUNDAE generation.
The issue with MLMs, however, is that they are trained in an unconditional setting, meaning they lack the cross-attention layers which are required for an encoder-decoder architecture.

In this work, we investigate how to best apply MLMs to initialize models for non-AR training, and in particular how to add cross-attention to transformers trained without it.
As initializing MT \emph{encoders} from MLMs is already known to be effective, we initialize \emph{both} the encoder and decoder from a single MLM, a method which we refer to as RUM (ReUsing MLMs) initialization.
Using the XLM-R~\citep{xlmr} MLM for RUM initialization of a SUNDAE encoder-decoder, we demonstrate improvements of 2.3 BLEU and 7.5 BLEU  on the WMT'14 De-En~\citep{wmt14} and WMT'16 Ro-En~\citep{wmt16} datasets compared to training with SUNDAE from scratch.
On the CodeXGLUE dataset \cite{codexglue} we show 35.8 and 27.1 CodeBLEU improvements on Java-C\# and C\#-Java respectively.

\begin{comment}Masked language models (MLMs) are pretrained with full bidirectional attention but usually are trained as unconditional models, meaning they can be viewed as encoder-only models instead of as an encoder-decoder structure frequently used in AR generative models. A deep look into the SUNDAE reveals the similarity of its objective to the MLM objective, both of which are denoising objectives with token-level corruption. This observation motivates us to use the pretrained MLMs to initialize the SUNDAE-like non-AR generative models, to speed up the training% compared to training from scratch 
and to improve the performance.

In this work, we extend the SUNDAE non-AR model to RUM-SUNDAE (ReUsing Mlm for Step-Unrolled Denoising AutoEncoder), investigating how to best apply pretrained MLMs as the initialization for non-AR training, and in particular how to add cross-attention appropriately. We show by using the XLM-R~\citep{xlmr} model to initialize both the encoder and decoder of a seq2seq transformer, RUN-SUNDAE achieves improvements of 2.3 BLEU and 7.5 BLEU on the WMT'14 De-En~\citep{wmt14} and WMT'16 Ro-En~\citep{wmt16} datasets respectively compared to the SOTA non-AR model SUNDAE. In addition, on the CodeXGLUE dataset \cite{codexglue} we show 35.8 and 27.1 CodeBLEU improvements on Java-C\# and C\#-Java respectively. 
\end{comment}

\begin{figure*}[!htb]
    \centering
    \includegraphics[width=0.8\textwidth]
    {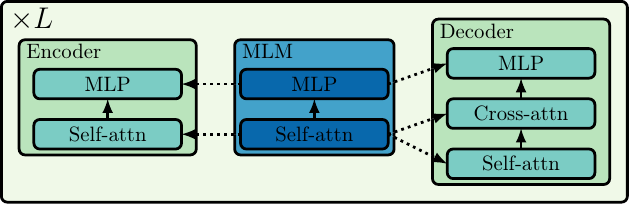}
    \caption{RUM initialization of encoder and decoder layer blocks from an MLM layer block. Dotted arrows denote weight copying.}\label{fig:init}
\end{figure*}

\section{Related Work}
\label{sec:relatedwork}
In this work we focus on SUNDAE, but RUM initialization should be useful for other non-AR generation methods as well. See \citet{xiao_survey_2022} for an overview.

The most salient related works are various attempts to generate from pretrained MLMs.
\citet{wang2019bert} and \citet{goyalexposing} both give methods for generating from MLMs without further tuning. \citet{su_non-autoregressive_2021} fit a CRF on top of a MLM, but do not update the MLM weights, and simply pass the source sequence as input rather than using an encoder-decoder architecture.

DeltaLM~\citep{ma_deltalm_2021} uses an MLM to initialize of an MT decoder as well, but for AR MT.
Their method is to use every other self-attention layer as a cross-attention layer.
As such, the second layer would receive the encoder hidden states as inputs rather than the outputs of the first layer, and so on.
This completely changes the function being computed by the MLM, so it is likely that the decoder initialization is not actually improving performance.
RUM, on the other hand, ensures that the decoder's computation is initially identical to the MLM's, so that the information acquired during pretraining can be used.

\section{Method: RUM-SUNDAE}
\label{sec:method}
% \subsection{Constrasting SUNDAE and MLM training}
The primary contribution of this work is identifying an effective means of finetuning pretrained MLMs into non-AR encoder-decoder models.
In this section we discuss our encoder-decoder initialization method, RUM, which unlocks the use of MLMs for SUNDAE MT training.

% While for each MLM training instance, a fixed fraction of tokens are replaced with [MASK], SUNDAE replaces a variable number of tokens with noise tokens uniformly sampled from the vocabulary.
% The more critical difference is that MLM training only uses one forward pass per instance, but SUNDAE training makes a second forward pass using a sample from the logits which were output by the first pass\footnote{Further repetitions of this process may be made, but \citet{sundae} showed limited improvement after two.}.
% \citet{sundae} showed that this ``unrolled denoising'' is essential for satisfactory performance. \textcolor{red}{(need to refine this part: why do we need this contrast?)}

\begin{comment}\begin{figure*}[h]
    \centering
    \includegraphics[width=0.75\textwidth]{figures/mlm_sundae.png}
    \caption{To finetune a pretrained MLM for non-autoregressive generation, we initialize both the encoder and decoder of a non-AR seq2seq model with the models' weights, add cross-attention layers to the decoder, and finetune the whole model with SUNDAE.}
    \label{fig:my_label}
\end{figure*}
\end{comment}

% \subsection{SUNDAE}

% In this section we discuss the specific initialization decisions which make doing so possible, as the most naive approach leads to failure.

\subsection{Adding modified cross-attention}
As an encoder-only pretrained model does not initially have the ability to attend to conditioning information, we cannot just directly use a pretrained MLM as the decoder.
To address this, we add a cross-attention layer after each self-attention layer, similar to the architecture from the original Transformer~\citep{vaswani2017attention}.

Another possible method would be to prepend the encoder hidden states to the decoder hidden states with an appropriate attention mask, but due to the quadratic cost of attention, this is a more expensive option.
Furthermore, since the MLMs we are experiment with use absolute position embeddings, this would cause a mismatch between the pretraining and finetuning inputs.
For these reasons, we chose to go with the option of adding the additional cross-attention layers.

First we define some notation: Let $S$ be the source sequence length, $T$ be the target sequence length, and $d_{\text{model}}$ be the model embedding dimension.
Let $h_i \in \mathbb{R}^{T \times d_{\text{model}}}$ be the output of the $i$-th decoder layer, and $h_\text{enc}\in\mathbb{R}^{S \times d_{\text{model}}}$ be the encoder output.
Finally, let $\text{Attn}(x, y)$ be the standard transformer attention using $x$ to compute the query vectors and $y$ to compute the key/value vectors (so $\text{Attn}(x, x)$ is just self-attention).

Using this notation, the function computed by layer $i$ in the MLM is:
\begin{align*}
    h_{\text{self}} &= \text{LayerNorm}\left(h_{i-1} + \text{Attn}(h_{i-1}, h_{i-1})\right)\\
    h_{i} &= \text{LayerNorm}\left(h_{\text{self}}+\text{MLP}(h_{\text{self}})\right)
\end{align*}
The simplest way to insert a new cross-attention layer into the above would be to just use the standard decoder architecture:
\begin{align*}
    h_{\text{self}} &= \text{LayerNorm}\left(h_{i-1} + \text{Attn}(h_{i-1}, h_{i-1})\right)\\
    h_{\text{xattn}} &= \text{LayerNorm}\left(h_{\text{self}} + \text{Attn}(h_{\text{self}}, h_{enc})\right)\\
    h_{i} &= \text{LayerNorm}\left(h_{\text{xattn}} + \text{MLP}(h_{\text{xattn}})\right)
\end{align*}
Recall that one of our motivations is that the MLM objective and SUNDAE objective are very similar, so we wish to leverage the strength of a pretrained MLM.
If we add cross-attention as above, it will warp the function which is computed by the decoder even at initialization time.
To address that concern, we would like to make it such that the cross-attention layers initially compute an identity function, so that at initialization the decoder output is the same as the unmodified MLM.
Due to the placement of the second LayerNorm, it is impossible to accomplish that via weight initialization alone, so we modify the cross-attention layer as follows:
\begin{align*}
    h_{\text{self}} &= \text{LayerNorm}\left(h_{i-1} + \text{Attn}(h_{i-1}, h_{i-1})\right)\\
    h_{\text{xattn}} &= h_{\text{self}} + w\times\text{LayerNorm}\left(\text{Attn}(h_{\text{self}}, h_{enc})\right)\\
    h_{i} &= \text{LayerNorm}\left(h_{\text{xattn}} + \text{MLP}(h_{\text{xattn}})\right)
\end{align*}
where $w$ is a scalar weight parameter which is initially $0$.\footnote{One could also initialize the second $\text{LayerNorm}$'s output weight to be $0$ but we chose this method to not complicate weight initialization further.}
This means that at initialization time, $h_\text{xattn} = h_\text{self}$, so the decoder's initial predictions will be identical to the original MLM's.

\subsection{Cross-attention copy initialization}
In our pilot experiments, we found that randomly initializing the cross-attention led to failed SUNDAE training.
Investigating the weight distribution of the pretrained MLMs showed that the parameter mean and variance differed greatly between layers, including between different weight matrices in the same block.
The simplest way for us to initialize the newly added layers to have those same statistics was to just copy the weights from the pretrained layers, as depicted in Figure~\ref{fig:init}.
This modification led to training converging successfully.

% The simplest way for us to initialize the newly added layers to have those same statistics was simply to copy the weights from the pretrained layers.
% This modification allowed training to converge successfully.

\section{Experiments}
\label{sec:exps}
In this section we demonstrate the effectiveness of RUM for machine translation and code-to-code translation.
We use SUNDAE to train non-AR encoder-decoder models both from scratch, and initialized with RUM (referring to the combination 
 as RUM-SUNDAE).\footnote{As \citet{sundae} did not provide a code release, we use our own reimplementation. Our implementation achieves comparable results on WMT'14 De-En, but has lower performance for En-De, which we discuss later in this section.}
% \textcolor{red}{(need to refine the following)}
% We use SUNDAE to train non-AR encoder-decoder models both from scratch, and initialized from pretrained MLMs as described in Section~\ref{sec:method}.
% We achieve comparable results on WMT'14 De-En, but find lower performance for En-De, which we discuss later in this section.

We find that our RUM-SUNDAE models consistently outperform models trained from scratch.
This effect becomes more pronounced as the amount of training data is reduced, similar to the results reported by~\citet{mbart} for the AR case.

\subsection{Models}
Our baseline models (i.e. encoder-decoders trained from scratch with SUNDAE) use the hyperparameters of \citet{sundae}, leading to models with a total of 61M non-embedding parameters.
For MT, we initialize our encoders/decoders from the XLM-R-base model~\citep{xlmr}, a RoBERTa~\citep{roberta} model which was pretrained on multilingual data.
The XLM-R-base initialized encoder-decoder has a total of 392M non-embedding parameters.
For code-to-code translation, we instead use CodeBERT~\citep{codebert}, leading to an encoder-decoder model with a total of 238M non-embedding parameters.

We also include a few AR models for reference: an AR Transformer~\cite{vaswani2017attention} for MT, and PBSMT~\cite{koehn-etal-2003-statistical}, a Transformer and CodeBert~\cite{feng-etal-2020-codebert} for code translation.
% There is one slight difference between our implementation and that of \citet{sundae}, which is discussed in Appendix~\ref{sec:length} along with experimental results on the impact of length prediction.
% Hyperparameters are discussed in Appendix~\ref{sec:hparams}.

\subsection{Datasets}
We use WMT'14 De-En~\citep{wmt14} and WMT'16 Ro-En~\citep{wmt16} for our MT experiments, computing BLEU scores with sacreBLEU~\citep{sacrebleu}.
For code-to-code translation, we use the Java-C\# translation dataset from CodeXGLUE~\citep{codexglue}. We use CodeBLEU~\citep{codebleu} metric for evaluation, which includes matching of the output's AST and semantics in addition to the BLEU score.

\subsection{Results}
\subsubsection{Machine translation}\label{sec:mt_results}
\begin{table*}[!htb]
    \centering
    \small
    \begin{tabular}{clcccc}
          & Model &  De-En & En-De & Ro-En  & En-Ro\\
         \midrule[0.8pt]
         % \cline{1-6}
         \multirow{2}*{AR Models}
         & Transformer Base~\citep{vaswani2017attention} & 31.8 & 27.3 & - & -\\
         & Transformer Base (reimplementation) & 31.27 & 27.48 & 34.05 & 33.70\\
         \midrule[0.8pt]
         \multirow{2}*{Baseline}
         & SUNDAE~\citep{sundae} & 30.8 & 26.3 & - & -\\
         & SUNDAE (reimplementation) & 28.9 & 22.7 & 26.3 & 20.1\\
         &CMLM-Raw ~\citep{ghazvininejad_mask-predict_2019} & - & 24.6 & - & \textbf{32.9}\\
         \midrule[0.2pt]
         \multirow{1}*{Ours}
         & RUM-SUNDAE & \textbf{31.2} & 23.0 & \textbf{33.8} & 24.6\\
         \midrule[0.2pt]
         \multirow{3}*{AR-distilled non-AR}
         & SUNDAE* & 32.3 & 28.5 & - & -\\
         & CMLM~\citep{ghazvininejad_mask-predict_2019}* & 30.5 & 27.0 & 33.3 & 33.1\\
         & ENGINE~\citep{tu2020engine}* & 32.0 & - & 33.2 & -\\
    \end{tabular}
    \caption{Performance comparison of RUM-SUNDAE, the baselines and the AR models on the MT task test set.}
    \label{tab:mt_bleu}
\end{table*}
\begin{comment}
We show our main machine translation results in Table~\ref{tab:mt_bleu}.
On WMT'14 De-En, RUM-SUNDAE initialized from XLM-R yields an improvement of 2.3 BLEU over SUNDAE (and 0.4 BLEU over the reported SUNDAE result), and an improvement by 1.1 BLEU over SUNDAE on WMT'14 En-De.
% but both our baseline and the finetuned model are behind~\citet{sundae}.
% It has been shown that in the autoregressive case, multilingual pretraining can be neutral or even detrimental for MT of high resource language pairs~\citep{mbart}, so this modest gain is unsurprising.
Although it has shown that in AR cases, multilingual pretraining can be neutral or even detrimental for MT of high resource language pairs~\citep{mbart}, but we successfully leveraged the multilingual pretrained model in MT with the non-AR generative models, which shed light on the using multilingual pretrained model in MT tasks. On WMT'16 Ro-En, RUM-SUNDAE initialized with XLM-R has gained a 5.7 BLEU score over the SUNDAE model, and a 4.5 BLEU score over SUNDAE on the En-Ro direction. Comparing with the AR models, we are also closing the gaps, especially on the X-EN direction.
\end{comment}
Table~\ref{tab:mt_bleu} shows our main machine translation results.
On WMT'14 De-En, RUM yields an improvement of 2.3 BLEU over the baseline of tranining from scratch.
For WMT'14 En-De, we see an improvement over the baseline by 1.1 BLEU.

\citet{mbart} showed that multilingual pretraining can be neutral or even detrimental for AR-MT of high resource language pairs~\citep{mbart}, so the relatively modest size of these improvements is to be expected.
On the medium resource WMT'16 Ro-En dataset, on the other hand RUM leads to lifts 5.7 and 4.5 BLEU lifts for the Ro-En and En-Ro directions respectively.

As our reimplemented baseline on WMT'14 En-De is over 3 BLEU behind that of \citet{sundae}, and our En-Ro results are over 6 BLEU behind our Ro-En scores (which is not typical), it is clear there was some issue in training for the X->En pairs.
Rather than exclude these results, we have chosen to leave them in for completeness.
Despite being lower in absolute BLEU than CMLM on WMT'16 En-Ro, we still find an improvement over the baseline for both translation directions.

% The results on WMT'16 Ro-En are much more dramatic, with XLM-R initialization gaining of 5.7 BLEU over the the randomly initialized baseline, and a 4.5 BLEU over the baseline for the En-Ro direction.

% As our reimplemented baseline on WMT'14 En-De is over 3 BLEU behind that of \citet{sundae}, as well as our En-Ro results being over 6 BLEU behind our Ro-En scores (which is not typical), we believe there is some issue in our handling of the X->En directions which is suboptimal.
% However, rather than exclude these results from the paper, we have chosen to leave them in for completeness.
% Despite being lower in absolute BLEU than CMLM, we still find an improvement in both cases over our reimplemented SUNDAE baseline.
% As Table~\ref{tab:mt_bleu} also shows, distillation from AR models has always improved over training non-AR models from scratch.
% As our finetuning method does not conflict with use of other finetuning methods or distillation, we believe future work should investigate the additive benefit of combining them.

\subsubsection{Code translation}
\begin{table*}[!t]
    \centering
    \small
    \begin{tabular}{clcc}
         & Model & Java-C\# & C\#-Java\\
         \midrule[0.8pt]
         \multirow{2}*{Non-AR Models}
         & SUNDAE (reimplementation) & 35.0 & 40.1 \\ % Val Java-c# 37.9
         & RUM-SUNDAE & \textbf{70.8} & \textbf{68.2}\\ % val Java-c% 76.7
         \midrule[0.8pt]
         \multirow{3}*{AR Models}
         & PBSMT~\cite{koehn-etal-2003-statistical} & 42.71 & 43.48 \\
         & Transformer~\cite{vaswani2017attention} & 63.74 & 61.59\\
         & CodeBERT~\cite{feng-etal-2020-codebert} & \textbf{85.1} & \textbf{79.4}
    \end{tabular}
    \caption{Comparison of CodeBLEU score when initializing SUNDAE training from CodeBERT vs from scratch on the CodeXGLUE Java-C\# test set}
    \label{tab:codebleu}
\end{table*}
Table~\ref{tab:codebleu} shows the performance of RUM-SUNDAE on the CodeXGLUE Java-C\# task.
RUM initialization from the CodeBERT model leads to improvement by 35.8 and 28.2 CodeBLEU on Java-C\# and C\#-Java respectively.
As this dataset only has 10K examples, it seems that there is not enough data to train a SUNDAE model from scratch, so using a pretrained model is not just beneficial, but necessary.
This success suggests that RUM-SUNDAE models offer a good way of deploying non-AR MT in domains where not enough labeled-data is available.

\subsubsection{Ablation Study}
\begin{table}[!htb]
    \centering
    \begin{tabular}{lc}
          Design & Val BLEU\\
         \midrule
         RUM-SUNDAE & \textbf{28.0}\\
         XLM-R-base init., encoder only & 25.6\\
         XLM-R-base random init. & 15.3\\
         Random init cross-attention& diverged\\
         \cite{vaswani2017attention} cross-attention& 27.7
    \end{tabular}
    \caption{Comparison of some modifications to our most successful model using the WMT'14 DE-EN validation set with oracle length.}
    \label{tab:init_ablations}
\end{table}
One obvious question is whether the larger size of the MLMs is the source of improvement, rather than the MLM pretraining.
To investigate that possibility, we trained an encoder-decoder with the same architectural hyperparameters as the pretrained MLMs, but with random initialization.
The result is reported in Table~\ref{tab:init_ablations} as ``XLM-R-base random init.'', and shows a drop of over 10 BLEU compared to RUM-SUNDAE.
This implies that the information gained in pretraining is indeed important.

Another question is whether all only the encoder is leveraging the MLM, obviating the need to carefully initialize the decoder.
To rule that scenario out, we trained a model with the same architecture as RUM-SUNDAE, but with a randomly initialized decoder.
As shown in Table~\ref{tab:init_ablations} (``XLM-R-base init. encoder only''), this led to a degradation of 2 BLEU.
This, in combination with the previous ablation, demonstrates that RUM indeed allows for the use of an MLM as an MT decoder.

The remaining two results in Table~\ref{tab:init_ablations} demonstrate that leaving out the cross-attention modifications described in Section~\ref{sec:method} degrades performance, verifying the effectiveness of RUM initialization.
Randomly initializing the cross attention leads to training divergence, so our copy-initialization trick is extremely beneficial.

\section{Conclusion and Future Work}
\label{sec:concl}
In this chapter, we proposed the use of pretrained MLMs for initialization of both towers of non-AR encoder-decoder MT models.
Our experiments on natural language and code translation demonstrated that RUM initialization allows the decoder to successfully make use of the information contained in pretrained MLMs, specifically XLM-R and CodeBERT.

This method opens up a new range of applications for all pretrained MLMs, contributing to our overall goal of increasing the utility of already extant NLP models.
Future research will focus on applying our method to other generative tasks with task specific pretrained models, and on exploration of performance improvements such as weight-tying between the encoder and the decoder.

\newpage
\section{Additional information}
\subsection{The Effect of Length Prediction}
\label{sec:length}
In this section we discuss how our length prediction implementation differs from \citet{sundae}, and also evaluate the difference in model performance when using predicted vs reference lengths.

The difference between our implementation and the method reported in \citet{sundae} is that we trained out length predictors separately from the training of the seq2seq model (for all models including baselines).
We expect that this should be strictly better because \citet{sundae} did not use the gradient of the length prediction loss to update the encoder weights.
As such, training a length predictor after the fact means that the length prediction module can learn to handle inputs from a a fixed encoder rather than one that is being updated over time.
Further, we can do early stopping for the length predictor only rather than jointly needing to stop training the length predictor and the MT model.

The possible drawbacks of this approach are that we had to use the reference sentence length rather than predicted lengths for early stopping on the MT model, and also added training complexity due to needing to do two separate training runs.

Table~\ref{tab:mt_oracle} shows the same results shown in Section~\ref{sec:exps}, with additional evaluations using the same translation model, but producing a number of tokens equal to the length of the reference translation rather than the predicted number.
This led to an increase in BLEU ranging from 0.2 to 2.2, depending on the setting, giving an estimate of how much extra performance may be attained just by improving the quality of length prediction.

The analogous results for code translation are shown in table~\ref{tab:code_oracle}.
In this case the gap between using the reference and predicted length was much higher than for natural language translation.
However, as CodeBLEU takes semantics as well as n-gram matches into account, it is perhaps unsurprising that this is the case.
Producing one extra incorrect token reduces BLEU by a fixed amount, but for programs may cause them to by syntactically incorrect or otherwise entirely change the semantics.

\begin{table*}[!t]
    \centering
    \small
    \begin{tabular}{lcccc}
          Model &  WMT'14 De-En & WMT'14 En-De & WMT'16 Ro-En  & WMT'16 En-Ro\\
          \midrule[0.8pt]
         SUNDAE~\citep{sundae} & 30.8 & \textbf{26.3} & - & -\\
         SUNDAE (reimplementation) & 28.9 & 22.7 & 26.3 & 20.1\\
         SUNDAE (reimpl.) + Oracle length & 30.5 & 23.8 & 26.4 & 20.3\\
         \midrule[0.2pt]
         RUM-SUNDAE  & \textbf{31.2} & \textbf{23} & \textbf{33.8} & \textbf{24.6}\\
         RUM-SUNDAE + Oracle length & \textbf{32.7} & \textbf{25.2} & \textbf{34.6} & \textbf{25.3}\\
    \end{tabular}
    \caption{Machine translation results (test set BLEU). Oracle length means the reference text length was used instead of a predicted length.}
    \label{tab:mt_oracle}
\end{table*}
\begin{table*}[tb]
    \centering
    \begin{tabular}{lcc}
         Model & Java-C\# & C\#-Java\\
         \midrule[0.8pt]
         SUNDAE (our impl.) & 35.0 & 40.1 \\ % Val Java-c# 37.9
         SUNDAE (our impl., oracle length) & 37.3 & 43.7\\ % Val Java-c# 37.9
         \midrule[0.2pt]
         RUM-SUNDAE & \textbf{70.8} & \textbf{68.2}\\ % val Java-c% 76.7
         RUM-SUNDAE (oracle length) & 80.0 & 76.2\\ % val Java-c% 76.7
    \end{tabular}
    \caption{CodeXGLUE Java-C\# translation results. Oracle length means the length of the reference was used instead of the predicted length}
    \label{tab:code_oracle}
\end{table*}

\subsection{Contrasting SUNDAE and MLM training}
\label{sec:background}
% \begin{comment}
In this section we will give a brief overview of SUNDAE~\citep{sundae} training and decoding procedures, which we use for our experiments in this work.
Conceptually, it is very similar to the MLM pretraining task, but has important changes which we will explain here.
We will only describe the method as it applies to encoder-decoder transformers, not the unconditional decoder-only case.

The first difference from MLM is that the noise distribution is replaced with uniform random token corruption.
For every training source sequence, a clean target sentence is tokenized, then a random proportion is selected uniformly in $[0, 1]$.
That proportion of tokens are replaced with a uniformly random token from the vocabulary.
The model then processes the noised sentence and attends to the encoder's embedding of the source sentence to produce the logits, $\ell \in\mathbb{R}^{T \times V}$ (where $T$ is sequence length and $V$ is the vocabulary size).
The cross-entropy loss is then used to score those predictions and compute gradients, using all positions not just the corrupted ones.

The second and larger difference is that the logits computed above are then sampled from independently at each position to generate a new, partially denoised, sentence.
This sample is then used as input to the decoder, exactly as the original noised sentence was, producing a second cross-entropy loss term.
The sum of these two loss terms is then used to update the model weights.

For decoding from a model trained with SUNDAE, \citet{sundae} proposed multiple methods, but we will only describe the ``low temperature sampling'' method.
At test time, there is only a source sequence given, so the output is initialized randomly.
The number of tokens to use is predicted by a length prediction module which takes the encoder hidden states as input.
Similarly to at training time, logits are computed for each decoder position, and sampled from after dividing by the temperature.
This process is then repeated a fixed number of times.
Because the model is exposed to a variety of noise levels at training time (everything from fully corrupted sequences to completely unchanged sequences), it is able to continue improving the partially denoised sequences until reaching a high quality translation.
% \end{comment}
% The similarity between the SUNDAE and MLM objectives motivates us to apply non-AR training to pretrained MLMs.
% Please see \citet{sundae} for a full description of SUNDAE.

While for each MLM training instance, a fixed fraction of tokens are replaced with [MASK], SUNDAE replaces a variable number of tokens with noise tokens uniformly sampled from the vocabulary.
The more critical difference is that MLM training only uses one forward pass per instance, but SUNDAE training makes a second forward pass using a sample from the logits which were output by the first pass\footnote{Further repetitions of this process may be made, but \citet{sundae} showed limited improvement after two.}.
\citet{sundae} showed that this ``unrolled denoising'' is essential for satisfactory performance, which suggests a possible reason why direct decoding from MLMs has been mostly unsuccessful in the past.

% Given the overall similarity of these two training techniques, there is good reason to suspect that the information captured in MLM pretraining will also lead to strong performance on the SUNDAE objective.
\chapter[Degenerate NLG outputs might not be due to model error]{Degenerate NLG outputs might not be due to model error\footnote{A significantly updated version of the work in this and the following two chapters appeared as \citet{modesacl}}}\label{chapter:modes_theory}
While the previous chapters have focused on transformer language models, in this chapter we will be discussing issues which affect all language models, in the sense of models which represent distributions over text.
We give a mix of new ideas, and ones which have been discussed in prior work, with the goal of encouraging the NLP community to think more closely about the issue of degenerate NLG outputs such as those identified by \citet{catgotyourtongue}.
We will start with a motivating analogy.

\subsubsection{Predicting the future with a TV}
Imagine you were watching a TV, and you wanted to predict what the next frame would be, based on the current frame.\footnote{Bear with us, we'll get back to NLP momentarily.}
If you had infinite training data, time, and compute, you'd eventually be able to do as well as possible at this task.

``As well as possible'' might not be perfect.
Realistically, there will be \emph{intrinsic uncertainty} in distribution of next frames.
Intrinsic uncertainty occurs when there are multiple possible outputs for a given input.
If the input is a picture of a baseball suspended in mid air, you might not be able to perfectly guess the next frame, since you don't know which way the ball is flying.
Even worse, it's possible that at any time, the feed will cut from one scene to another.
You'll have to train a model which represents this uncertainty.
That is, given a current frame $x$, your model will tell you $\prob(y | x)$ for any possible next frame $y$.

This model by itself isn't enough, you want to predict the next frame, not just know how likely each frame is.
The process of mapping from an entire distribution to a single prediction is called \emph{decoding}, and there are many different ways to do it.
There are at least two obvious strategies, which we'll refer to as \emph{sampling} and \emph{search}, as they're stand-ins for more general classes of methods.
The sampling strategy is to draw a single sample from the probability distribution your model learned.
The search strategy is to predict the single frame which the probability distribution thinks is most likely.
This particular search strategy is called \emph{maximum a posteriori} (MAP) decoding.

To make your job a bit harder, imagine the TV has a damaged cable which adds noise to the incoming data.\footnote{To be specific let's imagine that each frame is corrupted by additive spherical Gaussian noise in pixel space.}
If you trained your model on these noisy inputs, there'd now be a noticeable difference between sampling and search.
Sampling would usually give you an output which has the average amount of noise, but search would give you a noise-free image.\footnote{This might not be true if the distribution of images is such that noisy versions of distinct clean images overlap sufficiently, but for a low enough noise scale it will be.}
Your model knows enough about the data distribution to ignore the noise, and give you a clean frame.

You really want to know what happens in the next frame of the TV program, so you decide to go with search instead of sampling.
Now let's make things even worse, and suppose that on top of the additive noise, 1 in 1,000 frames gets dropped altogether.
When this happens, your TV just shows a black screen, with no image or even the noise.

Now we have a real problem.
If the distribution of noise-free images consists of more than 1,000 images which are all equally likely, any individual \emph{real} next frame will be lower probability than the blank screen!
Realistically, there are likely \emph{far} more than 1,000 similarly likely outputs, so the probability of the blank frame could be even smaller, and this problem would still occur.
When both the frame corruption \emph{and} the frame dropping noise are present, neither search nor sampling will give you a good output.
Sampling will give you noisy images, and 1 in 1,000 times it will give a blank output.
Search, on the other hand, will just give you a blank screen every time.

The problem here isn't in your model; it fits the data perfectly!
The problem is your decoding methods, which are vulnerable to the two types of noise in video stream.
Your model still knows what the  most likely non-noise frame is, and you can extract it by looking for the mostly likely frame which isn't blank, i.e.:
\begin{equation*}
    y_\text{pred} = \argmax\limits_{y} \prob \left( y \given[\big] x, \text{ ``$y$ is not blank''}\right)
\end{equation*}
This decoding method just extracts a \emph{conditional mode} instead of the unconditional mode.

\subsubsection{Relationship to NLG}
We will argue that the TV prediction game above is closely analogous to decoding from models in NLG.
In particular, we use decoding methods which are susceptible to specific types of noise in our datasets, but are surprised when those decoding methods give us low quality outputs!

This claim is primarily based on empirical and theoretical arguments which have already been made in the NLP literature.
For example, \citet{catgotyourtongue} in particular found that NMT models predict that the empty sequence is the modal translation for many inputs.
However, these findings are often assumed to be evidence of \emph{model error}.
The reason for this chapter is to try help the NLP community build intuition about this topic, as well as suggesting possible ways to handle noisy distributions.
We'll focus in particular on the ``bad mode problem'', the issue that NLG models will often assign a higher probability to degenerate outputs than good ones.

The important thing to understand about this chapter is that it is about \emph{sequence distributions}, as opposed to models trained on them.
So, if we convincingly argue that a distribution has some problem, it implies that no amount of improvements in model training can fix that issue.
The question is then whether the training data distribution itself has that problem.
We will argue that it is likely that real text corpora \emph{do} have noise modes which lead to undesired outputs.

The rest of this chapter consists of the following:
\begin{itemize}
    \item Section~\ref{sec:modestheory_notation} introduces notation for the rest of the chapter, and formalizes the issue we see with decoding methods via an ``oracle criterion.''
    \item Section~\ref{sec:modestheory_related_work} goes over past empirical findings on the bad mode problem, as well as how those findings have been interpreted.
    \item Section~\ref{sec:modestheory_examples} gives several examples of sequence distributions which display the bad mode problem.
    \item Section~\ref{sec:modestheory_realistic} argues that this phenomenon might be present in the distributions of our actual training sets as well.
    \item Section~\ref{sec:modestheory_proposal} discusses possible approaches for decoding from noisy distributions.
\end{itemize}
Chapter~\ref{chapter:modes_exact} will extend this by looking at the exact modes of several NLG models.

\section{Data distributions and the oracle criterion}\label{sec:modestheory_notation}
As we indicated in the introduction, many common decoding methods will produce undesirable results when used on distributions which contain noise.
To make this more concrete, we propose that we should reason about decoding algorithms in terms of how they behave when applied to the \emph{training distribution itself}.
Since we don't have access to the actual training data distribution, only a finite sample of it, we are only recommending this as a fruitful kind of thought experiment, not an empirical test.
In Section~\ref{sec:modestheory_examples} we'll apply this method by looking at the exact mode of a number of sequence distributions.

In the rest of this section we'll introduce some notation, and make the connection between the television example and NLG more explicit.
We'll then make the idea of decoding directly from the training distribution more concrete as the \emph{oracle criterion}.

\subsection{Notation}
\paragraph{The set of outputs} In all of our settings, we have some outputs which we want to predict, usually based on some inputs. We'll refer to the set of all possible outputs as $\seqs$. In the TV example $\seqs$ would be the set of all possible pixel assignments for our TV screen, while in the rest of the section it will be the set of all finite sequences from a vocabulary.

\paragraph{The ``clean'' subset.} The thing we really care about is outputs which are correct for a given input, which we'll call $\sperfect$.
When we need to specify that it depends on a particular input, $x$, we'll write $\sperfect(x)$.
(This generalizes to the case of unconditional generation by considering a single constant input).
In the TV example, for a previous frame $x$, $\sperfect(x)$ consists of frames which could follow $x$ in the absence of noise.
An example from NLP would be the set of valid translations for an MT task:
\begin{equation*}
    \sperfect(x) = \{y | y\text{ is a translation of }x\}
\end{equation*}
Note that this depends entirely on some task-specific notion of quality or correctness.
Consider the task of replying to Reddit comments.
If the goal is to simulate a Redditor, $\sperfect(x)$ includes all comments which might be replies to $x$.
On the other hand, if the goal was to give \emph{helpful} replies, $\sclean(x)$ would not include things like jokes, insults, etc.

\paragraph{The clean distribution.} We'll write $\dperfect$ for a distribution only supported on $\sperfect$.
This would be the distribution of next frames only including intrinsic uncertainty and no noise in the TV analogy.
In the task of summarization, for example, $\dperfect$ should be a distribution over all valid summaries of the input.

\paragraph{The training distribution.} For a distribution which potentially contains some noisy outputs, we'll write $\dtrain$. The noise could be due to $\dtrain$ being a mixture between $\dclean$ and some noise (as with the blank frames in the TV example), or due to corruption of outputs (the additive noise in the TV example).
In NLP this is the population distribution for a training set, including whatever noise is present.
We'll write $\ptrain(\cdot)$ for the probability that an event occurs under $\dtrain$, or the PMF of a random variable when the argument is a random variable.

\paragraph{Attributes.} We don't just want to view all texts as identical, so we'll need some way to categorize them.
We'll talk about \emph{attribute functions}, $A(x)$, which map an output $x$ to an set of possible values $\mathcal{A}$. For example, in our TV setting we might have $A(x)$ return the truth value of ``x is a blank frame''. One we'll make a lot of use of is length, so $A(x) = |x|$ and $\mathcal{A} = \mathbb{N}$.
Other examples include ``Is $x$ a fluent text?'', or ``Does $x$ have positive sentiment?''
These attributes are binary, so $\mathcal{A}$ would be the set $\{0, 1\}$.

\subsection{The oracle criterion}
We want to encourage the community to think more rigorously about what NLG behaviors are due to model error, and which are due to the difference between $\dtrain$ and $\dclean$.
If low-quality outputs are due to noise in the training data, we need to get better data or improve our decoding.

For analyzing decoding methods, we propose the following criterion:
\begin{mdframed}[backgroundcolor=gray!10, linewidth=0pt, innertopmargin=10pt, innerbottommargin=10pt]
    \textbf{Oracle Criterion:}\\
    An NLG decoding method should produce high-quality results if it is applied to the relevant oracle for $\dtrain$.\footnote{Different methods may need different oracles. For example, ancestral sampling needs the next token distribution $\ptrain(x_{t} | x_{<t})$. Non-autoregressive sampling methods may need $\ptrain(x_{t} | x_{<t}, x{>t})$.}
\end{mdframed}
Informally, the point is that if a decoding algorithm can't produce good outputs even when a model is trained perfectly, we should use a better method.

It may still be the case that we want to use methods which would not work well on $\dtrain$ in order to compensate for model error.
For example, nucleus sampling excludes the tail of the token distribution, which probably helps due to our models not accurately fitting that part of the text distribution.
However, we should be explicit about which of our decoding choices are made to compensate for model error, and would be necessary regardless of how well the model is trained.

\section{Prior empirical findings}\label{sec:modestheory_related_work}
In this section we'll look at past results related to the bad mode problem, as well as what the authors inferred from their findings.
Each of the snippets below is just the findings/claims from each paper which are directly applicable, not a full summary.

The repeated points are that on some (but not all) tasks NLG outputs can be high-likelihood while still being low quality.
The issue is generally attributed to some kind of model error or bias, and findings generally indicate that it gets worse on tasks where outputs are less constrained (although no one agrees on what measure of constraint to use).

\subsection{\citet{catgotyourtongue}}
\paragraph{Findings:} The authors ran exact MAP inference for a transformer MT model on the WMT \texttt{news-test2015} dataset.
They found that the empty sequence was the modal for 51.8\% of inputs, and were the first to report this kind of result.
On top of that, longer source sentences were \emph{more} likely to have the empty sequence as the modal output.

\paragraph{Interpretation:} They described these results as ``serious NMT model error,'' and ``evidence of NMT’s failure to properly model adequacy.''
More specifically, they say that ``at the root of the problem lies an inherent bias towards shorter translations.''

\subsection{\citet{inadequacyofmode}}
\paragraph{Findings:} Eikema and Aziz pointed out that \emph{samples} from NMT models don't have the same degeneracies as MAP or beam search outputs.
They found that samples were closer to the reference translations in terms of both length and n-gram statistics.
\paragraph{Interpretation}: They call the mode \emph{inadequate}. Specifically, they say: ``The mode often does not represent any significant amount of probability mass under the learnt distribution. We therefore argue that MAP decoding is not suitable as a decision rule for NMT systems.''
They also say that to criticise a model reasonably, you have to view it as an entire distribution instead of just looking at the mode.

This is a point we completely agree with them on, but our reasoning is different.
The reason they give is: ``the most likely translations under the model accumulate so little probability mass that the mode can be considered essentially arbitrary.''
We won't disagree with that, but we'll instead look at what causes bad modes to occur, and whether we can get any use out of MAP-type methods despite them.

\subsection{\citet{josifoski-likelihood-utility}}
\paragraph{Findings:} Josifoski et al. looked at the relationship between model-likelihood and utility of output across several NLG tasks and decoding methods.
They found that likelihood and utility were positively correlated for information extraction, but when using for machine translation, they were mostly negatively correlated.
For the other tasks, there were only weak correlations.

\paragraph{Interpretation:} They give a taxonomy of possible sources of disagreement between likelihood and utility:
\begin{itemize}
    \item \textbf{Training imperfections (TI)}: ``The model is trained on a different objective than the true utility, because of the finite size of the dataset and the approximation error in training''
    \item \textbf{Distribution shift (DS)}: ``When the training and testing distributions differ''
    \item \textbf{Utility drift (UD)}: ``UD occurs when the notion of utility changes, i.e. the labels for the same data points are changing.''
\end{itemize}
For example, they list NMT as a task where TI and DS are present, but UD is not.
They conclude that TI alone is present, likelihood is a strong predictor of utility, but when DS/UD are present this is not the case.

\subsection{\citet{riley2022continuum}}
\paragraph{Findings:} Riley and Chiang create a spectrum NLG tasks with varying levels of constraint by training NMT models using partially incomplete source sentences.
Training with 0\% of the source is just language modeling, using 100\% is MT, and other values give new tasks which are partially constrained.
For lower constraint levels, they say: ``beam search decoding that is too short, greedy-decoding which is too long and repetitive, and random samples that are disfluent.''
When they increase the level of constraint, length bias and repetition both improve.
\paragraph{Interpretation:} They point out that a unigram LM would have the same three problems as they found in beam search, so they say insufficient context-sensitivity might be the cause.

\subsection{\citet{nucleus_sampling}}
\paragraph{Findings:} Using beam search with GPT-2 produces repetitive outputs.
The authors also look at the per-token probability of beam-search generated and human text and find that the human text has much lower probability per token, and much more variation in probability between tokens.
\paragraph{Interpretation:} They argue that probability maximization is not appropriate for open-ended generation, as humans are not trying to output maximally probable sentences.

\subsection{\citet{zhang2021trading}}
\paragraph{Findings:} Using samples from GPT-2-774M~\citep{gpt2}, Zhang et al. observe that likelihood is correlated with crowdworker-rated quality up until a point, and negatively correlated afterward.
They also draw many samples across multiple hyperparameter settings for three sampling methods: temperature, top-k~\citep{topk}, and nucleus sampling.
When the settings are such that the decoding methods have similar output entropies, all methods perform similarly.
The exception is that nucleus sampling performs best when entropy is very low.
\paragraph{Interpretation:} The authors don't speculate about the source of the reversal in likelihood/quality correlation.

\subsection{\citet{wiher2022ondecoding}}
\paragraph{Findings:} They evaluate many decoding methods on several different NLG tasks.
The decoding methods are: greedy search, beam search, diverse beam search, pure sampling, top-k sampling, nucleus sampling, and Minimum Bayes Risk (MBR).
The tasks are MT, abstractive summarization, dialogue, story generation, and unconditional generation.
In general, beam search variants perform best for the tasks which have some constraint: MT, summarization, and dialogue.
For story generation, and unconditional generation, the sampling based methods perform best.
They find that for MT and summarization, quality and probability are positively correlated, while this relationship is flipped for story generation and unconditional generation.\footnote{This doesn't contradict other findings about high-likelihood/low-quality sequences in MT since none of the decoding methods can find super high-likelihood sequences.}
\paragraph{Interpretation:} Since the performance of decoding strategies varied so strongly between tasks, they say ``Practitioners should take great care in basing their choice of decoding strategy off of results reported for alternate tasks.''

\subsection{Summary}
One repeated finding is that likelihood-maximization based approaches tend to work less well for tasks that are less constrained.
Also, many authors indicated that the degeneracies they found are likely due to model error.
To be clear, this isn't a central point of their arguments, it's just the default assumption when we observe behavior like this.
\citet{inadequacyofmode} in particular argued that MAP was a poor decoding strategy, but their reasoning was that the mode is too small of a fraction of an MT model's output distribution.
This may also play a role in the bad mode problem, but in the next chapter we'll actually see that \emph{conditional modes} are often high quality, meaning that the low probability of the mode can't be the whole story.

\section{Distributions with bad modes}\label{sec:modestheory_examples}
Model error is definitely a plausible partial cause of the results we went over in the previous section, and we won't be saying that NLG models are somehow free of error.
However, in this section we'll look at the exact modes of some sequence distributions, and see that they often have undesirable properties.
This has two implications: First, even if we could train error-free models, we'd still be stuck with the issue; and second, if we want to know that an outcome \emph{is} due to model error, we need to know that it isn't implied by the training data distribution.\footnote{For example, \citet{inadequacyofmode} compared the \emph{average} length of samples from their MT models to the average length of reference sentences from the data. They found some model error, but much less than one would assume from the empty mode results of \citet{catgotyourtongue}.}

If $\dtrain = \dclean$, then of course model error is implied by these results. 
However, we'll also show that adding relatively small amounts of noise to the data distribution can lead to $\dtrain$ having a bad mode.
It's not just ``garbage in, garbage out'', it's  ``tiny amount of garbage in, sizeable amount of garbage out.''

\subsubsection{Typicality}
A concept which is important for reasoning about these results is that of \emph{typicality}.
\citet{dieleman2020typicality} provides an excellent overview of the topic, and \citet{typical_sampling} applied the concept to NLG to come up with the \emph{locally typical sampling} method.
Informally, a typical sample is one that isn't much more or less likely to be observed than an average sample.
More formally, we'll call the typical set for a random variable $X \sim\mathcal{D}$:
\begin{equation}
    S = \left\{
        x\; \given[\big] \;
        \lvert
        -\log\prob_\mathcal{D} \left(X = x\right)
         - H(X)
        \rvert < \epsilon
    \right\}
\end{equation}
For some small $\epsilon$.
This is slightly different from the standard formulation of the typical set, which instead normalizes by the number of tokens in the string (see e.g., \citealp{typical_sampling}).
However, in this work, we reason about distributions and models which are globally normalized over sequences of varying length.

The bad mode problem is a particular example of atypicality: the mode is not a typical sample unless the distribution is extremely low entropy.
Frustratingly, we'll also see cases where atypicality is helpful, meaning we need to take a more nuanced view of when to try to find typical sequences.

\subsubsection{The bad mode problem}
We can now easily summarize the bad mode problem, and why it's counterintuitive.
Using the notation from Section~\ref{sec:modestheory_notation}:
\begin{equation}\label{eq:noncommutative}
    \argmax\limits_{a \in \mathcal{A}} \ptrain\left(A(X) = a\right) \neq A\left(\argmax\limits_{x \in \mathcal{S}} \ptrain(X = x)\right)
\end{equation}
I.e. the most likely attribute value is not the same as the attribute value of the most likely string.
If $A(x)$ is the indicator function of $\sclean$, this is just the bad mode problem: A distribution can have an arbitrarily high probability of yielding high-quality samples, but still have a low quality mode.
When the attribute is length, $A(x) = |x|$, then Equation~\ref{eq:noncommutative} says that the mode being empty doesn't imply anything about the modal length.

The rest of this section goes over cases where the lack of equality in Equation~\ref{eq:noncommutative} occurs.
This is often undesirable, but we'll also see cases where it's beneficial.

\subsection{Warm-up: Biased coin flips}\label{sec:coins}
This example comes from ~\citet{dieleman2020typicality} (originally from \citet{yung}).
Consider flipping a biased coin, which comes up heads 60\% of the time, 100 times.
(So $\seqs = \left\{H, T\right\}^{100}$).
Let $A$ be the function that counts the number of heads, so $\mathcal{A} = \{0, 1, \dots,100\}$.
The output of $A$ with the highest probability is 60, and there is an approximately 97\% probability that $50 \le A(X) \le 70$.
However, for a sequence with $N_H$ heads, the probability of observing that sequence is $(0.6)^{N_H}(0.4)^{100 - N_H}$, so the modal sequence consists entirely of heads.

This result is counterintuitive because grouping outcomes by the number of heads is natural to us, but the probability distribution doesn't care how we choose to partition it, it just assigns a weight to each of the $2^{100}$ outcomes.
Despite the counterintuitiveness, there's nothing objectively good or bad about this fact since we don't have any task or notion of quality to appeal to.
As such, this instance of atypicality is just a benign curiosity.

\subsection{Variable length sequences}\label{sec:modestheory_length}
Text generation is a good bit more complicated than flipping coins, so here we'll describe a distribution over sequences of variable length.
Predictably, we'll show that it's possible for the \emph{modal sequence} to be the empty sequence while the \emph{modal length} is arbitrarily high.

Let $\seqs$ be the regular language defined by $(a|b)*$, i.e. all finite strings consisting of one of two characters, including the empty string.
The attribute function will be the length $A(x) = |x|$.
Consider a random variable $X\in \seqs$ such that all strings of a given length have the same probability.
That is, for a length, $\ell$, $\prob(X = x | |x| = \ell)$ is the uniform distribution.
We'll write $p_\ell$ for the marginal probability that $X$ is length $\ell$.

The modal attribute value is the most common sequence length:
\begin{equation*}
\ell^* = \argmax\limits_\ell p_\ell
\end{equation*}

On the other hand, a modal sequence (there may be many) is defined by:
\begin{equation*}\label{eq:modes_length_argmax}
x^* = \argmax\limits_{x \in \seqs} \frac{p_{|x|}}{2^{|x|}}
\end{equation*}

For simplicity consider a strictly unimodal length distribution, i.e.:
\begin{align*}
    \ell < \ell^* &\implies p_\ell < p_{\ell + 1}\\
    \ell > \ell^* &\implies p_\ell < p_{\ell - 1}\\
\end{align*}
This implies that $|x^*| \le \ell^*$, since all the sequences longer than $\ell^*$ have lower probability than $p_{\ell^*}$.

If $p_\ell$ grows slowly enough, the right hand side of Equation~\ref{eq:modes_length_argmax} will be maximized when $\ell = 0$, implying that $x* = \epsilon$, the empty string.
The specific condition for this to be the case is:
\begin{align*}\label{eq:modes_length_bound}
   \frac{p_\ell}{p_0} < 2^{\ell}
\end{align*}
So, if the probability of a given output length grows less than exponentially quickly, the mode will be empty.
By using a sufficiently slowly growing sequence, $p_\ell$, one can have the length of the modal sequence arbitrarily high, while also having the empty sequence be modal.

This distribution is still extremely unrealistic, but it shows one facet of the issue: A single bad sequence with arbitrarily low quality can still have a higher likelihood than a high-entropy set of good or typical sequences.
We haven't defined a task (so there's no concept of $\sperfect$), but for all common NLG tasks, this outcome would be an instance of bad atypicality.

\subsection{Bland word choice}\label{sec:modestheory_templates}
Outputs which are too short is just the most obvious case of bad modes, but we can also construct distributions which have the problem while having constant length.
Here we'll let $seqs$ be instantiations of the template: \texttt{<name><vb.><adj.><noun>}, e.g. ``Joe eats delicious toast.''
For the distribution over $\seqs$:
\begin{itemize}
    \item For each slot, define a single common word which has probability 10\%, and 90 rare words which each have probability 1\%
    \item Each slot is filled independently
\end{itemize}
Now consider the attribute $A(x) = \text{\# of rare words in $x$}$.
The most common attribute value is $A(x) = 4$, since the probability of all four words being rare is $0.9^4 \approx 66\%$.
On the other hand, the modal string is made of the four common words, and has probability $1/10000$.

If we're interested in quality in the sense of representativeness or creativity, it's easy to see that this is a bad mode.
\citet{typical_sampling} argue that we should use locally typical sampling to achieve similar information content to natural text.
On this example, locally typical sampling would lead to outputs which have four rare words, which is representative of a typical output.

On the other hand, the mode actually globally maximizes BLEU score against a reference randomly sampled from this distribution.
If we look at this example as analogous to the choices a translation system needs to make in producing an output, our system would get a higher BLEU score by using the mode than a sampled output.
More importantly (maximizing BLEU isn't very motivating), what kind of translation we want likely depends on what the translations are for.
If every human translator injects their own personal flair or style into translation, we probably want to reflect that in translations of fiction or other works where rhetoric is important.
On the other hand, in translating technical documentation, a modal translation is probably fine, or even preferable, to one which is more natural.

In summary, if we want creativity and representativeness, the mode for this distribution is bad.
On the other hand, if we want outputs which are as ``standard'' as possible, the atypicality of the mode can even be beneficial.

\subsection{Independent errors}\label{sec:modestheory_errors}
Now we'll look at a case which is essentially the same as the previous example, but in this case, the atypicality will be desirable rather than harmful.

Consider a distribution over texts of in which first a text of $N$ sentences is sampled from some high quality distribution, $\dclean$.
Then, define $\dtrain$ as a distribution which takes samples from $\dclean$ introduces a typo into each sentence independently with probability $p_e$.

A typical text will contain $N p_e$ typos, but provided that $p_e$ is low enough, the most likely sequence should contain zero typos!\footnote{How low $p_e$ has to be depends on how likely typos which could have come from multiple different clean texts are.}
This is mathematically the same as Section~\ref{sec:modestheory_templates}, but with ``sentences with errors'' playing the role of ``rare words.''
The difference is just that in this case we want to avoid the ``rare'' phenomenon, rather than encouraging it.

So here, we have a type of atypicality which is (essentially) unambiguously good.
Representative examples from $\dtrain$ will have errors, and we want to avoid them!
Once noise is introduced, looking at outputs purely from an ``information-per-token'' or typicality point of view runs into some problems.
It isn't clear how to distinguish ``Low-probability because useful information is being added'' and ``Low-probability because there's a misspelling'' from the model probabilities alone.

\subsection{Summary}
We've shown several examples of sequence distributions which have modes unrepresentative of a typical sample.
Importantly, a perfectly trained model on any of these distributions would have a mode with the same property.
This means that observing these kinds of behaviors in a model is insufficient to demonstrate model error, unless you can also show that $\dtrain$ does not have that property.

This atypicality can be harmful, benign, or even helpful, depending on our aim.
Estimating statistics on samples drawn from the model, as \citet{inadequacyofmode} did, is the way to actually investigate approximation error.
Returning to the ``Oracle criterion'' from Section~\ref{sec:modestheory_notation}, it's clear that even given an oracle for the distributions in this section, MAP inference would be inappropriate for most of them.

However, sampling based approaches also wouldn't be able to clean up the errors in the noisy distribution from Section~\ref{sec:modestheory_errors}.
MBR decoding (using samples as hypotheses) and typical sampling would both output text with the average number of errors\footnote{MBR with the right cost function might reduce the number of errors, but once texts get long enough that all the hypotheses also have errors, the output will have errors as well.}
If we consider a distribution which suffers from both a bad mode \emph{and} noise in typical samples (as in the TV story at the beginning), it doesn't seem like any standard methods are up to the task.

\section{Modes of real data distributions}\label{sec:modestheory_realistic}
In this section we'll argue that two common NLG tasks, machine translation and chatbot instruction-following, might suffer from noise which leads to the presence of bad modes in their training data distribution.
$\sclean(x)$ for an MT input $x$ is the same as we said in Section~\ref{sec:modestheory_notation}: the set of valid translations of $x$ into the target language.
For the chatbot case $\sperfect(x)$ should only contain sequences which are ``good'' responses to the prompt $x$.

Since the definition of ``valid translation'' is a philosophical question, we can imagine operationalizing this with something like ``a panel of 5 UN translators would unanimously agree that $y$ is a faithful translation of $x$.''
For chatbots, the definition could be ``$y$ receives the maximum human rating on all criteria'' for some set of criteria such as being helpful and harmless.

If we got our training data from $\dperfect$, a model assigning non-zero probability to a low-quality text (i.e. one not in $\sclean$) would imply that there is model error present.
So the question is whether the the gap between $\dclean$ and $\dtrain$ for our actual training datasets is low enough for this conclusion to hold in practice.

Consider a specific model of $\dtrain$, which consists of samples from $\dperfect$ but with corruption randomly applied.
To say the training distribution has a bad mode for input $x$ means that there's some $y_\text{bad} \notin \sperfect$ such that:
\begin{equation}\label{eq:bad}
    \ptrain\left(y_\text{bad} | x\right) \ge \max\limits_{y \in \sperfect} \ptrain(y | x)
\end{equation}
It's hard to say anything specific about the right hand side of Equation~\ref{eq:bad} without any assumptions on $\ptrain$.
However, if we restrict our attention to outputs which are typical among the high-quality outputs, we can give an intuitive condition under which this will occur.

Write $\stypical(x)$ for the set of outputs $y$ which are typical \emph{high-quality} output for $x$.
$y\in \stypical(x)$ means that:
\begin{equation*}\label{eq:good_typical}
    \log \ptrain(y | x\; y\in\sperfect) \approx -\mathcal{H}_{\text{train}}(y | x,\; y\in\sperfect)
\end{equation*}
where $\mathcal{H}_{\text{train}}(\cdot | \cdot)$ is conditional entropy.
All values of $y$ which have an NLL similar to the conditional entropy will also have similar NLL to each other, so we can just think about the uniform distribution over the typical set.\footnote{The only reason we need to say have an approximate equality instead of equality anyways is because we're working with discrete random variables.}

We'll get either a bad or atypical mode when there's some $y \notin \sperfect(x)$ such that:
\begin{equation*}\label{eq:modes_uniform}
    \ptrain(y | x) > \frac{1}{|\stypical(x)|}
\end{equation*}
As we discussed in Section~\ref{sec:modestheory_related_work}, some version of ``number of valid responses'' is generally thought to be related to the bad mode problem appearing in practice as well.
For example, \citet{wiher2022ondecoding} found that for more ``directed tasks'' (i.e. ones with strong input-output relationships), deterministic decoding tended to work better; \citet{stahlberg2022uncertainty} used an edit-distance based metric as a proxy for measuring uncertainty, and found that MAP-type decoding was better for less uncertain tasks. Refer to Section~\ref{sec:modestheory_related_work} for a few more examples.

Equation~\ref{eq:modes_uniform} generalizes the reasoning from the length example in Section~\ref{sec:modestheory_length}.
To see the connection, consider the simplified case where for a given input $x$, all good outputs have the same length, and there's a bad output which has probability $p$ independent of the input.
Equation~\ref{eq:modes_uniform} says that this bad output could be modal if the entropy rate of the outputs of length $L$ ever exceeds $-\log p / L$.

There's no reason to think the output entropy rate decreases much with length for NLG tasks, other than maybe grammatical error correction.
\cite{catgotyourtongue} found that longer source sentences were \emph{more} likely to be assigned the empty sequence as a mode, which makes a lot of sense if MT has a constant output entropy rate!\footnote{MT doesn't have the guarantee that all translations of a source will be the same length, but they will generally be similar in length}

None of the above rules out model error though, since the above doesn't matter if $p = 0$ for all bad sequences.
We'll now look at some actual MT training data to see if that's the case.

\subsection{Empty sequences in Europarl}
Europarl~\citep{europarl} is a dataset of paired text from the European Parliament, widely used for training MT systems.
As of EuroParl-v7, the empty sequence is the most commonly occurring ``text'' in the English split of the German-English pair.
About 0.39\% of the 1.9 million paired sentences are a German text containing at least one alphabetical character and an empty English ``text.''
The median length in characters of these German texts is 35 characters, and the 90th percentile is 122.
Whatever the cause, the empty sequence is the empirical unconditional mode of this training corpus.

$\log_2 (1/0.0039) \approx 8$, meaning that if a typical high-quality sentence involves more than 8 binary decisions, the number of good translations will be too large for any one of them to be more likely than the empty sequence (on average).
Figure~\ref{fig:empty_europarl} shows the longest German sentence (884 characters long) which is paired with an empty English text.

Conservatively, imagine that all good translations of the German input are no more than 500 characters long.
A conditional entropy rate of even 0.04 bits per character\footnote{I.e. one binary decision per 25 characters} would require the probability of the empty sequence to be less than 1 in a million, or else it will dominate the good translations.

\begin{figure}
    \centering
    \small
    \begin{subfigure}{0.45\textwidth}
        \centering
        \fbox{\parbox{\textwidth}{``Ein solcher Basisbeschluß wird im Allgemeinen Rat gefaßt, so daß es sich - anders als es bei den Europäischen Räten sehr oft der Fall ist - um einen wirklich sorgfältig ausgearbeiteten und durchdachten Beschluß handelt, der dem Europäischen Rat unterbreitet wird; im Allgemeinen Rat kann anschließend im Rahmen eines solchen Basisbeschlusses ein Beschluß mit qualifizierter Mehrheit gefaßt werden, wobei wiederum für einen Mitgliedstaat die Möglichkeit besteht, einen solchen Beschluß anzufechten, wenn einem wichtigen nationalen Interesse dadurch geschadet wird; um zu verhindern, daß eine solche Anfechtung allzu häufig erfolgt, gilt ein Verfahren, wonach Zwei Drittel der Ratsmitglieder die Möglichkeit erhalten, beim Europäischen Rat Berufung einzulegen, damit ein letztes Urteil gefällt wird, wenn eine solche Anfechtung ihrer Ansicht nach zu unrecht erfolgte\texttt{</s>}''}}
        \caption{German source}
        \label{fig:empty_europarl}
    \end{subfigure}
    \hfill
    \begin{subfigure}{0.45\textwidth}
        \centering
        \fbox{\parbox{\textwidth}{``\texttt{</s>}''}}
        \caption{English reference}
        \label{fig:subfigure2}
    \end{subfigure}
    
    \caption{A Europarl v7 De-En training example}
    \label{fig:figure}
\end{figure}

\subsection{Copying source sentences \citep{ott2018analyzing}}
\citet{ott2018analyzing} empirically showed that a particular kind of low entropy behavior had an outsized effect on NMT models in practice.
Specifically, they identify the issue of ``copy noise'', meaning that the target sentence in a training example is simply a copy of the training sentence.
They say: ``Even just 1\% of copy noise can lead to a drop of 3.3 BLEU for a beam of k = 20 compared to a model with no added noise. For a 10\% noise level, all [methods] but greedy search have their accuracy more than halved.''

It is easy to see that at a 1\% noise rate, when the number of equally probable valid translations of a given input passes 100 (or a larger number for unequally probable translations), the global mode of an MT model which perfectly fits the training data distribution will be to copy the input.

\subsection{Generalizing from truncation}
The clear objection at this point is that empty sequences should be filtered out during preprocessing.
If a model assigns a non-zero probability to the empty sequence in this case, there is definitely model error.
However, it could still be more appropriate to think of it as incorrect generalization.

Consider what's going on: Data is being sampled from a distribution which includes empty outputs, then being filtered to remove them.
Suppose you, the reader, were shown a training set of translations which included the following:
\begin{itemize}
\item German: ``Und der Liedtext geht weiter:'', English: ``And''.
\item German: ``(1) zur Hilfe der Europäischen Union für den Iran nach dem jüngsten Erdbeben'', English: ``(1)''
\end{itemize}
(These are also from Europarl, but it's much harder to estimate how much spurious truncation there is than it is to count how many empty English texts there are).

What betting odds would you be willing to take on a bet that the test set included precisely 0 empty outputs?
Very few people would put down even \$1,000 to win \$1, so their personal estimate of $\prob(\text{EOS})$ definitely can't be 0.
It shouldn't be too surprising that spurious EOS's sprinkled in the training data leads models to predict they might occur at the start of the sentence as well.

\citet{whynmtempty} conjectured that NMT models were oversmoothing due to a shared EOS token, and found that introducing per-position EOS tokens led to beam search producing empty outputs much less frequently.
This change makes it harder to make the incorrect generalization described above, and has the intended effect.

Besides the empty sequence problem, the examples above show we also need to worry about other types of degenerate translation.
Even if a model did put no probability on empty outputs, it may (correctly) have the modal output of producing a single word from a translation, \emph{then} EOS.
All the entropy-rate discussion from the previous sections applies to those types of outputs as well.

\subsection{Reinforcement learning-based instruction following}
In this section we'll give a sketch of how similar issues can arise even when training isn't done using MLE of the text distribution.

Consider a system like ChatGPT\footnote{\url{chat.openai.com}}, which is trained with Reinforcement Learning from Human Feedback (RLHF).
In this case, the model isn't being trained to fit any particular $\dtrain$, but the end result is still a distribution over sequences.
What should that distribution be?

One possible ideal distribution is the uniform distribution over all sequences which would get the maximum reward from the human raters.
RL doesn't actually elicit this distribution, since emitting a single output which receives maximal reward is easier and has the same expected reward.
Still, it's probably fair to say that a model which can give many different responses to ``Write a haiku'' is better than a model which just emits the same haiku every time.

One place noise can come from is incorrect human ratings, either due to misreading the text, or making data entry errors.
This is analogous to the corruption of $\dperfect$ discussed in the previous sections.

Just like with MT, we can come up with an input which has a high entropy output distribution.
Then we just have to show that there's a degenerate response with a negative log-likelihood lower than the entropy.

Consider the request ``Write a sonnet about NLP.''
Sonnets usually have around 100 words, so even at an extremely conservative entropy rate of 1 bit per word, there are $2^{100}$ valid responses, meaning we need to find a bad response which should have a probability higher than $1/2^{100}$.
One degenerate response is $y =$ ``I can't.''
Call the event that $y$ is maximally rated $R$, and the event that it's generated\footnote{by whatever distribution is being used to produce inputs to the rating process} $G$.
Then:
\begin{equation*}
    \prob(G | R) = \frac{\prob (R | G) P(G)}{P(R)}
\end{equation*}
The denominator on the right-hand side is at most 1, so the probability is bounded above by the numerator.
The numerator is just the product of the probabilities that ``$y$ is incorrectly given the maximal rating'' and ``$y$ is generated by the pre-RLHF model.''
In order for these probabilities to multiply to less than $2^{-100}$, one or both of them would need to be less than $2^{-50}$ (less than 1 quadrillionth).

Now, the ``equally probably outputs'' assumption isn't accurate for the current generation of chatbots.
On July 18, 2023, sampling outputs from GPT-4 for the above prompt 20 times led to 18 responses which started with ``Upon'', with the other two starting with ``In the realm.''
``Realm'' appears in the first line of 7 out of 20 responses. We'll look at modal outputs for some open source instruction-following chatbots in Chapter~\ref{chapter:modes_exact}, and show that this behavior is present in them as well.

So maybe beam search or other MAP approximations \emph{would} work well on these models, since their output entropy is so much lower than the ideal distribution.
We probably do want to produce better chatbots over time though, so the entropy of their outputs will necessarily rise (at least for certain inputs).

\section{Implications}\label{sec:modestheory_proposal}
Getting back to the ``oracle criterion'' again, it's doesn't seem like any common decoding method would consistently give us good samples from an oracle for $\ptrain (y_t | x, y_{1:t-1})$.
If the training distribution contains low-quality/low-entropy outputs (e.g. the empty sequence, ``I don't know'', ``As an AI language model, I cannot...''), then likelihood maximization strategies will fail.
On the other hand, if a typical sample is corrupted by a small amount of noise, sampling-based methods will also produce outputs with the same amount of noise.

The problem is at least partly that likelihood alone doesn't let us distinguish between low-probability but meaningful information, and low-probability noise such as typos.
As far as the distribution is concerned, they're both high-information!
Similarly, it can't let us distinguish between text which is high-probability because it is relevant to the input, and text which is high-probability because the distribution has low-quality modes.

At the very least, if you find the examples above convincing, don't conclude that your model learned the data poorly just because it outputs some low-quality text.
To assess whether a model has learned the distribution, you should draw samples from it and measure their statistics, as done by \citet{inadequacyofmode}.
You might need to use evaluations of single outputs for practical reasons, but it's important to view the results as a combined evaluation of \emph{both} the model and a decoding strategy.

For convenience we'll repeat Equation~\ref{eq:noncommutative} here:
\begin{equation*}
    \argmax\limits_{a \in \mathcal{A}} \prob\left(A(X) = a\right) \neq A\left(\argmax\limits_{x \in \mathcal{S}} P(X = x)\right)
\end{equation*}
We said before that the non-intuitive behavior of non-typical outputs is because these two quantities aren't equal.
The right hand side is what we're seeing when a model's mode is the empty sequence, and the left hand side is what we're seeing when a sampling-based method spits out noise at the same rate as it was present in the training data.

But this suggests a way to improve our decoding outputs, we can just condition on the argmax value from the left-hand-side of Equation~\ref{eq:noncommutative}.
Imagine translating this thesis into German, and call $l^*$ the modal translation length (\emph{not} the length of the mode).
$l^*$ is definitely greater than 10,000 but the modal translation would be much shorter (or even empty), due to the entropy rate issue discussed previously.
Instead of searching for the argmax translation $x \in \seqs$, we should be looking for the conditional argmax:
\begin{equation*}\label{eq:cond_mode}
    x_\text{cond}^* = \argmax\limits_{x \in \seqs} \prob \left( X = x \given[\big] |x| = l^*\right)
\end{equation*}
This is actually already common in another setting, although it's not usually framed this way.
Non-autoregressive (Non-AR) NMT systems usually \emph{have} to do this because of their architecture.
For example, SUNDAE~\citep{sundae} predicts the length of the target sentence from the source sentence, then searches for a high scoring sequence \emph{of that length}.
If someone suggested trying to maximize the product $\prob(\text{length}) \prob(X | \text{length})$ in that setting, everyone would think they were crazy!
This might partially explain why short outputs are often seen as a bias of autoregressive models.
Non-AR models avoid the problem by construction!

To avoid degeneracies other than brevity, we could imagine conditioning on other attributes as well.
If the modal response from a chatbot is ``I don't know'', it would be much more interesting to find the modal response \emph{conditional on attempting to reply}.

A reasonable question is whether this will actually give us good outputs, or if we'll just be patching up an endless sequence of failure modes.
For example, maybe the modal length-20,000 translation of this thesis is just the word ``NLG'' repeated 20,000 times.
In the next chapter, we'll look at some exact length-conditional modes of a pretrained NMT model and see that they're actually pretty high quality!

Since finding exact modes is expensive, if not intractable, in Chapter~\ref{chapter:modes_beam} we'll propose a beam search variant which can be used for search conditional on an attribute.
It doesn't just work for controlling length in MT, but also for applying beam search to an open domain LM.
We aren't proposing that writing a sonnet with MAP decoding is a good idea, but for settings where correctness is important, there's a lot to be gained by being able to do some sort of likelihood maximization.
The idealized end-goal of this line of research would be to decode from the conditional probability $\prob\left(x\given x\text{ is not degenerate}\right)$.

We should also point out that the ``attribute of mode does not equal modal attribute'' problem also suggests another solution.
If sampling-based repeats noise from the training distribution, you could for example sample from the conditional probability $\prob(x \given x\text{ is fluent})$.
More ambitiously, you could even consider sampling from $\prob(y \given x,\;\text{$y$ is a translation of $x$})$.
The idealized final result for RLHF training of LMs would be to directly sample from $\prob(y \given x,\; y\text{ is high-reward})$.

Sampling from conditional distributions has been researched before, although the conditioning information is generally things like topic or sentiment.
The most relevant method is FUDGE, which uses a classifier which predicts the final value of an attribute given a partial model output.
The way to use FUDGE for this particular task would be to train a classifier to predict the total number of typos in the output, or whether the final output will be a valid translation of the source sentence given the output so far.

Basically, we're suggesting using search on $\prob ( x | \text{$x$ is not degenerate})$, or sampling on $\prob (x | \text{$x$ is correct/not noisy})$.
Of these two options, we'll be focusing on search over sampling.
The first reason is that search (especially for language models) has received less attention than sampling, so we expect that there's more room for improvement.
Secondly, the relevant problem for fixing MAP degeneracies is predicting things like ``will the output be empty'' or ``does the output attempt to fulfill the prompt'', which we expect to be much easier to learn than ``will the output be factual.''
That being said, we hope that future work will investigate sampling from these conditional distributions.

\section{Conclusion}
The main point of this chapter has been to convince you that many observed degeneracies in NLG \emph{could} be due to properties of the training data distribution, rather than directly implying model error.
We described noise models for both MT data and RLHF trained LMs which would lead to these phenomena being present even in realistic settings.

Our decoding methods need to take this into account, since we'll be stuck with the issue unless we manage to only train on extremely clean data.
Our proposed path forward is to explicitly incorporate more information about the properties we want in our decoding outputs, rather than using methods which only depend on model likelihood.

\chapter{Exact Modes in Machine Translation, Story Completion, and Instruction Following}\label{chapter:modes_exact}

In the previous chapter we argued that the bad mode problem might be due to properties of our training data distributions, and therefore is not necessarily evidence of model error.
If correct, this implies that no amount of improvement in modeling will overcome the likelihood/quality mismatch.
Because of this, we suggested that one possible decoding method would be to search for the mode of conditional distributions of the (optimistic) form $\prob \left(x \given \text{$x$ is not degenerate}\right)$.

The main question to be answered is: if we condition away one issue, such as empty outputs, will we just run into another degeneracy such as repetition or disfluency?
In this chapter we give some evidence that \emph{conditional} model modes are often high-quality if we condition away the specific problem of length-degeneracy.
However, not only do we find that the problem hasn't abated with scale, we also find that there are several issues with exact modes in a family of larger language models when they are applied to an instruction following task.
These findings come from case studies of exact model modes for three tasks: Machine translation (MT), cloze completion, and open-ended instruction-following/question answering.
The interesting implication is that while many model modes of larger scale models are degenerate in various ways, many are not, suggesting that conditioning away the problems may be a viable path forward.

\paragraph{NMT and Cloze completion.} First, we replicate \cite{catgotyourtongue}'s finding of NMT models having the empty sequence as their mode, and find the same result with a language model trained to complete stories from the ROC stories~\citep{roc_stories} corpus.
However, when we instead search for \emph{length-conditional} modes, we find that they're generally high-quality, rather than just displaying other problems such as repetition.
This suggests that it might not be too hard to condition away the bad types of atypicality discussed in the last chapter.

\paragraph{Scaling up with LLaMA.} We also report, for the first time, on the exact modes of more modern language models, specifically from the LLaMA~\citep{llama} model family.
We find unconditional modes for instruction-following inputs for LLaMA-7B, Alpaca-7B~\citep{alpaca}, and Guanaco-7B~\citep{qlora}, and find:
\begin{itemize}
\item Despite these models having over an order of magnitude more parameters than the previous generation of models, their modes are still often empty.
\item When the mode for the instruction finetuned models is not empty, it is usually high quality.
\item The mode is more likely to be empty when the input is more open-ended. We verify this both qualitatively and quantitatively.
\item LLaMA-7B's modes \emph{do} display degeneracies other than emptiness.
\end{itemize}

Overall, our findings show that all the models we test have fairly high-quality global modes, \emph{except} for certain degeneracies which we discuss.
In the next chapter we will take advantage of this fact to develop a version of beam search that approximately searches for high-scoring sequences while avoiding degenerate outputs.

\section{Background: Exact modes of NLG models}
It's commonly assumed that exact MAP decoding from autoregressive NLG models is intractable due to the exponential size of the search space, but this is not correct.
\citet{catgotyourtongue} pointed out that simple depth-first search (DFS) with pruning can find the exact modal output of NMT models, since the properties of the search problem make it so that aggressive pruning is possible.
For example, \citet{stahlberg2022uncertainty} showed that the exact mode could usually be found for a transformer MT model while searching less than 1 million states.

To show how pruning works, here's a quick example.
Suppose that during tree search, a complete translation with a (model) probability of $1/5000$ has been found.
Two observations are necessary:
\begin{enumerate}
    \item Any partial hypothesis with a probability less than $1/5000$ can be discarded, since adding any more tokens can only lower the probability.
    \item Every search node gives you a possible complete hypothesis to consider, since you'll already be evaluating the probability of every possible continuation, including the next token being EOS. (And complete hypotheses are exactly the sequences that end in EOS).
\end{enumerate}
These features make it possible to prune an enormous fraction of the search space, making exact inference tractable.

As mentioned in the last chapter, this method showed that over half of the source sentences they tested on had the empty sequence as their modal translation, and longer source sentences were \emph{more} likely to have an empty modal translation.

It's already implicit in how we use beam search that we're not \emph{really} looking for high scoring outputs.
If someone wanted to find higher scoring outputs with beam search, they could make (beam size) $\times$ (sequence length) complete hypotheses by just adding EOS to each partial hypothesis considered during search.
This takes no extra model forward-passes, since the probability of every next token (including EOS) is already calculated when new hypotheses are being generated.
This would let ``beam search'' with a beam size of 1 find the empty sequence when it's the modal translation.

We don't want to find these outputs, so we don't make this ``improvement'' to beam search.
\citet{meister2020ifbeamsearch} suggested that beam search may actually be the most successful decoding method because it has a useful inductive bias, which could explain why it is still useful, even if making the beam size larger or considering more hypotheses is harmful.

\citet{stahlberg2022uncertainty} look at search for exact modes in more depth.
They look at the relationship between intrinsic uncertainty (whether an input allows many different outputs), search errors, and how many states DFS explores before finding the mode.
They operationalize intrinsic uncertainty as average edit distance between references divided by average reference length, for multi-reference tasks.
They find that more uncertainty leads to DFS taking longer to find the mode. 
They also find the $N$ highest scoring sequences for each task, and find that those sequences do actually represent a larger fraction of the total probability mass for inputs with lower intrinsic uncertainty (i.e., GEC inputs or shorter MT inputs).

\section{Optimizing memory usage for DFS on transformers}\label{sec:modesexact_dfs_mem}
In this section we'll briefly describe a method for reducing the memory usage necessary for running DFS on transformer NLG models.
Generating from a transformer requires caching the key and value vectors for previous timesteps, in order to avoid a large amount of recomputation for each token.
Running DFS to depth $k$, and storing a KV-cache of length $k$ at each nodes, leads to storage that is quadratic in $k$, even though only $k$ nodes are active at a time.

For the MT and GPT-2 models we experiment with, the maximum search depth and hidden state dimension are both small enough that we can do this with no issue.
The LLaMA-based models however, are over an order of magnitude larger, and also frequently lead to searching subtrees that are hundreds of tokens deep.
As such, we need some way to avoid actually storing a full KV-cache at each node.

Empirically, the DFS search order for these models often involves some path through the search tree being greedily expanded, without any branching.
That is, many search nodes will only have one child which gets explored.
For these nodes, storing the KV-cache for later is a waste of memory since it will never be re-used.
We don't know up-front which nodes will and won't be re-used, but we can still save some memory without losing too much performance by taking a heuristic approach.

To reduce storage while still avoiding running the model on a prefix more than once, each search node initially only stores the hidden state for the token that it used as input.\footnote{Our implementation of DFS is just recursive, so when we say that a search node stores something, we mean that it's stored in the Python interpreter's stack frame for that call to the DFS function.}
Once a node's second child is about to be expanded, the full KV-cache is reconstituted from the keys and values stored at that node and in its ancestors.
Specifically, the node uses back pointers to go back up the search tree until some node that has a full KV-cache is found.
This way, a greedy search path out to depth $k$ will only require $O(k)$ memory instead of $O(k^2)$.

In summary, when search node at depth $k$ is evaluated, a $k \times d$ key/value cache\footnote{The KV-cache is actually a list of a key and value cache for each layer, so the size-$d$ dimension should be seen as a concatenation of all these different values.} $\bm{h}_{1:k}$ is produced
It is then processed as follows:
\begin{enumerate}
    \item The search node saves the vector $\bm{h}_{k}$
    \item The full cache $\bm{h}_{1:k}$ is passed to the first child node, which uses it for a forward pass then frees it
    \item If a second child node will be expanded, the search node recomputes the full cache $\bm{h}_{1:k}$ using its cached vector and those cached in its ancestors. This time the cache is saved for use in the third, fourth, etc. child instead of being freed after the second child uses it.
\end{enumerate}

This heuristic isn't optimal by any means, but it lets us avoid running out of memory when the search state gets hundreds of nodes deep.
Some potentially better methods include using a LRU cache that limits the number of full caches in memory, or using the next-token probabilities at each node to make a smarter decision about whether a full or partial cache should be kept.

\section{Exact modes of an MT model}
In this section, we use DFS to find the exact modes of the pretrained Chinese-English MarianMT model~\citep{marianmt} on the WMT'17 Zh-En \texttt{newsdev} dataset~\citep{wmt17}.
The model is a standard transformer using the architecture of \citet{vaswani2017attention}.
Just like \citet{catgotyourtongue}, we find that the model's mode is often empty, and that this occurs more often as the length of the source sentence increases.
In addition to replicating their results, we also qualitatively examine the modal outputs to see if the the empty modes were just hiding further degeneracies (e.g., disfluency or repetition).

One important thing to note about this model is that it was trained with 10\% label smoothing.
This means that the model is trained with a (token-wise) 90-10 mixture distribution of actual translations and the uniform distribution on its vocabulary.
This mixture distribution puts a hard upper bound on how long the modal output can be (See \citet{liang2022implicit}), but \citet{riley2022continuum} found that adding label smoothing didn't tend to change the length or repetition bias in beam search, so this level of label smoothing might be too small to change the actual modal sequence.

\subsection{Unconditional modes}
\begin{figure}
  \centering
  \begin{subfigure}{0.4\textwidth}
    \centering
    \includegraphics[width=\linewidth]{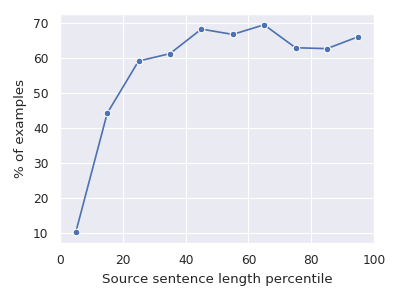}
    \caption{Percent of source sentences that have the empty sequence as their modal output.}
    \label{fig:mt_empty_percent}
  \end{subfigure}%
  \hfill
  \begin{subfigure}{0.5\textwidth}
    \centering
    \includegraphics[width=\linewidth]{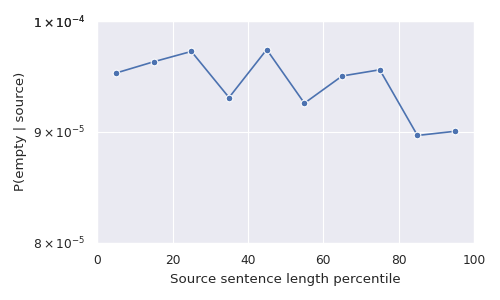}
    \caption{Geometric mean of the model's probability of the empty sequence given the input.}
    \label{fig:mt_empty_prob}
  \end{subfigure}
  \caption{MarianMT Zh-En predictions of empty outputs on WMT17 Zh-En \texttt{newsdev-2017} set (2002 examples). Source sentences are grouped into 10 equally sized bins by length.}
  \label{fig:mt_empty_srclen}
\end{figure}

As expected, we find that the MarianMT Zh-En model predicts that the empty sequence as the mode for 57.7\% of the 2002 source sentences.\footnote{For reference, \citet{catgotyourtongue} found that 51.8\% of inputs on the WMT \texttt{news-test-2015} En-De dataset had an empty mode under their model.}
Figure~\ref{fig:mt_empty_srclen} shows the relationship between the length of the input sentence and empty outputs.
Just like \citet{catgotyourtongue}, we find that longer source sequences are \emph{more} likely to have an empty modal output (Figure~\ref{fig:mt_empty_percent}. 
We can also just read the probability of generating an empty output directly off the decoder logits, which is shown in Figure~\ref{fig:mt_empty_prob}.
It shows that the average log-probability of the empty output only goes from about 1 in 10,000 to 9 in 100,000 as the source length increases.

These results are consistent with the explanation given in Chapter~\ref{chapter:modes_theory}: The main cause of the empty mode problem is that the entropy of valid outputs increases with input length, but the probability of the empty output does not decline fast enough to offset this effect.

\subsection{Length-conditional modes}
\begin{table}[h]
    \begin{CJK*}{UTF8}{gbsn}
    \centering
    \caption{Global and length-conditional modal translations of ``泰国旅游景区炸弹爆炸致四人死亡'' by the MarianMT Zh-En translation model. The reference translation is ``Thailand Bomb Blasts At Tourist Hotspots Kill Four'' (14 tokens).}\label{tab:marian_example}
    \begin{tabular}{lll}
         Length constraint (tokens)&  Log-probability & Text\\
         \midrule
        Global mode & -7.91 & \texttt{</s>} \\
        4 & -9.22 & Four people died\texttt{</s>} \\
        6 & -9.77 & Four people were killed.\texttt{</s>} \\
        8 & -10.37 & In Thailand, four people died.\texttt{</s>} \\
        10 & -10.63 & A bomb blast in Thailand killed four people.\texttt{</s>} \\
        12 & -9.60 & The bombing of the Thai tourist zone killed four people.\texttt{</s>} \\
        \bottomrule
    \end{tabular}
\end{CJK*}
\end{table}
We'll now look at \emph{conditional}-modes of this model, which are the exact values of:
\begin{equation*}
   y^* = \argmax\limits_y \pmodel\left(y \given[\big] x, |y| = L\right)
\end{equation*}
for a given target length, $L$.

The main thing we want to know is whether conditioning away the degenerately short outputs uncovers new issues such as disfluency and repetition, or if the length-conditional modes are high-quality.
Our main finding is that these conditional modes are indeed high-quality, provided the length constraint is long enough.
We'll mostly look at these modes qualitatively, for a more quantitative investigation of length-conditional modes, see \citet{catgotyourtongue}.\footnote{They instead condition on the length being \emph{at least} some target value, whereas we search for outputs of exactly a given length.}

One computational problem we face is that finding length-conditional modes is much harder than finding global modes.
This is because there are far fewer opportunities to prune the search tree than in unconditional search.\footnote{For finding a conditional mode of length 12, you can only prune a depth 5 subtree if it's lower probability than the best length 12 sequence found. That will rarely be the case, so you need to check more of these small trees than you would when you're searching for a global mode.}
Because of that, the highest length we targeted with conditional search was 12 tokens.

\subsubsection{Characteristics of length-conditional modes}
A common pattern in the conditional modes is that the shortest modes will not be complete sentences, but those of length 10-12 will be grammatical (while still leaving out some content by necessity).
Table~\ref{tab:marian_example} shows an example of this behavior for an input that is assigned the empty sequence as its global mode.
The length 4 mode is incomplete, but each increase in the length constraint adds more information.
The length 8 mode adds that the blast happened in Thailand, the length 10 mode adds that it was a bomb, and the length 12 one adds that it was in a tourist zone.

Table~\ref{tab:mt_modes} shows 20 more search outputs, that were randomly selected to avoid cherry-picking.
Sources A-J were randomly selected from the sequences with reference lengths between 5 and 15 tokens, so that we could search for outputs of similar length.
The problem is that the short sequences are least likely to have empty modes, so even the global modes are high quality.
Sources K-T were sampled from the sources with reference lengths between 25 and 35 tokens for which the model \emph{did} predict that the mode was empty.

\paragraph{Modes for short reference sentences.} For all the sources with short references, almost all of the length-conditional modes are grammatical.
The only strange sentence is the length 12 mode for Source G, which has spurious punctuation.
The differences in meaning from the reference translations are due to lack of details in source sentences.
Notably, the global modes for sources A, C, and D are either ungrammatical or not good translations of the source.

\paragraph{Conditional modes when the mode is empty.}
Examples K-T in Table~\ref{tab:mt_modes} show length-conditional modes for some sources for which the model predicts the empty sequence as the modal output.
Interestingly, the mode of length 12 is grammatical for 8 of the 10 sources (all but N and S).
There is no repetition, the most commonly observed other search degeneracy in beam search.\footnote{Repetition is locally super high probability: If the model outputs ``cat'' 20 times, the event that it is output a 21st time is extremely high probability. The issue is that at some point the model has to output the EOS token, which is getting lower and lower in probability the more times the repetition happens.
This makes it so that repetition can trap methods like beam search that need to make local decisions, but it doesn't actually give high-scoring full sequences.}.

The main cause of ungrammatical outputs seems to be the prescribed length being too short for all the information in the source sentence.
The reason we're stuck with this problem is that the sentences that tend to have empty model modes have reference lengths too long for us to do exact search for.
Language modeling tasks have higher entropy than MT, so we'll see empty modes at much shorter output lengths when we look at LMs in the next section.

\subsection{Discussion: Exact modes in machine translation}
Running exact search on the Marian MT Zh-En model replicated \citet{catgotyourtongue}'s finding that the unconditional mode is often empty.
On the other hand, length-conditional modes seem to generally be high quality.
The other ``degeneracy'' we see in the conditional modes is that they may be truncated when the conditioning is too short. 
That seems to be mostly due to the fact that we have to condition on lengths much shorter than the reference translations.\footnote{In the next chapter when we look at approximate search for conditional modes, we'll instead be able to condition on the length being equal to the reference length, which will fix that problem.}

Overall, we consider these findings an encouraging sign, since we didn't see repetition or disfluency in the conditional modes.
This supports our suggestion in Chapter~\ref{chapter:modes_theory} that finding the conditional argmax of a probability distribution might avoid the bad mode problem.

\section{Exact modes of a cloze completion LM}
The issue with looking at length-conditional modes of NMT models is that their output entropy rate is relatively low.
That makes it so that by the time outputs get long enough for degenerate modes to occur, they're also too long for length-conditional search to be practical.

To move to a more open-ended setting that also has outputs of tractable lengths, we used the ROC stories dataset~\citep{roc_stories}.
ROC Stories is a dataset of 5 sentence stories, where the goal is to predict the last sentence based on the first 4.
The intended way to evaluate models is to have them choose between two alternatives, but we just used it as a language modeling dataset since it has a very constrained format.
We finetuned\footnote{Hyperparameters: Adam~\citep{adam} with a learning rate of $2 \times 10^{-5}$, and a batch size of 32.} the 345M parameter GPT-2 model~\citep{gpt2} on the dataset and searched for unconditional and length-conditional completions on the validation set.
We used early stopping based on validation set perplexity over the 12 epoch training run.\footnote{This technically leaks information since we also look at modes on the validation set itself. However, we're just selecting between 12 models (1 after each epoch), which means there aren't very many bits of leakage.}

\subsection{Unconditional modes}
\begin{figure}
  \centering
  \begin{subfigure}{0.4\textwidth}
    \centering
    \includegraphics[width=\linewidth]{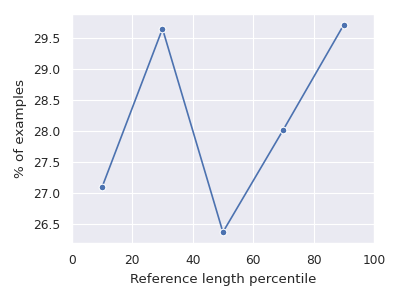}
    \caption{Percent of stories that have the empty sequence as their modal continuation.}
    \label{fig:roc_empty_percent}
  \end{subfigure}%
  \hfill
  \begin{subfigure}{0.5\textwidth}
    \centering
    \includegraphics[width=\linewidth]{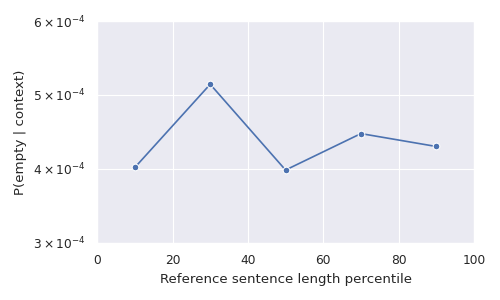}
    \caption{Geometric mean of the model's probability of the empty sequence given the first four sentences of the story.}
    \label{fig:roc_empty_prob}
  \end{subfigure}
  \caption{Finetuned GPT-2-345M predictions of empty outputs on the ROC Stories validation set (1571 Stories). Stories are grouped into 5 equally sized bins by reference continuation length.}
  \label{fig:roc_empty}
\end{figure}
The empty sequence was the mode for 28.71\% of the 1571 ``Winter 2018'' validation set stories.
Figure~\ref{fig:roc_empty} shows that, unlike the NMT case, there's not a clear correlation between length and the probability of the mode being empty.
This is probably just due to the fact that the output lengths don't vary much.

The probability of the empty sequence averages to around 4 or 5 in ten thousand, which is quite a bit higher than in our MT experiments.
Since the ROC stories data is much cleaner than MT training data, this definitely represents model error.
The reason for it is likely just that finetuning didn't completely overwrite the base GPT-2 models' distribution of when EOT should be emitted.

\subsection{Length-conditional modes}
\begin{table}[h]
    \centering
    \caption{Modal continuations of several lengths for prefix: ``Sarah always had a fascination with the night sky. Noticing her passion, Sarah's father bought her a new telescope. She was ecstatic. She went outside every night to diligently view the night sky.'' The reference continuation is ``Sarah loved her new telescope.''}\label{tab:roc_example}
    \begin{tabular}{lll}
    Length Constraint (tokens) & Log-probability & Text\\
    \midrule
    Global mode & -7.79 & \texttt{<|endoftext|>}\\
    5 & -9.14 &  Sarah loved astronomy!\texttt{<|endoftext|>}\\
    6 & -7.97 &  Sarah never looked back.\texttt{<|endoftext|>}\\
    7 & -8.59 &  Sarah loved her new telescope.\texttt{<|endoftext|>}\\
    8 & -9.38 &  Now, Sarah is an astronomer.\texttt{<|endoftext|>}\\
    9 & -8.68 &  Sarah was happy with her new telescope.\texttt{<|endoftext|>}\\
    10 & -8.77 &  Sarah was very happy with her new telescope.\texttt{<|endoftext|>}\\
    12 & -8.91 &  Sarah was amazed by the beauty of the night sky.\texttt{<|endoftext|>}
    \end{tabular}
\end{table}

\begin{table}[h]
    \centering
    \caption{Modal continuations of several lengths from a GPT2-345M model finetuned on the ROC stories corpus. The input was: ``Kaylee always wanted a puppy. On her birthday her parents took her to a farm. There were lots of beagle puppies there. Her parents told her she could pick a puppy for her birthday.'' The reference continuation is ``Kaylee was thrilled!''}\label{tab:roc_example2}
    \begin{tabular}{lll}
    Length Constraint (tokens) & Log-probability & Text\\
    \midrule
    Global mode & -6.55 &  Kaylee picked a beagle puppy.\texttt{<|endoftext|>}\\
    5 & -9.00 &  Kaylee cried.\texttt{<|endoftext|>}\\
    6 & -7.66 &  Kaylee said yes.\texttt{<|endoftext|>}\\
    7 & -6.73 &  Kaylee was so happy.\texttt{<|endoftext|>}\\
    8 & -7.25 &  Kaylee picked a black lab.\texttt{<|endoftext|>}\\
    9 & -6.55 &  Kaylee picked a beagle puppy.\texttt{<|endoftext|>}\\
    10 & -7.01 &  Kaylee picked a black and white puppy.\texttt{<|endoftext|>}\\
    12 & -7.98 &  Kaylee picked a black and white beagle puppy.\texttt{<|endoftext|>}\\
    \bottomrule
    \end{tabular}
\end{table}

Just like with the MT model, the GPT-2 model's length-conditional modes are high-quality, even when its global mode is not. Table~\ref{tab:roc_example} shows one example of this behavior. 
The mode is empty, but the length conditional modes are all plausible completions of the story, and don't display any degeneracies such as repeating earlier text from the story.

An interesting feature of these constrained modes is that the content can be correlated with the length in clear ways. Table~\ref{tab:roc_example2} shows an example where the mode of length 5 is significantly different from all the other modes.
It may be impossible to produce a 5 token output that has the right content, but the model ``prefers'' to output something grammatical, so we see different content.
This is different from the short NMT modes, which were often truncated when the constraint was too short to express the content of the source sentence.

In order to show that these patterns aren't just cherry-picked, randomly sampled examples of modal outputs are shown in Table~\ref{tab:roc_modes}.
All 30 of the conditional modes are grammatical, relevant to the context, and don't show any evidence of degenerate behavior.
This is further evidence that conditional MAP inference may be a promising direction of investigation.

\subsection{Discussion: Exact modes for cloze completion}
Our findings for this finetuned GPT-2-medium model were essentially the same as what we found for the MT model.
Specifically, the empty sequence is often the modal output, but conditioning on a desired length can yield better results.
The length conditional modes are essentially always grammatical, unlike for the MT model where they were often truncated for shorter target lengths.
This is probably because we were able to condition on lengths that were similar to the length of the actual reference continuations, which was not the case for our MT experiments.

To our knowledge, this is the first time exact and conditional modes have been investigated for a language model, so this generalizes the prior findings of \citet{catgotyourtongue}.
A priori, we were worried that length-conditional modes from GPT-2 might just be a single word repeated over and over, or a repetition of an earlier sentence from the context.
The fact that they're high-quality even for this open-ended task is extremely encouraging.

\section{Exact modes of LLaMA-based models}
The earlier experiments replicated~\cite{catgotyourtongue}'s results, and extended them from MT models to a language model as well.
The length-conditional search results showed that these models' modes are usually high-quality, as long as the length is constrained to an appropriate value.

However, those experiments used models that are much smaller than modern LLMs, as well as the LM experiment using a synthetic task (no one actually wants a model that adds a 5th sentence to 4 sentence stories).
To extend our work to a more realistic setting, we also run exact search on three variants of the 7B parameter LLaMA model~\citep{llama} using prompts from the \texttt{databricks-dolly-15k} dataset~\citep{dolly15k}.
The three models are: the base LLaMA-7B model, Alpaca 7B~\citep{alpaca}, and Guanaco-7B~\citep{qlora}.
LLaMA is trained as a general language model, while the other two are finetuned from it on instruction following specifically.

Since these models are over 10 times larger than the models in the previous sections, we only run exact mode search and not length-conditional mode search.
The goal of examining this modes is to verify the assertion that ``the bad mode problem won't go away with scale or improved training'', from Chapter~\ref{chapter:modes_theory}.

\subsection{Details of mode search for LLaMA-family models}
We sampled 1000 instructions from the \texttt{databricks-dolly-15k} dataset, filtered to be less than 256 tokens long, and to be in the instruction/response format rather than the instruction/context/response format.
For Alpaca and Guanaco, we use the prompt format used during finetuning, but LLaMA isn't trained for instruction following so we just use the Alpaca format for it as well (See Appendix~\ref{chapter:prompts_appendix} for the exact text).

We ran DFS on the same set of inputs for each of these models, using the caching strategy we described in Section~\ref{sec:modesexact_dfs_mem}.
One issue is that Guanaco was trained on multi-turn conversations, which means it will generate more messages instead of outputting EOS.
To fix that we treat ``\texttt{\textbackslash{}n\#\#\#}'' as an alternative EOS marker in addition to \texttt{</s>}, which prevents it from generating more than one message.

\subsection{Quantitative Results}
\begin{table}[h]
    \centering
    \begin{tabular}{lrr}
        Model & \% modes empty & $P(\text{empty})$\\
        \midrule
        LLaMA & 70.7 & $1.13 \times 10^{-4}$\\
        Alpaca & 16.0 & $1.80 \times 10^{-5}$\\
        Guanaco & 7.70 & $2.48 \times 10^{-6}$\\
    \end{tabular}
    \caption{Statistics of 7B parameter LLaMA variant models' predictions of empty modes.}
    \label{tab:llama_empty_stats}
\end{table}

\begin{figure}
  \centering
  \begin{subfigure}{0.45\textwidth}
    \centering
    \includegraphics[width=\linewidth]{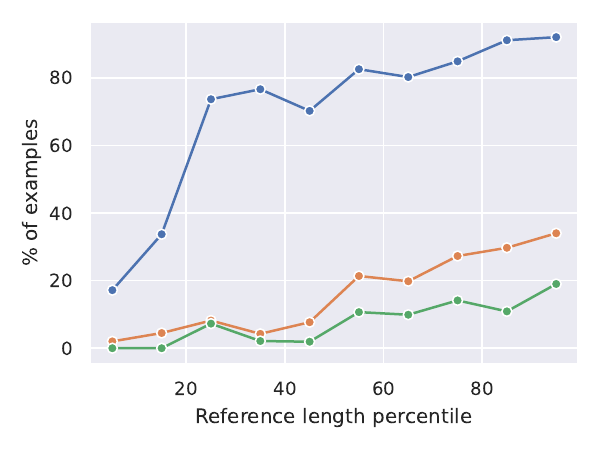}
    \caption{Percent of prompts for which each model's modal response is the empty sequence.}
    \label{fig:llama_empty_percent}
  \end{subfigure}%
  \hfill
  \begin{subfigure}{0.45\textwidth}
    \centering
    \includegraphics[width=\linewidth]{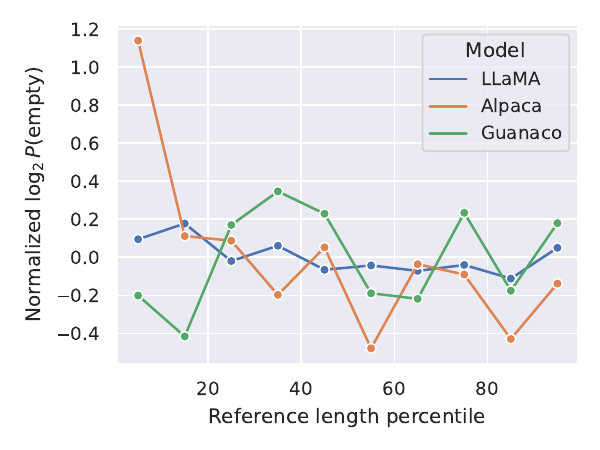}
    \caption{Difference between the average base-2 log-probability of the empty sequence for a particular reference length decile, and the average over all reference lengths.
    This relative view is to put all the log probabilities in the same range, see Table~\ref{tab:llama_empty_stats} for the absolute values.}
    \label{fig:llama_empty_prob}
  \end{subfigure}
  \caption{LLaMA 7B, Alpaca 7B, and Guanaco 7B predictions of empty outputs on 1000 prompts from the \texttt{databricks-dolly-15k} dataset. Prompts are grouped by the decile of the length of the reference response.}
  \label{fig:llama_empty}
\end{figure}

\begin{figure}[htbp]
  \centering

  \begin{subfigure}{0.45\linewidth}
    \includegraphics[width=\linewidth]{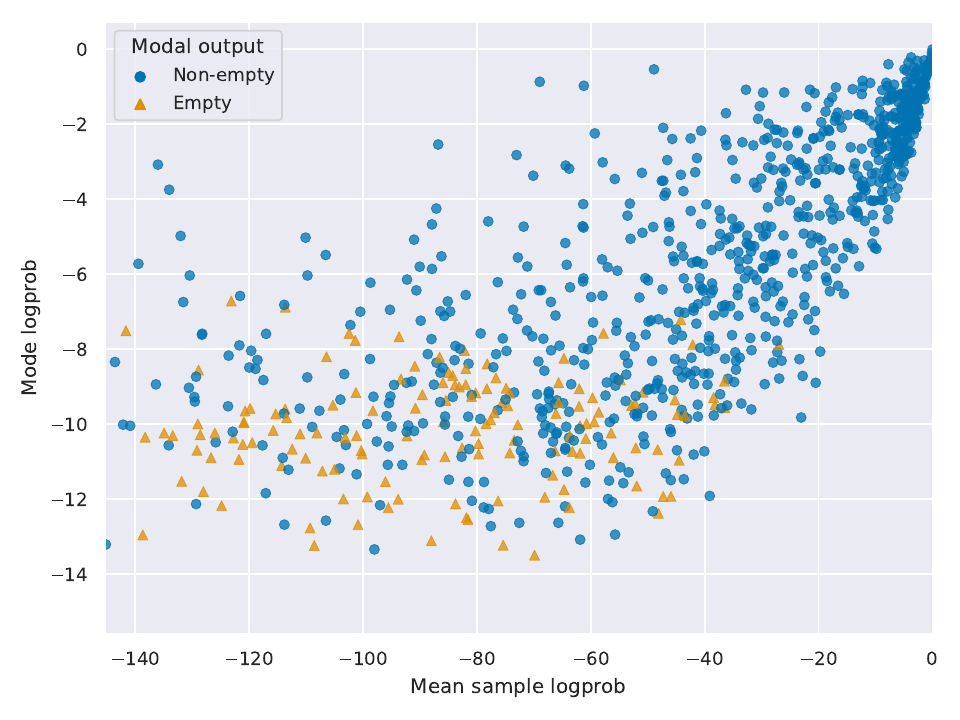}
    \caption{Alpaca}
  \end{subfigure}
  \hfill
  \begin{subfigure}{0.45\linewidth}
    \includegraphics[width=\linewidth]{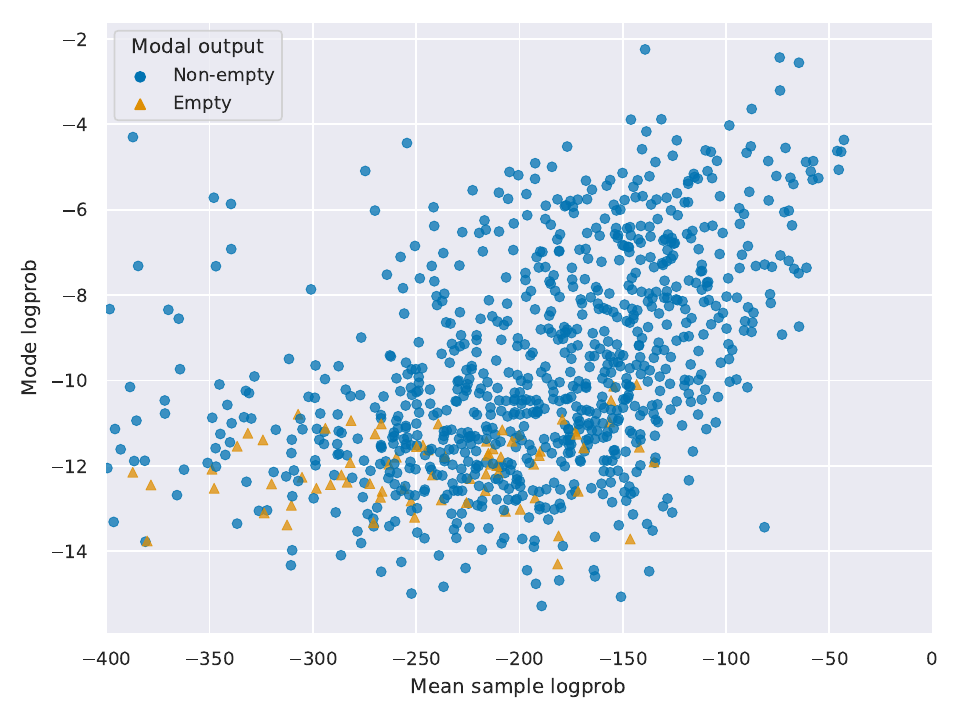}
    \caption{Guanaco}
  \end{subfigure}

  \begin{subfigure}{0.6\linewidth}
    \includegraphics[width=\linewidth]{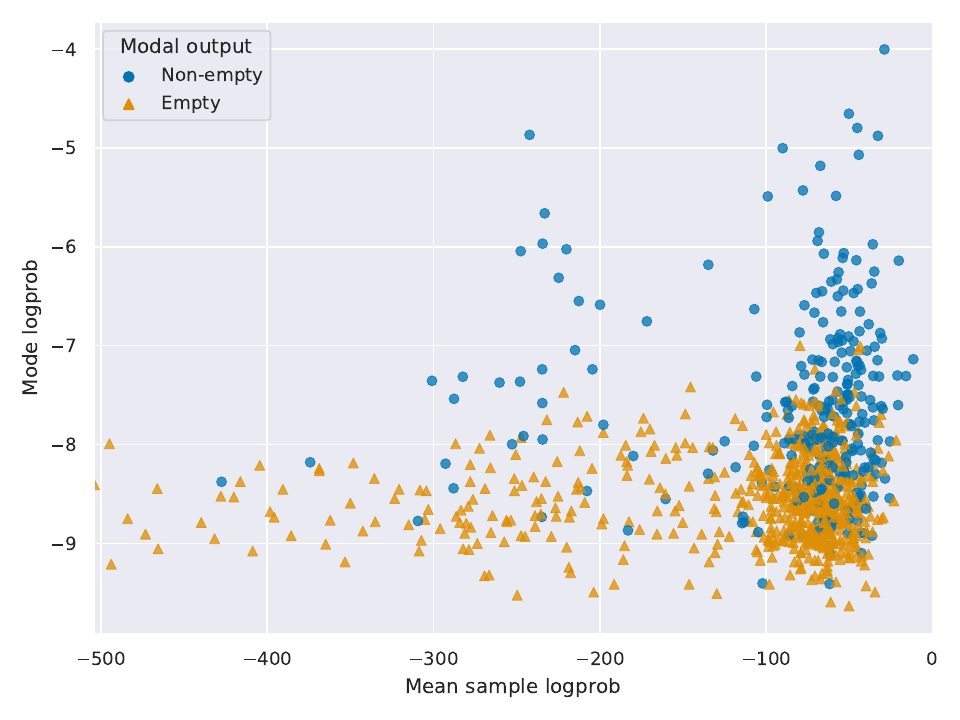}
    \caption{LLaMA}
  \end{subfigure}

  \caption{The log-probability of the modal (y-axis) output compared to the mean log-probability of 5 samples for each of 1000 prompts from the \texttt{databricks-dolly-15k} dataset. Empty modes tend to only occur for lower sample-logprob (higher-entropy) inputs for Alpaca and Guanaco. The 5\% of points with the lowest ``Mean sample logprob'' values have been truncated for readability.}
  \label{fig:sample_logprobs}
\end{figure}

Just like with the smaller models, all three of the LLaMA model variants often have an empty modal output.
Table~\ref{tab:llama_empty_stats} shows the percent of prompts that cause each model to predict an empty mode.
LLaMA has an empty mode more than 4 times as often than the other two models, which isn't surprising, since Alpaca and Guanaco were trained on data where the LM always gives a full response to the user.

Figure~\ref{fig:llama_empty} shows the same pattern happens with LLaMA as in the earlier experiments: Inputs with longer reference lengths have empty modes more often, but the probability of empty outputs either declines or is relatively unchanged with length.
In Figure~\ref{fig:llama_empty_prob}, Alpaca is the only model that shows a clear trend in terms of empty probability, but even it only changes by a factor of about $2 / 2^{-0.45} \approx 3$.

As mentioned in Chapter~\ref{chapter:modes_theory}, it's commonly thought that the bad mode problem occurs when a task is less-constrained (meaning there are many correct responses).
We don't have access to multiple references, but we can estimate how spread the \emph{model}'s estimate of output distribution is, then see how that relates to mode emptiness.

Figure~\ref{fig:sample_logprobs} shows this relationship, by showing (for each prompt) the logprob of the mode, the average logprob over 5 samples\footnote{The average log-probability of a sample is just the entropy, so this is an estimate of the negative conditional entropy for a prompt. A lower value means that the model's distribution is more spread out for that prompt.}, and whether the modal output is empty.

The Alpaca and Guanaco modes support the ``level-of-constraint'' hypothesis, with the empty modes only occurring when the estimate of the entropy is below some threshold.
LLaMA, on the other hand, has empty modes even for prompts with fairly low entropies (high sample logprobs).
Emitting an EOS token right away is an error for the Alpaca and Guanaco training tasks (conversation/QA), but could actually be correct for the web data LLaMA was trained on.

\subsection{Qualitative analysis: Which prompts lead to empty modes?}
Instruction-following is a much more varied task than either MT or story completion, in terms of how constrained the set of responses is.
In MT, we'd expect that the entropy of the output roughly scales with the input length.
With instruction-following, it's easy to come up with short inputs that have extremely high output or extremely low entropy.

In Chapter~\ref{chapter:modes_theory}, we suggested that the bad mode problem shouldn't be thought of in terms of length, but in terms of the entropy of the set of valid outputs.
This is often related to length, but can easily be decoupled from it.
For example, when prompted with ``Recite the Declaration of Independence.'', Alpaca's global mode is the first 151 tokens of the preamble, ending with ``\ldots to effect their safety and happiness.''
On the other hand ``Write a brief paragraph of the benefits of attending Arizona State University'' is assigned the empty mode.
The latter prompt has many valid outputs shorter than 151 tokens, but the entropy of those valid outputs is clearly \emph{far} higher than the entropy of the set of valid responses to the recitation prompt.

In the previous section, we quantitatively looked at the relationship between open-endedness and empty modal outputs.
Here, we also look at it quantitatively by giving more examples of prompts that lead to empty or non-empty outputs.
Tables~\ref{tab:alpaca_empty_prompts}, \ref{tab:guanaco_empty_prompts}, and \ref{tab:llama_empty_prompts} show 10 of each for each of the three LLaMA-based models, which were randomly sampled to avoid cherry-picking.

For Alpaca and Guanaco, the prompts that have empty modes are generally asking much more open ended questions, which don't have a single clear answer.
Most of the prompts for which LLaMA has a non-empty mode are factoid requests that have a single answer.

\subsection{Qualitative analysis: When the mode is non-empty, is it degenerate?} We observe that when these models' modes are non-empty, they are often high-quality, although we do see problems that weren't present with the models trained on more constrained tasks.

Tables~\ref{tab:alpaca_modes}, \ref{tab:guanaco_modes}, and \ref{tab:llama_modes} show the model modes for the prompts that have non-empty modes.
While some modes are correct responses, there are also degenerate behaviors besides empty outputs.
We'll briefly summarize the issues here.

\paragraph{Alpaca:} Nine of the ten Alpaca outputs are high quality, but Alpaca Prompt I has the output ``\textless{}nooutput\textgreater{}'' which is a tag appearing 28 times in the 52,000 Alpaca training examples.
This exact string was the mode for 46 out of the 1,000 prompts we searched.

\paragraph{Guanaco:} Eight of the modes are high quality (albeit with some factual errors), but Guanaco prompts B and G lead to a mode which is just a phrase from the prompt instead of an answer.
It's hard to measure how often that happens \emph{and} is undesirable, since many of the Dolly prompts require an answer which is part of the prompt.

\paragraph{LLaMA:} Prompts B, E, F and J repeat all or part of the prompt rather than responding.
In general all of these modes are very short, but that isn't surprising since 70\% of LLaMA modes are empty overall.
The prompts that \emph{do} have non-empty modes will tend to be ones that can be answered very concisely.

These modal outputs show that for these LMs, there are multiple types of degenerate outputs.
This is different from the situation with the ROC stories LM or NMT model, where the empty mode seemed to be the only problematic mode.

Addressing this problem will be harder than just conditioning on a single deterministic property like length.
For example, when we re-ran exact search on LLaMA-7B, but blocked the empty output from being produced, 17.2\% of the modal outputs were the string ``\textbackslash{}end\{code\}''.
All the above issues \emph{could} be addressed by heuristic modifications to search, but those heuristics have issues:
\begin{enumerate}
    \item \textbf{Short/Empty outputs.} One option is to maximize logprob per token instead of the overall probability, but repetition of strings from the prompt will have high probability. Also, extremely long strings of one token repeated many times have extremely high logprob per token, since if you've said the word ``hello'' 500 times, the probability of saying it again is very high.
    \item \textbf{Repetition of prompt.} Direct ngram blocking can help, but as we mentioned in many cases repeating parts of the prompt is desirable. One common use case of chatbots is for proofreading, which often requires repetition of large amounts of text.
    \item \textbf{LLaMA's Love of LaTeX.} When specific substrings like ``\textbackslash{}end\{code\}'' are degenerate modes, we can prevent them from being produced, but there might be a large number of such outputs, and of course there's also the problem that they might sometimes be the appropriate response.
    A harder example of this same problem is the ``safety'' outputs that many current chat models produce, where they refuse to answer a request.
    Since the refusal texts are low entropy, they might be modal even when the model thinks there's a 1\% chance it should refuse.
\end{enumerate}

As we said in Chapter~\ref{chapter:modes_theory}, what we really want is to be able to search for the output, $x$, that maximizes:
\begin{equation*}
    \prob \left(x \given[\big] \text{``$x$ is a well-formed response to the prompt''}\right)
\end{equation*}
Doing exact search on that objective is probably intractable, for basically any definition of ``well-formed.''
In the next chapter we investigate how to approximate search of this kind using beam search.

\section{Conclusion}
In this section we looked at the exact modes of NLG models in three settings: Chinese-to-English machine translation, story cloze completion, and instruction-following.
We reproduced the results of \citet{catgotyourtongue}, and extended them to a GPT-2 based LM, as well as three LLaMA-based LMs.

For MT and cloze completion, we found that searching for \emph{length-conditional} modes instead of unconditional modes usually produces high-quality outputs.
This is encouraging, since prior work (See Section~\ref{sec:modestheory_related_work}) had identified that model likelihood often diverges from quality past some point.
These results provide support for the hypothesis we proposed in Chapter~\ref{chapter:modes_theory}: There may be low-quality outputs that are low-probability in absolute value, but due to the high entropy of the set of valid outputs, they still become the mode.
Our finding for MT that the empty output receives a lower probability for longer inputs, but is also \emph{more} likely to be the mode for those inputs is exactly what we would predict based on our reasoning in the last chapter.

We also found that, apart from empty modes, the LLaMA-based language models often have high-quality modes, but sometimes fall into new problems such as repeating their prompt.
In the next chapter, we take advantage of these findings to improve outputs from LLaMA-7B using beam search.

\chapter{Approximate mode search with conditional beam search}\label{chapter:modes_beam}

\section{Introduction}
In the previous two chapters we looked at the bad mode problem both in theory and with several actual models.
Chapter~\ref{chapter:modes_exact} showed that while all these models display the bad mode problem, conditioning on output length could get it under control for the MT model and story completion model.
The LLaMA model derivatives instead displayed a wider variety of issues, such as modes that repeated the prompt or consisted of a single \LaTeX{} fragment, on top of the empty mode issue.

The current way we force LLMs to give acceptable outputs is a combination of supervised finetuning (SFT), and reinforcement learning from human feedback (RLHF).
Both of these methods alter the model's output distribution to get it to output text that is more relevant, factual, non-toxic, etc.
The downside of these methods is that they don't include any incentive for the LM to maintain good coverage of the output space.

This is not just a theoretical problem: GPT-4, the best publicly available LLM at present, appears to put most of its output probability mass on just a few sequences.
When GPT-4 was given the input\footnote{On July 31, 2023, using a temperature of 1.} ``Write a haiku'' 20 times, the outputs had extremely low variation.
For example, 17 of the 20 outputs had the string ``whispers'' in them, and 10 mentioned the moon.

As we argued in Chapter~\ref{chapter:modes_theory}, at least part of what is happening is that our decoding methods don't appropriately handle various types of noise in our training data.
This means that if we want our inadequate decoding methods to produce high-quality results, we need to make sure the model we're using them on is \emph{extremely} unlikely to produce noisy outputs.
This is likely one of the pressures leading to the mode collapse described above.

While in this section we won't be tackling the problem of \emph{sampling} from models, we are still interested in the question of how much we can improve decoding while leaving the model fixed.
In the last chapter, all the models we ran exact MAP inference on showed degenerate modes.
However, the modes were often high-quality, and this was much more likely to be the case when we instead searched for an appropriate \emph{conditional mode}.
Encouraged by these results, we developed a variant of beam search to search for conditional modes, in the same way that ordinary beam search approximates search for the unconditional mode.

The rest of this chapter's contents are as follows.
In Section~\ref{sec:modesbeam_beam}, we derive \emph{conditional beam search}.
Section~\ref{sec:modesbeam_alternatives} discusses the connection between conditional beam search and existing methods for conditional sampling.
In Section~\ref{sec:modesbeam_length} we demonstrate that using conditional beam search while conditioning on length usually leads to higher-likelihood and more grammatical outputs of a desired length than merely constraining standard beam search does.

While the length constraint results support our hypotheses about the bad modes problem and the functionality of conditional beam, they aren't of much use for practical NLG applications.
In order to bridge this gap, in Section~\ref{sec:modesbeam_llama}, we apply conditional beam search to LLaMA-7B, despite the fact that beam search is usually seen as a poor choice for more open ended tasks.
We find that even using just 500 labeled examples, we're able to significantly improve the output compared to ordinary beam search.
These experiments support our repeated claim that while MAP may not be a good decoding method, MAP on the conditional distribution $\pmodel\left( x \given[\big] A(x) = a\right)$ can give good results.

\section{Conditional beam search}\label{sec:modesbeam_beam}
In this section, we'll discuss how to modify beam search to search for \emph{conditionally} high likelihood sequences, rather than the unconditional mode.
Our method is very similar to FUDGE\citep{fudge}, but there are some additional details necessary to support doing beam search instead of sampling.
We discuss the relationship of our method to theirs in Section~\ref{sec:modesbeam_alternatives}, as well as some possible alternative methods.

Beam search is a heuristic method for finding outputs that an autoregressive model assigns a high likelihood to (usually conditional on some other text such as a sentence to be translated).
As we've discussed in the previous two chapters, disagreement between the actual quality of an output and the likelihood our model assigns it is a well-known problem (see the discussion of prior work in Chapter~\ref{chapter:modes_theory}).

In standard beam search, we're doing approximate search for the unconditional mode of the model's output distribution.
Beam search assigns each partial hypotheses of length $t$ a score $S(x_{1:t})$.
The score is just its log-likelihood under $\pmodel$ calculated using the chain rule of probability:\footnote{For a conditional task like MT, the probability would instead be $\pmodel\left(x_i \given[\big] x_{<i}, s \right)$ for a source sequence $s$. For brevity (and to keep notation uniform), we'll leave out the dependence on the conditioning information.}
\begin{equation}\label{eq:beam}
    S(x_{1:t}) = \sum\limits_{i=1}^t \log \left(\pmodel(x_i | x_{<i})\right)
\end{equation}
If $x_{1:t}$ is a complete sequence\footnote{I.e., it ends in the end-of-sentence token}, this score is exactly its log-probability under the model.

Instead of searching for the argmax of $\pmodel(x)$ (the unconditional mode), we want to change this to the argmax of $\pmodel(x | A(x) = a)$ for some attribute $a$ (the conditional mode).
To start with, we can replace the marginal distribution Equation~\ref{eq:beam} with one that conditions on the event that $A(x) = a$:
\begin{equation}\label{eq:cond_beam}
    S'(x_{1:t}, a) = \sum\limits_{i=1}^t \log \left(\pmodel(x_i | x_{<i}, A(x) = a)\right)
\end{equation}
This can be simplified by applying Bayes' rule to any term in the sum:
\begin{equation}\label{eq:bayes}
    \pmodel\left(x_i \given[\big] x_{<i}, A(x) = a\right) = \frac{
        \pmodel\left(A(x) = a \given[\big] x_i, x_{<i}\right) \pmodel\left(x_i \given[\big] x_{<i}\right)
    }{
        \pmodel\left(A(x) = a \given[\big] x_{<i}\right)
    }
\end{equation}
This shows that the denominator in $i$th summand in Equation~\ref{eq:cond_beam} is the same as the numerator in the $i-1$th summand, so part of the sum telescopes:
\begin{align*}
S'(x_{1:t}, a) &= -\log\pmodel\left(A(x) = a\right) + \left(
    \sum\limits_{i=1}^t \log \pmodel\left(x_i \given[\big] x_{<i}\right)
\right) + \log\pmodel\left(A(x) = a \given[\big] x_{1:t}\right)\\
&= -\log\pmodel\left(A(x) = a) + S(x_{1:t}\right) + \log\pmodel\left(A(x) = a \given[\big] x_{1:t}\right)\\
\end{align*}
The first term is just the marginal log-likelihood of the attribute occurring, but it doesn't differ between hypotheses.
Since we only care about the relative value of the score between different hypotheses, we discard it to get our final conditional score:
\begin{align*}
    S_\text{cond}(x_{1:t}, a) &= S(x_{1:t}) + \log\pmodel\left(A(x) = a \given[\big] x_{1:t}\right)\\
\end{align*}
So to do conditional beam search, we just need to add one term to the normal beam search score.
To clear up a possible point of confusion, $\log\pmodel(A(x) = a | x_{1:t}$ is \emph{not} the probability that $x_{1:t}$ has the desired attribute.
It is the probability that the final output will have that attribute, conditional on a prefix of $x_{1:t}$.
The problem is that it's intractable to compute, since we'd need to marginalize over all sequences which have $x_{1:t}$ as their first $t$ tokens.

To avoid this, we will train a classifier to estimate this probability, calling the predicted probability $\pclf\left(A(x) = a \given[\big] x_{1:t}\right)$.
The important thing is that we're trying to specifically predict the probability of a \emph{model output} having the attribute, which means the classifier should be trained on model outputs rather than naturally occuring text.
Substituting in the classifier prediction gives us a score we can actually use for decoding:
\begin{equation}\label{eq:beam_final}
    S_\text{clf}(x_{1:t}, a) = S(x_{1:t}) + \log\pclf\left(A(x) = a \given x_{1:t}\right)\\
\end{equation}
Notice that the probability of the attribute being satisfied doesn't need to be accumulated across timesteps.
This means that $\pclf$ is only used for selecting new hypotheses, but doesn't contribute to the running beam score.
If we didn't take advantage of the telescoping simplification, the value of $\pclf(A(x) = a | x_{<t}$ calculated by marginalizing $\pclf(A(x) = a | x_{1:t})$ (i.e., predicted on step $t$) will be different from the value that was predicted on step $t-1$, since the $\pclf$ may not be temporally consistent.
Using the simplification above lets us avoid accumulating error in the scores.

The last detail to specify is that we only calculate $\log\pclf(A(x) = a | x_{1:t})$ for the $k$ tokens with the highest likelihood, for computational efficiency.\footnote{The reason for doing this is discussed more in Section~\ref{sec:modesbeam_alternatives}.}.
The full method is formally shown in Algorithm~\ref{alg:beam_cond}.

\algnewcommand{\LineComment}[1]{\State \(\triangleright\) #1}
\begin{algorithm}
\caption{Conditional beam search. This searches for an output with a sequence $x$ with a high value of $\pmodel(x, A(x) = a)$ for some target attribute $a$. In order to execute this efficiently one needs to efficiently compute $\pmodel$ and $\pclf$ to avoid recomputation, rather than just fully re-executing them as the pseudocode implies. How this is done will depend on the model architectures, see Section~\ref{sec:modesbeam_alternatives} for discussion.}\label{alg:beam_cond}

 \hspace*{\algorithmicindent}\textbf{Input:} \\
 \hspace*{\algorithmicindent} $\: a$: The target attribute value\\
 \hspace*{\algorithmicindent} $\: V$: Vocabulary\\
 \hspace*{\algorithmicindent} $\: B > 0$: Beam size\\
 \hspace*{\algorithmicindent} $\: L > 0$: Maximum output length\\
 \hspace*{\algorithmicindent} $\: k > 0$: Number of continuations to score for each hypothesis.\\
 \hspace*{\algorithmicindent} $\: \alpha > 0$: Optional attribute weight.\\
 \hspace*{\algorithmicindent}\textbf{Output:} A string in $V*$ with length at most $L$
\begin{algorithmic}[1]
\State $X[b] = \epsilon$ for $b \in {1, \dots, B}$ \qquad\{Initialize all hypotheses to the empty sequence\}
\State $S \gets [0, -\infty, -\infty, \dots, -\infty]$ \qquad\{A length $B$ array of cumulative scores, only $S[0]$ is initially finite.\}
\State $t \gets 1$
\While{Any hypothesis $X[i]$ on the beam is not complete AND $t \le L$}
    \State Init empty $b \times k$ array $U_t$ \qquad\{Unconditional scores\}
    \State Init empty $b \times k$ array $C_t$ \qquad\{Conditional scores)\}
    \State Init empty $b \times k$ array $C'_t$ \qquad\{Weighted conditional scores, optional\}
    \State Init empty $b \times k$ array $W$ \qquad\{Continuations\}
    \For{$b = 1, \dots, B$}
        \If {$t < L$}
            \State $L[w] \gets \log\pmodel (w | x_{1:t} = X[b])$ \{for all $w \in |V|$\}
        \Else
            \State $L[:] \gets -\infty$
            \State $L[\text{\tt </s>}] \gets 0$ \qquad\{If at the max sequence length, all hypotheses are forced to be complete\}
        \EndIf
        \State $W[b, :] \gets$ words with top $k$ scores in $L$
        \For {$i = 1, \dots, k$}\qquad{}
            \State $w \gets W[b, i]$
            \State $U_t[b, i] \gets S[b] + L[w]$
            \State $C_t[b, i] \gets U[b, :] + \log\pclf(a | x_{t+1} = w, x_{1:t} = X[b])$
            \State $C'_t[b, i] \gets U[b, :] + \alpha\log\pclf(a | x_{t+1} = w, x_{1:t} = X[b])$
        \EndFor
    \EndFor
    \State \{We select new beam elements using $C'_t$, but update $S$ using $U_t$.\}
    \For{$b = 1, \dots, B$}
        \State $b_\text{prev}, i \gets$ $b$-th largest pair of indices into $C'_t$
        \State $S[b,i] \gets U_t[b_\text{prev}, i]$
        \State $X[b] \gets \mathrm{concatenate}\left(X[b_\text{prev}], W[b_\text{prev}, i]\right)$
    \EndFor
\EndWhile
\If {Attribute is deterministic}
    \State $b_\text{best} = 1$
\Else
    \State $b_\text{best} \gets \argmax\limits_{i} C_t[i]$\qquad\{If the classifier is uncertain about the value of the attribute even for a complete output, we take that into account when selecting the output.\}
\EndIf
\State \textbf{Return} $X[b_\text{best}]$
\end{algorithmic}
\end{algorithm}

\section{Related work on conditional generation}

The main goal of this chapter is to start bridging the gap between exact search for conditional modes (tractable, but not efficient), and methods that are actually efficient enough to run in practice.
In the last section, we started with ordinary beam search, and derived a version of it that can be used to instead search for conditional modes.

Our goal is to do approximate search on a conditional version of an LM or MT models' unconditional distribution, $\pmodel(x)$.
Because of that, we'll focus on two sampling methods that can be seen as approximately sampling from such a conditional distribution (FUDGE in particular explicitly tries to do so).
We discussed other controllable generations that aren't concerned with conditional distributions in Section~\ref{sec:relatedwork_generation}

\subsection{Two ways to estimate $\pmodel\left(a | x\right)$}\label{sec:modesbeam_alternatives}
 Here we'll look at two approaches to estimating $\prob(a | x_{1:t})$ from partial hypotheses, the methods that introduced them, and the tradeoffs they make.
\paragraph{FUDGE~\citep{fudge}.} FUDGE is a method for conditional sampling, specifically approximately sampling from $\pmodel(x | A(x) = a)$. They train classifiers to predict $\pmodel(a | \cdot)$ conditional on partial hypotheses, just like we need for Equation~\ref{eq:beam_final}.
They apply them to conditional sampling rhyming couplet generation, topic constrained generation, and formality constrained machine translation.

Their approach (and ours) requires calculating $\pclf(A(x) = a \given x_t, x_{<t})$ $x_t$ as an \emph{input} to the classifier.
The problem is that we need to do this for \emph{every} possible continuation token, so the attribute classifier needs to be very lightweight.
On top of that, both FUDGE and our method only apply the classifier to the top $k$ possible next tokens, in order to further reduce compute.

An alternative would be for the classifier to take as input just $x_{<t}$, and then output a $|V| \times |\mathcal{A}|$ matrix that simultaneously predicts the $|V|$ categorical distributions $\pmodel(A(x) = a | x_t, x_{<t})$ (one for each value of $x_{t}$).
This would allow us to avoid the ``top-$k$'' restriction, but would also reduce the capacity of the classifier since it would do computation per possible continuation.

\paragraph{GeDi~\citep{krause2020gedi}.} GeDi (Generative-Discriminators) instead trains class-conditional LMs (CC-LMs) to predict $\prob(x_t | A(x) = a)$ directly.
To use them to modify sampling from an unconditional LM, they use Bayes' rule to estimate $\prob(A(x) = a | x_{1:t})$, which is then applied in the same way as shown in Equation~\ref{eq:bayes}.
In their experiments they train their CC-LMs on labeled datasets rather than model outputs, but to estimate $\pmodel(x_t | A(x) = a)$ you could just train them on model outputs instead.

The upside of GeDi compared to directly predicting $\pmodel(A(x) = a | x)$ is that they don't need to restrict their attribute prediction to the $k$ tokens with the highest marginal scores.
The downside is that training CC-LMs should intuitively be much harder than training classifiers, since they need to learn the entire text distribution.

Imagine training a trivial attribute that was just ``Is the first word of the output 'The'.''
An attribute classifier would only need to predict the marginal probability of seeing ``The'' at the first token, then output either 0 or 1 on all future timesteps depending on what token was output.
A CC-LM would instead try to learn the conditional distribution of all the text following the first token, which is both difficult and completely irrelevant to the classification task.

\begin{table}[h]
    \centering
    \caption{Tradeoffs between the two alternatives discussed for conditional generation (generation from $\pmodel(x | A(x) = x)$).}\label{tab:clf_tradeoffs}
    \begin{tabular}{p{0.1\textwidth}lp{0.2\textwidth}p{0.3\textwidth}}
        \toprule
        Auxiliary model & Predicts & Pros & Cons\\
        \midrule
        Classifier & $\pmodel(A(x) = a | x_t, x_{<t})$ & Doesn't need to learn extraneous information. & Expensive to apply to possible continuations, so only $k$ continuations are considered.\\
        Class-Conditional LM & $\pmodel(x_t \given A(x) = a, x_{<t})$ & Can consider all possible continuations. & Needs to learn next-token distribution in addition to discriminative information.\\
    \bottomrule
    \end{tabular}
\end{table}
\paragraph{Evaluation of tradeoffs} Table~\ref{tab:clf_tradeoffs} summarizes the pros and cons of the two methods.
We chose to use the classifier-based approach since the downside of needing to learn the distribution of text in addition to discriminitave information seemed too large.
That's just an intuitive justification, not an empirical fact, so using conditional beam search with CC-LMs should be tested in future work.

\subsection{Value guided beam search}
Another method for controllable generation which doesn't attempt to do \emph{conditional} generation is Value-Guided Beam Search (VGBS, \citealp{vgbs}).
Our score function in Equation~\ref{eq:beam_final}, is extremely similar in form to the score function used by VGBS.
Specifically, they use:
\begin{equation*}
    S_v(x_{1:t}) = (1 - \alpha) S(x_{1:t}) + \alpha V(x_{1:t})
\end{equation*}
Where $V(x_{1:t})$ predicts the value of a continuation (BLEU score in their experiments).

The differences between VGBS and conditional beam search are due to a difference in goals: Our goal is to find sequences which our LM assigns a high conditional probability to, whereas VGBS is just interested in finding the highest reward sequence.
This difference causes us to select the final best hypothesis differently, and train classifiers differently.

Since VGBS wants to find a high value sequence, they include the weight $\alpha$ in the final score which is used to choose what sequence to output.
They also train their value function on data produced using beam search, since since the ideal value network will be one which can guide beam search specifically to a high scoring output.

On the other hand, we want to actually estimate the quantity $\pmodel (x | A(x) = x)$.
This leads us to use model samples in training, otherwise the classifier will not be trying to predict the conditional under $\pmodel$.
It also means that we take the output from the final beam which is the highest probability under that conditional distribution.
Because of that, when we do use a weight $\alpha$ we only use it during search, and not for final hypothesis selection.

The other difference is that in conditional beam search we may evict a complete sequence which was already found.
This isn't very relevant in standard beam search, since log-likelihoods only decrease as generation continues.
However, using an attribute classifier, it can be the case that the estimate of $\pmodel (a | x_{1:t})$ changes dramatically as a sequence is produced.
A better method would be to keep \emph{all} complete sequences encountered during search, but this isn't usually done, and we leave testing it to future work.\footnote{For regular beam search this would lead to the output being empty whenever the empty sequence is the modal output. This is because the probability of all possible first tokens is calculated, including the EOS token. So ordinary beam search often ``finds'' the global mode, but just doesn't keep track of it and loses it again! This is a manifestation of \citet{meister2020ifbeamsearch}'s point that beam search is probably doing something other than actually trying to find high likelihood sequences.}

\subsection{Monte-Carlo tree search with a value network}
\citet{liu2023making} propose using the value network which is produced used for PPO-based RLHF to perform Monte-Carlo Tree Search (MCTS) at inference time.
In terms of the objective being optimized, this is essentially identical to VGBS, but much more computationally expensive, and complex to implement.
However, if one is trying to maximize the estimated value of outputs (rather than trying to find high-likelihood sequences), this may be a promising approach.

\subsection{COLD decoding}
\citet{cold_decoding} propose COLD, a decoding method which allows one to generate sequences which optimize a given energy function.
Similar to VGBS and the method of \citet{liu2023making}, COLD does not search sequences which are assigned high-likelihood by an NLG model, but instead finds one which strongly satisfy a soft constraint.
As these constraints are soft and being optimized, a sufficiently low energy would be preferred, regardless of the likelihood the underlying LM assigned it.

\subsection{Neurologic decoding}
Neurologic A*-esque (NA) decoding~\citep{lu2021neurologic} is a class of methods for constrained decoding.
NA allows one to specify lexical constraints on the generation which will be produced.
Similar to conditional beam search, NA needs to re-rank tokens early in generation based on the eventual content of the final generation.
While we approach this by training classifiers, NA uses a variety of lookahead heuristics involving generating partial continuations of the input.

In our setting, this would not be feasible as rollouts using sampling, greedy decoding, or beam search will generally find no sequences satisfying the constraint.
Further, for some of the constraints we consider (long sequence length), rollouts would need to be as long as 50 tokens long.
Finally, we consider generation towards soft targets, where it is not possible to deterministically label a sequence as satisfying a constraint or not.

\section{Experiments: Length controlled generation}\label{sec:modesbeam_length}
In this section and the next we discuss our experiments with conditional beam search.
Our first experiments are on length-constrained generation for language modeling and MT.
Specifically, our goal is to generate a sequence of exactly length $L$ that has a high score under the relevant model.

Our experiments verify that our method is able to decode outputs of a given length with higher likelihood than those found by beam search.
This shows that conditional beam search can find high-probability sequences which satisfy a constraint.
We also find that, qualitatively, the sequences found using conditional beam search are better than the ones found by just constraining beam search to output at the target length.

\subsection{Models and data}
Our NLG models are the same ones used to find length-conditional exact modes in Chapter~\ref{chapter:modes_exact}: The MarianMT Chinese-English model for MT, and the ROC stories finetuned GPT-2 model for LM.
We train our classifiers using sequences sampled from these models.
In the case of the GPT-2 model we can do this in a data free way (simply generating each text from scratch), but for the MarianMT model we sample translations from the model conditional on sources from the \texttt{news-commentary-v12-zh-en} training data~\citep{wmt17}.

\subsection{Conditioning target: Length}
Because the number of possible output lengths is very high, we train a classifier to predict the length remaining rather than the absolute length of a sequence.
We also reduce the number of categories in the classification problem by bucketing the lengths into coarser groups as the length increases.
Specifically, lengths 0-16 each have a unique class, lengths 17-32 are split into 4 groups, lengths 32-64 are split into two groups, then all lengths 65 and higher are assigned to a single group (24 classes in total).

\subsection{Classifier architecture and training}
For both models, we train a classifier that takes the decoder's hidden states as input.\footnote{The decoder in the case of the LM is the entire model, while the MT model has a separate encoder and decoder.}
This section describes the exact architecture, and how to use it for conditional beam search.
The specific hyperparameters used are given in Appendix~\ref{chapter:beam_hparams_appendix}.

The tricky thing about the classifier is that we want it to be able to make predictions about the class of possible continuation tokens, while not needing to run the full decoder model on each of those continuations to make a hidden state.
To make that possible, the classifier uses the model hidden states at time $t$ (and earlier) to make a classification prediction for time $t + 1$.

We'll formalize the prediction process here.
Let $\bm{h}_{1:T}^{(\ell)}$ be the layer $\ell$ hidden states of the decoder, with the decoder having $M$ total layers and an embedding dimension of $d_\text{model}$.
Suppose we want to predict the final label for an output sequence with a prefix of $x_{1:t}$, and a candidate next token $x_{t+1}$.
In order to make the computation for each continuation lightweight, we frame it in terms of the word embedding of that token\footnote{These embeddings come from the input embedding table from the underlying decoder model.}, $w$.
For some dimension sizes $d_\text{clf}$ and $d_\text{out}$, the classifier output is computed as follows:
\begin{alignat*}{2}
    \bm{h}_{1:t}^{(\text{stacked})} &= \mathrm{Concat} \left([\bm{h}_{1:t}^{(0)}; \bm{h}_{1:t}^{(1)}; \dots; \bm{h}_{1:t}^{(M)}]\right) \qquad && \text{// Concatenate all hidden states layerwise}\\
    \bm{h}_{1:t}^{(\text{in})} &= \mathrm{Linear}\left(\bm{h}_{1:t}^{(\text{stacked})}\right) && \text{// Project down to input dimension: $\mathbb{R}^{T \times (M d_\text{model})} \to \mathbb{R}^{t \times d_\text{clf}}$}\\
    \bm{h}_{1:t}^{(\text{out})} &= \mathrm{Transformer}(\bm{h}_{1:t}^{(\text{in})}) && \text{// $\mathbb{R}^{t \times d_\text{clf}} \to \mathbb{R}^{t \times d_\text{out}}$}\\
    \bm{c}_{t+1} &= \mathrm{Concat}([
            h_t^{(\text{out})},
            w
    ]) && \text{// Concatenate output for time $t$ to token emb.}\\
    \mathrm{Logits}_t &= \mathrm{MLP}(\bm{c}_{t+1}) && \text{// Apply MLP without output dimension $|\mathcal{A}|$}
\end{alignat*}
At training time, the classifier is only passed the tokens that actually occur in the training example, so this is done in parallel for every position in the sequence.
At inference time, it needs to evaluate $k$ candidate continuation tokens (See Algorithm~\ref{alg:beam_cond}).
Since the the transformer output from time $t$ is shared across all evaluations of tokens for time $t+1$, it only needs to be run once for each position, while the MLP is run $k$ times.\footnote{To avoid possible confusion, there are two transformers: the one in the underlying NLG model, and the one being used for classification. Both of them only need to make one forward pass per sequence during training, and one forward pass per token during inference.}

The reason to include the transformer instead of just the MLP is that it allows the classifier to use as many of the NLG model's hidden states as possible, instead of just the hidden states from time $t$.
This is similar to the architecture used by FUDGE, but using a transformer instead of a LSTM.

\subsection{Beam search details}
We use beam sizes of 5 and 20 for our experiments, and always evaluate $k=100$ candidate next tokens for conditional beam search.
We use $\alpha = 1$ as defined in Algorithm~\ref{alg:beam_cond}, so this is exactly the theoretical version of conditional beam search derived in Section~\ref{sec:modesbeam_beam}.

\subsection{Results}\label{sec:modesbeam_length_results}
\begin{figure}
    \centering
    \begin{subfigure}{0.45\textwidth}
        \includegraphics[width=\linewidth]{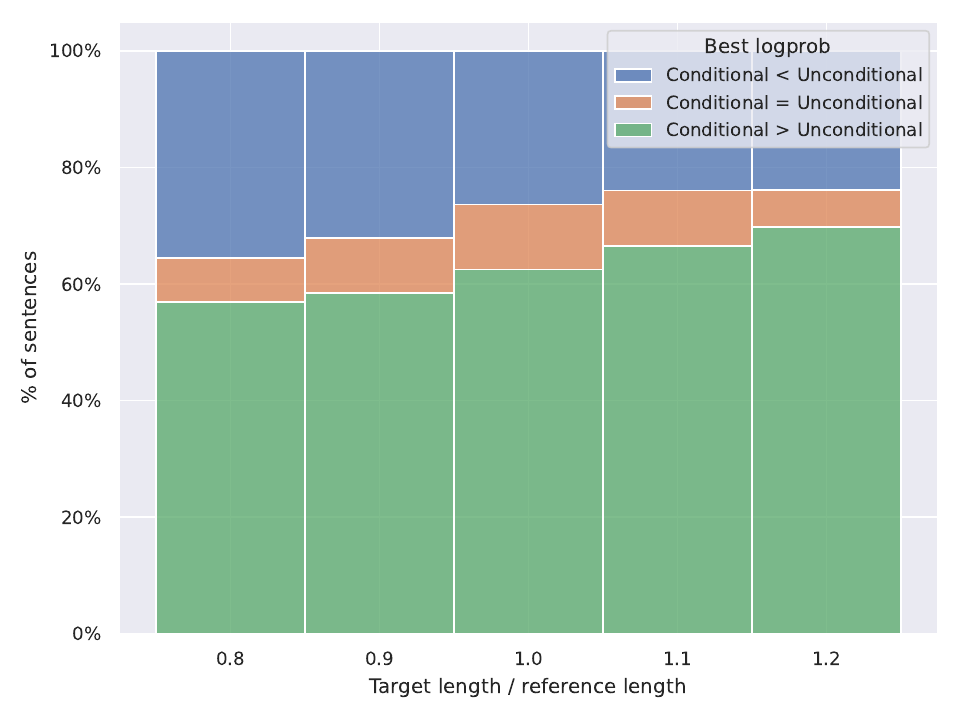}
        \caption{Beam size 5.}\label{fig:marian_beam5}
    \end{subfigure}
    \hfill
    \begin{subfigure}{0.45\textwidth}
        \includegraphics[width=\linewidth]{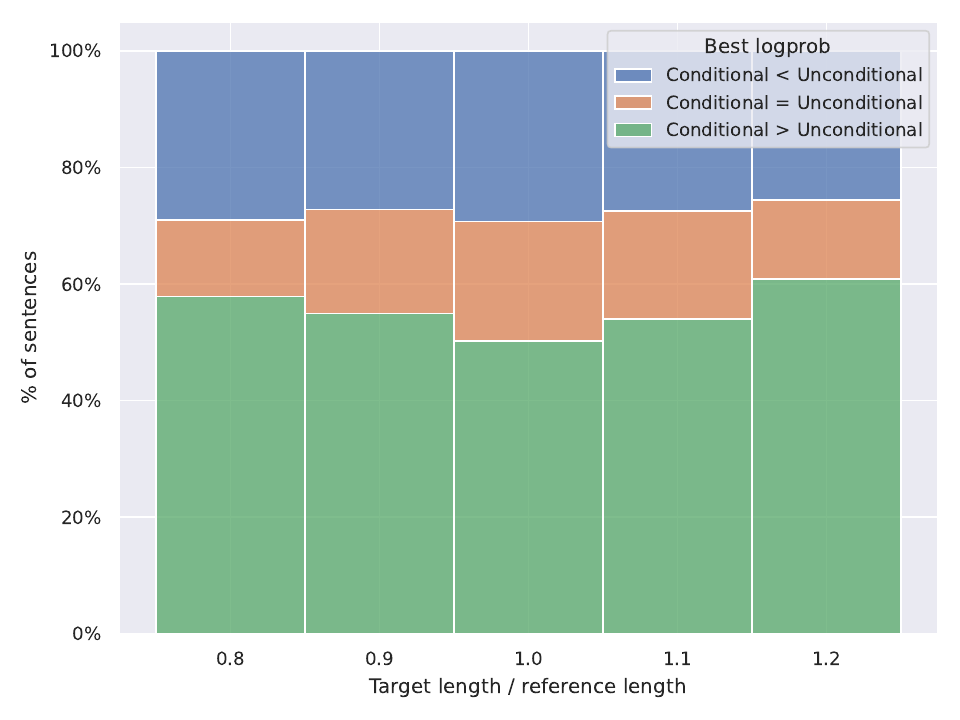}
        \caption{Beam size 20.}\label{fig:marian_beam20}
    \end{subfigure}
    \caption{A comparison of classifier-guided conditional beam search and beam search when generating outputs that must be a certain length, using the MarianMT Zh-En translation model on the WMT17 dev. set. Each bar shows how often each method finds the higher log-probability output for a given length constraint. The length constraints are relative to the reference sentence, so for a 20 token reference constraints of 16, 18, 20, 22, and 24 tokens were used.}\label{fig:marian_beam_scores}
\end{figure}
    
\begin{figure}    
    \centering
    \begin{subfigure}{0.45\textwidth}
        \includegraphics[width=\linewidth]{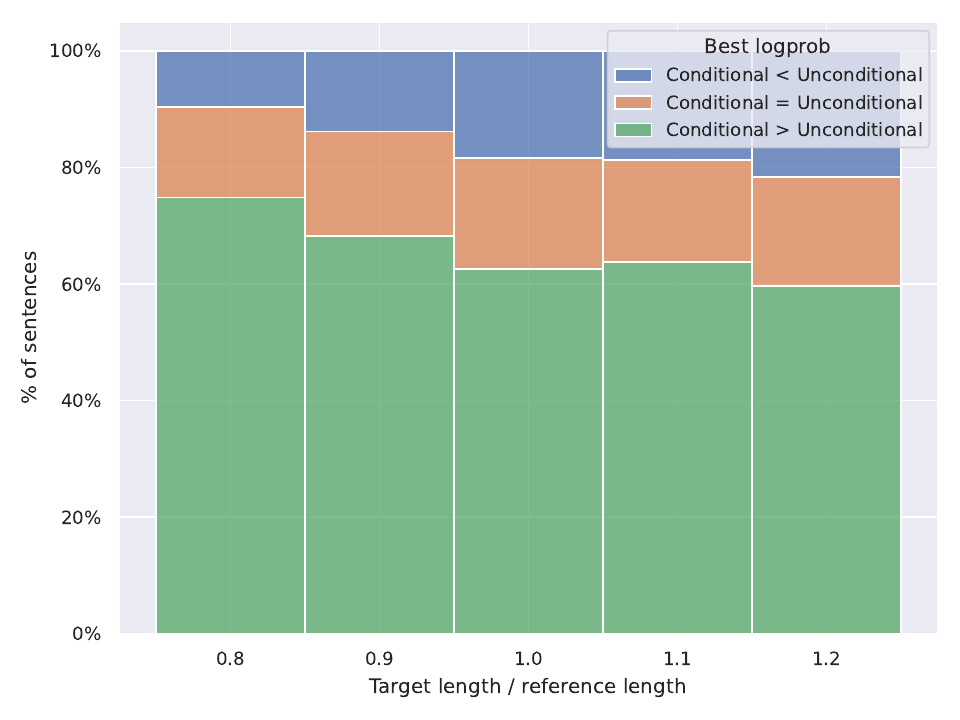}
        \caption{Beam size 5.}\label{fig:roc_beam5}
    \end{subfigure}
    \hfill
    \begin{subfigure}{0.45\textwidth}
        \includegraphics[width=\linewidth]{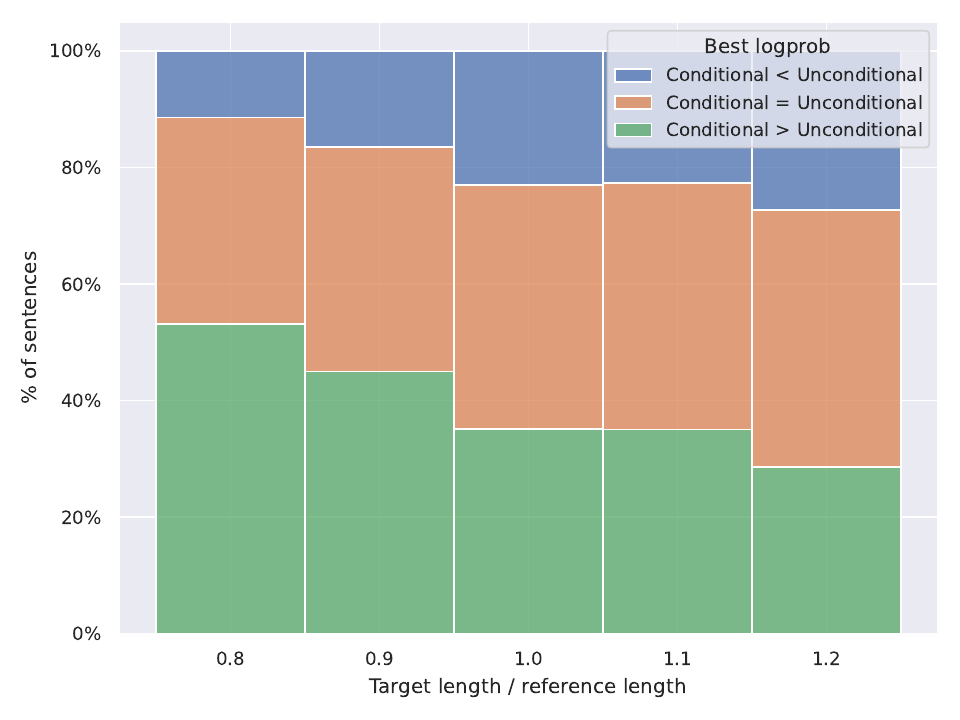}
        \caption{Beam size 20.}\label{fig:roc_beam20}
    \end{subfigure}
    
    \caption{A comparison of classifier-guided conditional beam search and beam search when generating outputs that must be a certain length, using our finetuned GPT-2-345M model on the ROC Stories dev. set. See Figure~\ref{fig:marian_beam_scores} for explanation.}
    \label{fig:roc_beam_scores}
\end{figure}

Our main finding from our length-control experiments is that conditional beam search typically manages to find higher probability outputs of a given length than beam search does.
In these experiments, both beam search and conditional beam search are constrained to output the EOS token at the target length, and never produce the EOS token prior to that point.
We do this by setting the probability of the EOS token to 0 and renormalizing the log-probabilities.\footnote{An alternative would be to just take the top-k tokens that aren't EOS, which would lead to different scores being accumulated across timesteps and could affect the final output. We opted to use the renormalization method since we know that $\pmodel(x_i = \mathtt{</s>} | |x| = L) = 0$ when $i < L$.}

Figures~\ref{fig:marian_beam_scores} and \ref{fig:roc_beam_scores} summarize our findings when controlling the output length.
These plots show how often each method finds the higher likelihood output for a given input and a target length that is rounded from a multiple of the corresponding reference length.
For MT, the source/reference pairs come from the WMT17 dev set.
For the ROC Stories LM, the inputs are the first four sentences of examples from the dev set, and the references are the correct fifth sentence.

Overall, conditional beam search outperforms unconditional beam search in terms of finding high-likelihood sequences satisfying the length constraint.
The closest unconditional beam search comes to being better is at a length ratio of 1.2 on ROC Stories with a beam size of 20, where conditional beam search only finds the better output about 1 percentage point as often.
Unsurprisingly, there is less of a gap between unconditional and conditional search with a beam size of 20.\footnote{If the beam size is increased sufficiently, both methods will eventually find the conditional mode of length $L$.}

One interesting difference between the MT results (Figure~\ref{fig:marian_beam_scores}) and LM results (Figure~\ref{fig:roc_beam_scores}) is that the relationship between improvement and length is inverted.
For a beam size of 5, unconditional search beats conditional search more often at higher lengths for MT, but less often at higher lengths for ROC stories.
A priori we might have expected a trend more like the one shown in Figure~\ref{fig:marian_beam20}, where the benefit of conditional search is the lowest at exactly the reference length, but higher for ``weirder'' lengths which aren't equal to the reference.

\subsubsection{Qualitative properties of outputs}
\begin{CJK*}{UTF8}{gbsn}
\begin{table}[h]
    \centering
    \caption{Selected decoding outputs from MarianMT Zh-En to compare conditional and unconditional beam search (beam size 5). Conditional beam search more consistently leads to grammatical outputs.}\label{tab:marian_beams_selected}
    \begin{tabular}{llp{0.6\textwidth}}
         Type & $\log P(y | x)$ & Text\\
         \toprule
         Input & - & ▁该仓库当初就不应建造在距离住宅楼那么近的地方。 \\
Reference (20 tokens) & - & The storage depot should never have been built so close to residential buildings in the first place. \\
Unconditional (16 tokens) & -6.04 & The warehouse should not have been built so close to the residential building.\\
Unconditional (18 tokens) & -10.85 & The warehouse should not have been built so close to the residential building at first.\\
Unconditional (20 tokens) & -21.73 & The warehouse should not have been built so close to the residential building at the time of the\\
Unconditional (22 tokens) & -14.53 & The warehouse should not have been built so close to the residential building at the time of the incident.\\
Unconditional (24 tokens) & -25.18 & The warehouse should not have been built so close to the residential building at the time of the incident..\\
Conditional (16 tokens) & -6.04 & The warehouse should not have been built so close to the residential building.\\
Conditional (18 tokens) & -9.12 & The warehouse should not have been built in such close proximity to the residential building.\\
Conditional (20 tokens) & -11.66 & The warehouse should not have been built in such close proximity to residential buildings at the time.\\
Conditional (22 tokens) & -20.40 & The warehouse should not have been built in the vicinity of the residential building at the time of writing.\\
Conditional (24 tokens) & -22.71 & The warehouse should not have been built in the vicinity of the residential building in the first place of its construction.\\
\midrule
Input & - & ▁例如,生吃10个土豆可导致毒性反应。 \\
Reference (16 tokens) & - & For example, eating 10 raw potatoes can result in a toxic response. \\
Unconditional (12 tokens) & -15.79 & For example, a raw diet of 10 potatoes can\\
Unconditional (14 tokens) & -14.62 & For example, eating 10 potatoes raw can lead to toxic effects\\
Unconditional (16 tokens) & -8.73 & For example, raw eating of 10 potatoes can lead to toxic effects.\\
Unconditional (17 tokens) & -8.88 & For example, a raw diet of 10 potatoes can lead to toxic effects.\\
Unconditional (19 tokens) & -19.39 & For example, a raw diet of 10 potatoes can lead to toxic effects..\\
Conditional (12 tokens) & -10.99 & For example, 10 raw potatoes can cause toxicity.\\
Conditional (14 tokens) & -8.57 & For example, eating 10 potatoes raw can cause toxic effects.\\
Conditional (16 tokens) & -8.73 & For example, raw eating of 10 potatoes can lead to toxic effects.\\
Conditional (17 tokens) & -8.88 & For example, a raw diet of 10 potatoes can lead to toxic effects.\\
Conditional (19 tokens) & -20.07 & For example, a raw diet of 10 potatoes can lead to toxic effects of toxicity.\\
\bottomrule

    \end{tabular}
\end{table}
\end{CJK*}
\begin{table}[h]
    \centering
    \caption{Selected decoding outputs from ROC stories finetuned GPT2-345M to compare conditional and unconditional beam search (beam size 5). Conditional beam search more consistently leads to grammatical outputs.}\label{tab:roc_beams_selected}
    \begin{tabular}{llp{0.6\textwidth}}
         Type & $\log P(y | x)$ & Text\\
         \toprule
         Input & - & Kelly hated math class and struggled to learn the concepts. She struggled a lot with the work and often sought help from teachers. She worked very hard and it paid off with good grades. She was entering college in the fall. \\
Reference (7 tokens) & - & Kelly graduated with good grades. \\
Unconditional (5 tokens) & -14.33 &  She was so excited\\
Unconditional (6 tokens) & -18.39 &  When she got to college\\
Unconditional (7 tokens) & -16.35 &  She was so excited to start\\
Unconditional (8 tokens) & -20.58 &  When she got to college she was\\
Conditional (5 tokens) & -10.09 &  Kelly got accepted.\\
Conditional (6 tokens) & -7.63 &  Kelly graduated with honors.\\
Conditional (7 tokens) & -10.16 &  Kelly graduated with a B.\\
Conditional (8 tokens) & -10.01 &  Kelly graduated with honors in math.\\
\midrule
Input & - & Yesterday I played the Powerball game. I picked my numbers from our family's bible. I purchased my tickets from a reputable online lottery agent. I prayed nervously as the winning numbers were drawn. \\
Reference (6 tokens) & - & I didn't win. \\
Unconditional (4 tokens) & -16.02 &  I won the\\
Unconditional (5 tokens) & -14.07 &  I won the lottery\\
Unconditional (6 tokens) & -13.69 &  I won the Powerball\\
Unconditional (7 tokens) & -18.21 &  When the numbers were called,\\
Conditional (4 tokens) & -7.22 &  I won!\\
Conditional (5 tokens) & -11.61 &  I was ecstatic.\\
Conditional (6 tokens) & -7.24 &  I won the lottery!\\
Conditional (7 tokens) & -7.13 &  I won the jackpot!\\
\bottomrule

    \end{tabular}
\end{table}
A consistent pattern in the outputs is that conditional beam search finds grammatical outputs of the requested length, while unconditional beam search does not.
Tables~\ref{tab:marian_beams_selected} and \ref{tab:roc_beams_selected} show several examples of this issue.\footnote{To avoid cherry-picking, many more randomly selected outputs are shown in Tables~\ref{tab:marian_beam_samples} and \ref{tab:roc_beam_samples}, that also consistently show the same pattern.}
While conditional beam search finds different outputs for each length such that the final length is correct, unconditional beam search often outputs truncated sequences.

The reason that unconditional beam search is failing is that it can't plan ahead.
If the target length is $L$, ordinary beam search might have no examples on the beam at step $L-2$ that could lead to a fluent ending within two tokens.
The generation models do ``know'' that it's the wrong time to output EOS, so they assign a low-probability to the overall sequence, but that isn't sufficient to find a good output if there aren't any good partial hypotheses on the beam.
This also partially explains why conditional beam search is able to find higher scoring sequences than unconditional beam search.

\section{Experiments: Control of completeness and relevance with LLaMA}\label{sec:modesbeam_llama}
In the last section, we looked at training classifiers to control output length.
This at least directly address the empty-mode problem, but is not of much practical interest.
However, as we saw in Chapter~\ref{chapter:modes_exact}, the 7B parameter LLaMA~\citet{llama} and its derivatives\footnote{No pun intended.} suffer from problems other than brevity.
Some of the problems that exact modes of LLaMA-7B on the instruction-following data displayed were:
\begin{enumerate}
    \item The mode was empty 70.7\% of the time;
    \item Many outputs exactly repeated the prompt;
    \item When the empty output was blocked, the next most likely output was often a \LaTeX fragment;
\end{enumerate}
However, when the output distribution has low enough entropy, the modal output was sometimes a response to the prompt.
The question we investigate in this section is whether we can condition away these problems and get high-quality responses from LLaMA-7B despite it not having been trained for anything other than pure language modeling.

We define two attributes to use for conditioning: ``Completeness'' and ``Irrelevance'', defined fully in Section~\ref{sec:modesbeam_attributes}.
We want outputs with high completeness, but low irrelevance.
With only 200 labeled training examples for each of these categories, we are able to steer LLaMA-7B to produce beam search outputs that display the problems above much less frequently.
This is extremely promising for our goal of improving generation without needing to destroy the information stored in the underlying language model.

\subsection{Model and data}
These experiments use the LLaMA-7B model, which has only been trained for language modeling, not instruction following.
We chose to use this over the finetuned models such as Alpaca~\citep{alpaca}, since those models have been directly trained to respond to prompts, while we're interested in trying to get high-quality behavior out of a generic LM.
Rather than using LLaMA in its 16-bit precision, we use a 4-bit quantized version (See Section~\ref{sec:modesbeam_llama_clf} for details).

We used 500 inputs from the Alpaca~\cite{alpaca} as inputs for classifier training.
For each input, we generated a completion using the ``beam search'' prompt format shown in Appendix~\ref{chapter:prompts_appendix}.
450 of these are used for training the classifiers, and 50 are used as a validation set for early stopping.
These completions were generated using beam search, instead of sampling.
We discuss the reason for making that choice in Section~\ref{sec:modesbeam_concessions}.

For testing, we use inputs from the \texttt{databricks-dolly-15k} dataset~\citep{dolly15k}.
We switch datasets for evaluation to make it less likely that any improvements we see are just due to the classifier exploiting idiosyncracies of the training data.
In order to make our output length limit not too constrictive, we use the 248 prompts from the dataset which led to a prompt length of under 100 tokens.

\subsection{Conditioning targets: Completeness and Irrelevance}\label{sec:modesbeam_attributes}
We train two separate classifiers for the following two attributes:
\begin{itemize}
    \item Completeness: The degree to which an output attempts to fully answer the prompt. For the question ``Who was the first president of the United States?'', ``Abraham Lincoln'' would be a complete answer despite being incorrect. ``I don't know'' would be incomplete, as would an empty output.
    \item Irrelevance: The amount of unnecessary text in the output. An empty output has minimal irrelevance by definition, while ``\textbackslash{}text\{George Washington\}'' would be somewhat irrelevant due to the unnecessary TeX markup.
\end{itemize}
By controlling for both of these attributes at the same time, we are implicitly assuming that the events ``$x$ is complete'' and ``$x$ is relevant'' are independent conditional on a prefix of $x$.
That is definitely untrue, so this is an approximation.

For training data, we labeled 500 outputs from LLaMA-7B for completeness, while the GPT-4 was used to predict the irrelevance. (See Appendix~\ref{chapter:prompts_appendix} for the exact prompt used).
The labeling was done on a scale of 1-5, but the final classification task was binary classification.
All empty outputs were automatically marked as a 1 for Completeness, and a 1 for Irrelevance.
The class balances were:
\begin{itemize}
    \item Completeness: 70\% incomplete, 30\% complete;
    \item Irrelevance: 13.8\% irrelevant, 86.2\% relevant;
\end{itemize}
See Section~\ref{sec:modesbeam_concessions} for discussion of concessions we had to make in terms of training data quality.

\subsection{Classifier architectures}\label{sec:modesbeam_llama_clf}
The classifiers we use in this section consist of taking the LLaMA-7B model, adding a linear classifier on top of its final layer hidden state, and partially finetuning the model.
The reason for using this much more heavyweight approach than in the last section is that we only have 500 labeled training examples, so leveraging the pretrained information available in LLaMA\footnote{Both about the distribution of text and the meaning of its hidden states.} is critical.

However, if we made classifiers exactly as stated above, the memory for the parameters alone would take:
\begin{equation*}
(7\text{ billion}) \times (16\text{ bits}) \times (3\text{ models}) \approx 42 \text{ GB}
\end{equation*}
We ran these experiments on a single 24 GB GPU, and the total above doesn't even include the model activations, so we use several methods to reduce the footprint.

Our method is based on QLoRA~\citep{qlora}, a parameter-efficient fine-tuning method for large models consisting of the use of 4-bit blockwise quantization, and LoRA~\citep{lora}.
We'll briefly go over each of those methods here as background, then discuss what modifications we made.
For specific hyperparamters, see Appendix~\ref{chapter:beam_hparams_appendix}.

\paragraph{4-bit blockwise quantization.} We use the blockwise quantization method introduced by \citet{qlora}, but with the AF4-64 code\footnote{See: \url{https://github.com/davisyoshida/abnormal-floats}.} instead of their NF4 code.
Blockwise quantization of a transformer replaces each floating point weight matrix with a matrix of integers, although with a scaling value per block (in our case each block consists of 64 values).
When the weight is needed for a quantization, the values are turned back into 16-bit floating point values by taking the relevant value in the quantization code, and multiplying it by the scaling value for the block.
In the case of LLaMA-7B, this reduces the GPU memory needed to store the weights from 13.5 GB to 3.7 GB.

\paragraph{LoRA \citep{lora}.} Low-rank Adaptation (LoRA) is a method for finetuning a large model without needing to update its weights, and in particular without needing to maintain optimizer states for those weights.
For each weight matrix $W \in\mathbb{R}^{m \times n}$ to be trained, new weights $A \in \mathbb{R}^{r \times n}$ and $B \in \mathbb{R}^{m \times r}$ are introduced.
Then, each multiplication $Wx$ in the network is replaced by the computation $Wx + BAx$.
$B$ is initialized to be the zero matrix, so this initially makes no difference to the model's output.
The combined set of $A$'s and $B$'s for the whole model is often also referred to as a LoRA.\footnote{Presumably short for low-rank \emph{adapter} instead of \emph{adaptation} in this case.}

Instead of training $W$, it is left frozen and $A$ and $B$ are trained instead.
Since $r \ll \min (m, n)$, this leads to far fewer parameters being trained, and is equivalent to learning a rank $r$ update.
Readers should see \citet{lora} for more information.

\paragraph{Batching LoRAs.} With limited GPU memory, the idea of loading more than one 7 billion parameter model into memory sounds too expensive to be practical. However, if each one of those models is just a LoRA finetune of the same underlying base model, we can load a single copy of the quantized base-model, and the much smaller LoRA parameters for each finetuning.
On top of the memory savings, we can vectorize forward passes by batching the $A$ and $B$ matrices from each LoRA together.
In our case we'll have three LoRAs: One for each of the two classifiers, and an additional one that has $A = B = \bm{0}$ to retain the behavior of the original LM\footnote{This is inefficient since computing $ABx$ for the zeroed out parameters is a waste of computation, but it lets us simplify the implementation.}.
So, for each ``model'' a matmul that originally mapped from $\mathbb{R}^{m}$ to $\mathbb{R}^{n}$ is batched to map from $\mathbb{R}^{3 \times m}$ to $\mathbb{R}^{3 \times n}$.

This is implemented without any changes to the model code, since we use a custom JAX~\citep{jax} transform for applying LoRA to models that were written without LoRA in mind\footnote{The transformation is available at \url{https://github.com/davisyoshida/lorax/tree/master/lorax}, and should be compatible with most JAX models.}.
We then use JAX's \texttt{vmap} transformation to turn a function that expects a single LoRA into one that takes multiple LoRAs and produces multiple outputs from a single input.

\paragraph{Shared trunk.} This would still use way too much memory, since we'd still need to store three full copies of the KV-cache during inference. 
For beam search, this is also multiplied by the beam size, which would make running decoding on a single GPU impractical.
We reduce this amount by only training LoRA parameters on the last three of LLaMA's 32 layers.
This way, the LoRA finetuned classifiers can still take advantage of LLaMA's pretrained knowledge to some extent, but they only need distinct key-value caches for the final few layers.

The final outcome of this is that we only need to store a cache that is about $(29 + 3 \cdot 3)/32 \approx 1.19$ times the size that would be needed for just the LM alone.
We chose to do this instead of the transformer method used for length prediction in order to significantly increase the capacity of the classifiers, as well as letting them make better use of LLaMA's pretraining in order to learn from just 500 examples.

\paragraph{Lookahead batching.} Recall that Algorithm~\ref{alg:beam_cond} requires applying the classifier to $k$ candidate continuations for each of the $B$ beam hypotheses.
The trick to doing this efficiently is to batch over the candidates, but have all candidates share a single key-value cache.\footnote{This works since all candidates for a given hypothesis on the beam share a history. If we're looking at candidate outputs for position $t+1$, all hidden states up until $t$ will be identical.}

For this type of batching, we do need to modify the model code as well.
The reason is that naive batching would also lead to a batched output of the updated key-value cache, i.e., $B \times k$ copies of it.
To avoid that, we modify the LLaMA code to instead only output the key and value vectors for the $t$-th timestep.\footnote{This also requires modification of the attention mechanism. All reductions that go from time $1$ to $t$ are modified to handle timesteps $1$ to $t-1$ in a group, then the result for the present timestep is handled separately.}
After we select which tokens will be used for the next timestep, we make a single updated cache for each beam element, so that we never need to instantiate more $B$ key-value caches.

\paragraph{Summary of modifications to LLaMA.} The original LLaMA model is a function that has the following inputs:
\begin{enumerate}
    \item Model parameters
    \item Key-value cache
    \item Input token (At inference time just a single token would be input)
\end{enumerate}
and outputs:
\begin{enumerate}
    \item Classification or Next-token prediction logits
    \item New key-value pairs
\end{enumerate}
We modify it into a function that has the following inputs:
\begin{enumerate}
    \item Model parameters
    \item LoRA parameters (Batch axis: \texttt{lora})
    \item Key-value caches for each beam element and group of LoRA parameters (Batch axes: \texttt{lora} and \texttt{beam})
    \item One token for each of $k$ continuations of each beam element (Batch axes: \texttt{beam} and \texttt{continuation})
\end{enumerate}
and outputs:
\begin{enumerate}
    \item Classification \emph{and} next-token logits (Batch axes: \texttt{lora}, \texttt{beam}, \texttt{continuation})
    \item New key and value \emph{vectors} for the most recent timestep, which can be used to update the caches (Batch axes: \texttt{lora}, \texttt{beam}, \texttt{continuation})
\end{enumerate}
Note that for the key and value vectors, the \texttt{lora} batch axis is only present for the last three layers, since those are the layers we finetune.

\subsection{Beam search details}
We use a beam size of 5 for all experiments in this section.
To reduce the cost of training and inference, we limit the max sequence length to 196 tokens (including the prompt).
We evaluate $k=100$ candidates for conditional beam search, and use a weight of $\alpha = 2$.

\subsection{Weaknesses in classifier training}\label{sec:modesbeam_concessions}
Unlike in our length control experiments, we have to make various compromises in classifier training just due to resource constraints.
All the changes here should make the classifier \emph{worse} so we expect an implementation of conditional beam search without these limitations would have \emph{better} final performance.

\paragraph{Labeling budget.} The main constraint is that we have an extremely limited data labeling budget (both in terms of time and money), so we only use 500 training examples for each classifier.
Contrast this with the hundreds of thousands of training examples we were able to use for the length control classifier.

\paragraph{Unrepresentative training data.} Related to the above point, we had to generate the training data using beam search on LLaMA, rather than sampling.
The issue is that there are two things we need for training data to be effective for training an attribute classifier.
The first is that it should be representative of the model's output distribution, $\dmodel$.
The second is that the data should be representative of the types of inputs that the classifier will be used on at inference time.

These are conflicting goals, which we can explain with an example.
Consider the prompt ``Name three scifi/fantasy book series.''
Imagine that the beam search completion is ``1. Harry Potter, 2. Lord of the Rings, 3. Foundation.'', and a sample output is ``Some popular series are: Mistborn, The Expanse, and The Three Body Problem series.'' We want the classifier to be able to handle prefixes of the beam search output as input, but its predictions are \emph{about} the sampling output, which has different properties than the beam search output.

Due to our limited budget, we optimize for input validity over output validity, and use beam search completions from LLaMA-7B as our training data.
The ideal solution would be to use a wide enough range of samples that they achieve good coverage of model behavior, or to use outputs that start by using beam search but sample completions.\footnote{This latter strategy is the opposite of what VGBS does: \citet{vgbs} sample the start of the model decoding and complete it using beam search.}

For context on the difference in scales, the recently released LLAMA-2 Chat model was trained using over 1.4 \emph{million} human preference labels\citep{llama2}.
Using that scale of human labeling, we wouldn't need to make this compromise in terms of our data distribution.

\subsection{Results}
We find that, despite the weaknesses in classifier training discussed in the previous section, we're able to generate much more acceptable results from LLaMA-7B using conditional beam search than ordinary beam search.

\subsubsection{Classifier satisfaction}
\begin{table}[h]
\centering
\caption{Effect of conditioning on LLaMA-7B beam search outputs. ``\% of outputs'' is the percent of outputs for which the classifier predicted the attribute was the majority class. $\prob_\text{clf}(a)$ is the average of the classifiers predicted probability over the outputs.
The outputs in the conditional row are the same for both classifiers (i.e., both classifiers were used simultaneously to guide generation).}\label{tab:classifier_satisfaction}
\begin{tabular}{llllll}
  & Method & \multicolumn{2}{c}{Completeness $(\uparrow)$}  & \multicolumn{2}{c}{Irrelevance $(\downarrow)$} \\ \cmidrule(lr){3-4} \cmidrule(lr){5-6}
  &                    & \% of outputs & $\prob_\text{clf}(a)$ & \% of outputs & $\prob_\text{clf}(a)$ \\ \midrule
 & Beam search             & 92.34             &  0.5999 & 45.56     &  0.4996\\
  & Conditional beam search & \textbf{98.39}    & \textbf{0.6289} & \textbf{27.02}    & \textbf{0.3583}\\ \bottomrule 
\end{tabular}
\end{table}
The easiest measurement we can do is to look at how often the attribute classifiers label complete outputs as having satisfied the target.\footnote{We didn't do this in the previous section since we could directly constrain the outputs to be the right length. The classifier satisfaction would have been 100\% in all cases (or whatever the highest value it could output is).}
Table~\ref{tab:classifier_satisfaction} shows the classifier prediction statistics for the unconditional and conditional LLaMA-7B outputs.
For both completeness and irrelevance, using conditional beam search satisfies the classifier more often.

One thing to point out is that the classifier seems to predict that outputs are complete far more often than they actually are.\footnote{This can be verified by looking back at the statistics about degenerate outputs from the start of the chapter.}
Even though it isn't good in absolute terms, the classifier is still effective in terms of guiding beam search, as we'll see in the next few sections.

\subsubsection{Improvement in identified degeneracies}
\begin{table}[h]
    \centering
    \caption{Improvement in occurrence of degenerate outputs from using conditional beam search with LLaMA-7B. These were measured using 248 prompts from the \texttt{databricks-dolly-15k} dataset.}\label{tab:beam_heuristics}
    \begin{tabular}{lll}
        Behavior & Beam search & Conditional beam search\\
        \midrule
        Empty output & 18.6\% & 0.0\%\\
        Spurious TeX & 6.5\% & 0.0\%\\
        Repeating prompt & 29.0\% & 4.8\%\\
        \bottomrule
    \end{tabular}
\end{table}
 At the start of this chapter, we mentioned several problems we observed, namely: empty outputs, repetition of the prompt, and emission of unecessary TeX code.
These problems were dramatically improved by the addition of conditioning.

Table~\ref{tab:beam_heuristics} shows the frequency of each problem under each type of beam search.\footnote{Even though LLaMA's exact mode is empty about 70\% of the time, beam search doesn't usually find that out.} The rate of these three undesired behaviors drops significantly when attempting to condition on Completeness and Relevance.
The classifier seems to be learning non-trivial behavior, not just lexical rules.
For example, given the input ``Output a LaTeX table'', it still TODO.

\subsection{Qualitative analysis}
We'll now look at some examples of interesting behavior that the conditional beam search outputs show.
Appendix~\ref{chapter:beam_appendix} shows the outputs with and without conditioning for a non-cherrypicked subset of the Dolly-15k prompts.

\subsubsection{Responses to factual but high-entropy prompts}
As we saw in Chapter~\ref{chapter:modes_exact}, inputs with a high-number of possible responses are more likely to have the empty sequence as their mode.
While we wouldn't recommend using MAP for something like poetry generation, it's easy to come up with prompts that are requesting concrete factual information, but also have a large number of possible responses.
Here's an example:
\begin{mdframed}[backgroundcolor=gray!10, linewidth=0pt, innertopmargin=10pt, innerbottommargin=10pt]
    \textbf{Prompt:} ``Suggest three restaurants in Chicago.''\\
    \textbf{Beam search:} \texttt{<empty>}\\
    \textbf{Cond. beam search:} ``1. Lou Malnati's\\
    2. Giordano's\\
    3. Gino's East''
\end{mdframed}
Here, any three restaurants will do, but there are many possible responses, so it's unsurprising that unconditional beam search gives an empty response.

Here's another example where there are many ways to choose to answer, but once a choice is made there's no ``creativity'' required:
Here's an example:
\begin{mdframed}[backgroundcolor=gray!10, linewidth=0pt, innertopmargin=10pt, innerbottommargin=10pt]
    \textbf{Prompt:} ``Translate the phrase "PhD thesis" into 5 languages''\\
    \textbf{Beam search:} \texttt{<empty>}\\
    \textbf{Cond. beam search:} ``* French: thèse de doctorat\\
* German: Dissertation\\
* Italian: tesi di dottorato\\
* Portuguese: tese de doutorado\\
* Spanish: tesis de doctorado''
\end{mdframed}
The German translation is incorrect, but the conditional beam search output is high-quality, while beam search only finds the empty sequence.

\subsubsection{Factual uncertainty}
There may also be cases where the model is uncertain about the answer, but conditional on the answer the output is highly constrained. This example shows how even factual questions can cause degenerate outputs from the base model:
\begin{mdframed}[backgroundcolor=gray!10, linewidth=0pt, innertopmargin=10pt, innerbottommargin=10pt]
    \begin{CJK*}{UTF8}{gbsn}
    \textbf{Prompt:} ``What is the literal and figurative meaning of "雪中送炭"?''\\
    \textbf{Beam search:} ``雪中送炭\\\\雪中送炭\\\texttt{<continues repeating>}''\\
    \textbf{Cond. beam search:} ``The literal meaning of "雪中送炭" is "to send coal in the snow".''
    \end{CJK*}
\end{mdframed}
Conditional beam search only half answered the question, but unconditional beam search ends up in meaningless repetition.\footnote{This particular case wouldn't have been caught by our ``repeating the prompt'' heuristic since the output isn't an exact suffix of the prompt.}

\subsubsection{TeX outputs}
Here are two examples of outputs where the ``TeX problem'' is fixed:
\begin{mdframed}[backgroundcolor=gray!10, linewidth=0pt, innertopmargin=10pt, innerbottommargin=10pt]
    \textbf{Prompt:} ``Name three famous AI researchers''\\
    \textbf{Beam search:} ``\textbackslash{}begin\{itemize\}\\
\textbackslash{}item Alan Turing\\
\textbackslash{}item John McCarthy\\
\textbackslash{}item Marvin Minsky\\
\textbackslash{}end{itemize}''\\
    \textbf{Cond. beam search:} ``Alan Turing, John McCarthy, and Marvin Minsky''
\end{mdframed}
In this one, the same response is output, just without the unnecessary markup.

\begin{mdframed}[backgroundcolor=gray!10, linewidth=0pt, innertopmargin=10pt, innerbottommargin=10pt]
    \textbf{Prompt:} ``Can you share a response of politely informing the customer that we dont support a feature''\\
    \textbf{Beam search:} ``\textbackslash{}begin\{blockquote\}\texttt{\textbackslash{}n\textbackslash{}n}\textbackslash{}begin\{blockquote\}\texttt{\textbackslash{}n\textbackslash{}n}\texttt{...<repeated>}''\\
    \textbf{Cond. beam search:} Thank you for your inquiry. I am sorry to inform you that we do not support your request at this time.
\end{mdframed}
In this one, beam search collapses to a degenerate output, while conditional beam search finds a bland but correct response.\footnote{ ``Bland but correct'' is basically what we want out of a good MAP-style algorithm. See Section~\ref{sec:modestheory_templates} for discussion of why this is expected.}

\subsection{Discussion: Conditional beam search with LLaMA}
Despite only using a tiny amount of labeled data, we were able to guide LLaMA-7B to produce better outptuts using conditional beam search.
In particular, the rates of each of the three degeneracies we targeted decreased significantly.
Qualitatively, we see that we are able to find outputs of reasonable quality using conditional beam search.
This is true even for inputs with high output entropy, which ordinary beam search will tend to produce trivial responses to.

However, there are still problems present in the outputs of conditional beam search.
At least some of these are likely due to the classifier being too weak.
As shown in Table~\ref{tab:classifier_satisfaction}, the Completeness classifier predicted that 92\% of beam search outputs are complete, which is very much at odds with our measured rates of output degeneracy.
To address this, we will need to train better classifiers, both by using more labeled data, and by improving the training procedure.
A more accurate classifier should lead to even higher quality outputs, allowing us to investigate how far we can go without any tuning of the underlying model.

\subsection{Future work: Performance improvements}\label{sec:modesbeam_perf}
Our main goal in these experiments was to find out what was possible by purely guiding beam search, instead of modifying the model weights themselves.
This led to a method that worked, but was very slow.
Generation from LLaMA-7B with two attribute classifiers, $k=100$ and a beam size of 5 runs at about 1 token/second on an NVIDIA Quadro RTX 6000 GPU, which is far too slow to be practically useful.
In this section we discuss how conditional beam search can be made more efficient.

\paragraph{Custom GPU kernels.} Our experiments use pure JAX implementations of 4-bit dequantization and attention.
Using optimized GPU kernels would significantly improve performance.
See \citet{qlora} for information on 4-bit matmul kernels, and \citet{flashattention2} for an optimized attention implementation.

\paragraph{Varying $k$.} We fixed the value of $k$ (the number of candidate next tokens to consider) to 100 in all our experiments. Because we only need to look at high likelihood tokens, it's possible that a much lower value of $k$ would not change generation outputs. To find the minimal value of $k$ that would leave the outputs unchanged, we could just record the rank of each token selected during decoding (in terms of the original model's likelihood of each token). If the top 5 tokens in conditional likelihood only come from the top 20 tokens in terms of unconditional likelihood, we could get a large speedup.

\paragraph{Classifier architectures.} The reason we used an expensive classifier architecture was so that we could take advantage of LLaMA's pretraining to manipulate its own hidden states.
Using more training data should allow us to train classifiers from scratch with a similar architecture as used in Section~\ref{sec:modesbeam_length}.
Alternatively we could still train the costly LoRA-based classifiers, then use model distillation to create a cheaper classifier.

\subsection{Further Future work}
As we said in Chapter~\ref{chapter:modes_theory}, we see both conditional sampling and conditional search as promising ways of improving output quality.
In this chapter we focused on conditional search, but hope to look at conditional sampling in the future.
Instead of conditioning away the problems of search outputs: emptiness and reptition, we would instead try to condition on things such as fluency and factual correctness, which suffer when sampling.

The main direction for future work on conditional search is to scale up the experiments, both by training on more data, and by testing on larger models.
We discussed implementation efficiency as a particular focus in Section~\ref{sec:modesbeam_perf}.

\section{Conclusion}
In Chapter~\ref{chapter:modes_exact} we found that length-conditional modes for NMT and story completion were often high quality, as opposed to continuing to suffer from the ``bad mode'' problem even after conditioning.

This inspired us to see if we could find a more practical approach to search for conditional modes, which we did in this section.
Our length-constrained generation experiments that used beam search with the conditional distribution, $\pmodel(x | |x| = L)$, successfully allowed us to find higher scoring sequences satisfying the length constraint than those found by beam search.
These sequences tended to be grammatical far more often than those found by ordinary beam search, so in this case the difference in likelihood was reflected by a difference in quality.

We then moved on to the more ambitious goal of improving the instruction-following behavior of LLaMA-7B, a model that was not trained for instruction following.
Using only 500 labeled training examples for ``completeness'' and ``irrelevance'' we trained classifiers with a novel architecture that allowed us to implement conditional beam search.
The outputs were higher quality than ordinary beam search, both in terms of our automated measures of degenerate outputs, and qualitatively.

We aren't claiming that this method is currently ready to replace sampling based methods for NLG, but we see it as a promising step towards more powerful MAP based methods.
Using attribute classifiers to fix output degeneracy lets us get high quality outputs \emph{without} needing to finetune the underlying language model at all, which is the goal we laid out in Chapter~\ref{chapter:modes_theory}.

\chapter{Conclusion}
This thesis has offered a number of novel methods and insights for the application of pretrained language models.
These methods are generically applicable, and can especially provide value for models which have been trained on specific domains, which are not directly targeted by the state of the art LLMs.
Chapters~\ref{chapter:recurrence} and \ref{chapter:mt} offered new methods for finetuning pretrained models, opening up new possibilities for pretrained LMs and MLMs.
Chapters~\ref{chapter:hso} and \ref{chapter:modes_beam} suggested new inference techniques for producing higher quality outputs from a pretrained model without any further finetuning
In Chapters~\ref{chapter:modes_theory} and \ref{chapter:modes_exact}, we argued that the NLP community should consider the bad mode problem in a more nuanced way, rather than chalking it up to model error and trying to finetune it away.
The work in this thesis represents important progress on making more efficient use of pretrained models, which should have a multiplicative effect with the ongoing work of training ever better NLP models.

\section{Future work}
The main question we hope to see answered in future work is: ``What is the best quality we can reach using a fixed language model.''
The fact that we were able to extract much better behavior from LLaMA-7B using conditional beam search as opposed to standard decoding methods implies that the information necessary to behave that way is contained in the model.
Our current decoding/inference techniques seem to not be up to the task of extracting those high-quality outputs though, even if the model \emph{knows} they are good.

The next step is to significantly up the scale and quality of classifier training for conditional beam search, in order to find out at what point the quality improvements will saturate.
We will also attempt to use more advanced search techniques such as MCTS, which has been used to try to maximize reward, but has not been considered as a conditional search technique.

\bibliography{bibliography,anthology}
\bibliographystyle{acl_natbib}

\appendix

\chapter{RUM-SUNDAE Hyperparameters}
We used the same training hyperparmeters as \citet{sundae} where possible.
On WMT'14 De-En we trained the randomly initialized baseline with a batch size of 4096 for 570K steps, which corresponds to about 518 epochs.
We trained the pretrained models for only 115K steps, with a batch size of 512, which is approximately 13 epochs.
On WMT'16 RO-EN the steps used were baseline: 136K (910 epochs), pretrained: 37K (31 epochs).
On CodexGLUE Java-C\#, the steps used were baseline: 10K (4096 epochs), pretrained: 10K (512 epochs).

For finetuning the pretrained models, we disabled weight decay and label smoothing, with the expectation that they would damage the information in the pretrained models.\footnote{Label smoothing in particular is likely detrimental, as XLM-R has a large multilingual vocabulary, most of which is unused for any given language pair. As a result, label smoothing would encourage it to allocate weight to tokens which will never be used.} For test time, we used 15 unroll steps for all models, took the best of 16 samples, and used sampling temperatures in the range $[0.1, 1.1]$. All hyperparameter decisions were made based on results on validation splits.

\chapter{Global and Conditional Modes}\label{chapter:modes_appendix}
This appendix contains the modal outputs for various NLG models which are discussed in the main text.

\begin{CJK*}{UTF8}{gbsn}
\begin{longtable}{lcp{0.6\linewidth}}
    \caption{Unconditional modes and length-conditional modes of the MarianMT Zh-En model}\label{tab:mt_modes}\\
    Type & $\log P(y | x)$ & Text\\
    \toprule
    Source A & - & 需要带上大量的水。\\
Reference (11 tokens) & - & Take lots of water.\\
Mode & -6.17 & Needs a lot of water.\\
Mode (length 8) & -6.36 & A lot of water is needed.\\
Mode (length 10) & -6.26 & A lot of water needs to be brought.\\
Mode (length 12) & -8.55 & There is a need to bring a lot of water.\\
\midrule
Source B & - & 幸运的是，他们安全通过了。\\
Reference (13 tokens) & - & Fortunately they worked.\\
Mode & -2.65 & Fortunately, they passed safely.\\
Mode (length 8) & -4.49 & Fortunately, they have passed safely.\\
Mode (length 10) & -7.83 & Fortunately, they're safe to pass.\\
Mode (length 12) & -9.92 & Fortunately, it's safe for them to pass.\\
\midrule
Source C & - & 实际应该是2014年而非2013年。\\
Reference (8 tokens) & - & It was 2014, not 2013.\\
Mode & -5.04 & Rather than 2013.\\
Mode (length 8) & -5.70 & It should be 2014 instead of 2013.\\
Mode (length 10) & -7.60 & It was supposed to be 2014 instead of 2013.\\
Mode (length 12) & -9.83 & In real terms, it should be 2014 instead of 2013.\\
\midrule
Source D & - & 把父母放心交给我。\\
Reference (11 tokens) & - & Leave your parents to me.\\
Mode & -5.35 & Give me your parents.\\
Mode (length 8) & -7.46 & Put your parents in my hands.\\
Mode (length 10) & -8.44 & Leave it to me to trust my parents.\\
Mode (length 12) & -10.59 & Leave it to me to be assured of my parents.\\
\midrule
Source E & - & 有8名遇难者的遗体一直没有找到。\\
Reference (14 tokens) & - & Eight bodies have never been found.\\
Mode & -4.97 & The remains of eight victims were never found.\\
Mode (length 8) & -8.33 & The remains of eight victims remained.\\
Mode (length 10) & -4.97 & The remains of eight victims were never found.\\
Mode (length 12) & -5.13 & The remains of eight of the victims were never found.\\
\midrule
Source F & - & 不过，有些人则没那么乐观。\\
Reference (13 tokens) & - & But some are not that optimistic.\\
Mode & -3.52 & Some, however, are less optimistic.\\
Mode (length 8) & -4.01 & However, some are less optimistic.\\
Mode (length 10) & -4.65 & Some people, however, are less optimistic.\\
Mode (length 12) & -7.86 & There are, however, some who are less optimistic.\\
\midrule
Source G & - & 现在我就是你们的家人。\\
Reference (14 tokens) & - & I am a member of your family now.\\
Mode & -2.52 & Now I'm your family.\\
Mode (length 8) & -2.52 & Now I'm your family.\\
Mode (length 10) & -7.67 & Well, now I'm your family.\\
Mode (length 12) & -10.73 & \# Now I'm your family. \#\\
\midrule
Source H & - & 我们九点在码头集合。\\
Reference (13 tokens) & - & We meet on the quay at nine.\\
Mode & -5.31 & We meet at 9.\\
Mode (length 8) & -8.16 & We'll meet up at 9.\\
Mode (length 10) & -6.00 & We meet at the docks at 9:00.\\
Mode (length 12) & -6.83 & We'll meet at the docks at 9:00.\\
\midrule
Source I & - & 我忍不住会想到谁出现在了赛场上。\\
Reference (13 tokens) & - & I couldn't help who was here.\\
Mode & -7.62 & I can't help but wonder who showed up.\\
Mode (length 8) & -11.26 & I can't help it.\\
Mode (length 10) & -10.22 & I cannot help but wonder who showed up.\\
Mode (length 12) & -7.62 & I can't help but wonder who showed up.\\
\midrule
Source J & - & 我不想沉默。\\
Reference (10 tokens) & - & I don't want silence.\\
Mode & -2.69 & I don't want to be silent.\\
Mode (length 8) & -6.59 & I don't want silence.\\
Mode (length 10) & -2.69 & I don't want to be silent.\\
Mode (length 12) & -9.01 & No, I don't want to be silent.\\
\midrule
Source K & - & 企业集团就网络安全法向中国提诉求\\
Reference (26 tokens) & - & Business Groups Appeal to China Over Cybersecurity Law\\
Mode & -7.89 & \texttt{<empty>}\\
Mode (length 8) & -9.58 & Group claims to China on cyber security\\
Mode (length 10) & -9.91 & Corporate groups complain to China about cyber security laws\\
Mode (length 12) & -11.95 & Corporate groups complain to China about cyber-security laws.\\
\midrule
Source L & - & 当我们前往解决池水变绿的问题时，对最佳化学物质进行过讨论。\\
Reference (32 tokens) & - & When we went to fix the green, there was a discussion about the best chemicals.\\
Mode & -7.85 & \texttt{<empty>}\\
Mode (length 8) & -11.47 & The best chemical substances were discussed.\\
Mode (length 10) & -13.97 & The best chemical substances were discussed when we.\\
Mode (length 12) & -13.43 & Best chemicals were discussed when we turned the pool green.\\
\midrule
Source M & - & 爱尔兰超级足球联赛：费恩哈普0-5不敌德利城\\
Reference (27 tokens) & - & League of Ireland Premier Division: Finn Harps 0-5 Derry City\\
Mode & -7.60 & \texttt{<empty>}\\
Mode (length 8) & -12.10 & Irish Super Football League: Finn Harper\\
Mode (length 10) & -10.34 & Irish Super Football League: Finn Harper 0-5\\
Mode (length 12) & -12.27 & Irish SuperSoccer: Finn Harper 0-5.\\
\midrule
Source N & - & 不出所料的是，在他们总共34粒稀松的进球中，有十几粒进球出自定位球。\\
Reference (33 tokens) & - & Predictably, a dozen of their sparse total of 34 came from set pieces.\\
Mode & -8.85 & \texttt{<empty>}\\
Mode (length 8) & -13.04 & Not surprisingly, of their total 34\\
Mode (length 10) & -13.62 & Not surprisingly, out of a total of 34\\
Mode (length 12) & -14.83 & Unsurprisingly, a dozen of their 34\\
\midrule
Source O & - & 赢得联赛冠军的赔率（通过Oddschecker统计）为1,000-1\\
Reference (26 tokens) & - & Odds to win the league (via Oddschecker) 1,000-1\\
Mode & -6.45 & \texttt{<empty>}\\
Mode (length 8) & -13.34 & (ddschecker)\\
Mode (length 10) & -14.43 & (by Oddschecker)\\
Mode (length 12) & -13.87 & The odds of winning the League championship are 1,000-1.\\
\midrule
Source P & - & 基尔马诺克武士刀“血洗”案兄弟俩被判入狱\\
Reference (30 tokens) & - & Brothers jailed for samurai sword 'bloodbath' in Kilmarnock\\
Mode & -7.80 & \texttt{<empty>}\\
Mode (length 8) & -11.26 & Two brothers were sentenced to prison.\\
Mode (length 10) & -14.28 & The two brothers in the Kilmanok.\\
Mode (length 12) & -14.88 & In this case, two brothers were sentenced to prison.\\
\midrule
Source Q & - & 夺冠反应：西蒙·曼努埃尔的历史时刻看起来如何\\
Reference (28 tokens) & - & Golden Reaction: What Simone Manuel's Historic Moment Looked Like\\
Mode & -7.45 & \texttt{<empty>}\\
Mode (length 8) & -10.90 & What does Simon Manuel look like?\\
Mode (length 10) & -8.61 & How does Simon Manuel's history look?\\
Mode (length 12) & -9.72 & Champ: How does Simon Manuel's history look?\\
\midrule
Source R & - & 泰国领导人认为针对旅游景区的袭击与宪法更替有关\\
Reference (30 tokens) & - & Thai Leader Links Attacks on Tourist Sites to Constitution Change\\
Mode & -8.50 & \texttt{<empty>}\\
Mode (length 8) & -12.75 & Thai leaders believe that attacks on tourist\\
Mode (length 10) & -14.45 & Thai leaders believe that the attack on the tourist\\
Mode (length 12) & -12.33 & Thai leaders consider attacks on tourist sites related to constitutional change\\
\midrule
Source S & - & 塔塔钢铁的消息来源警告称，该公司仍可能卖掉塔尔伯特港工厂。\\
Reference (30 tokens) & - & Tata Steel sources have warned it could still sell Port Talbot.\\
Mode & -7.28 & \texttt{<empty>}\\
Mode (length 8) & -14.00 & plant in the port of Talbot.\\
Mode (length 10) & -15.72 & Chargé d'affaires a.i.\\
Mode (length 12) & -14.97 & Tata steel sources warned that it could still sell the\\
\midrule
Source T & - & 后场球员、中场球员和前场球员，我们都必须加强。\\
Reference (33 tokens) & - & The back players, midfield players and front players, we have to strengthen.\\
Mode & -8.78 & \texttt{<empty>}\\
Mode (length 8) & -12.67 & All of us must be strengthened.\\
Mode (length 10) & -12.83 & We must strengthen rear, middle and front.\\
Mode (length 12) & -11.13 & We must all strengthen rear, middle and front players.\\

\end{longtable}
\end{CJK*}

\newpage
\begin{longtable}{lcp{0.6\linewidth}}
    \caption{Unconditional and length-conditional modes of our ROC Stories finetuned GPT2-345M model}\label{tab:roc_modes}\\
    Type & $\log P(x_{\ge t} | x_{<t})$ & Text\\
    \toprule
    Story A & - & Janice usually wears jeans to work every day. However, now she has been promoted to manager. She decides she needs to dress a little more formally. Janice buys a few pairs of khakis for work.\\
Reference (9 tokens) & - & She also bought plenty of blouses.\\
Mode & -6.44 & \texttt{<empty>}\\
Mode (length 8) & -8.31 &  She feels more confident at work.\\
Mode (length 10) & -7.45 &  Janice is happy with her new look.\\
Mode (length 12) & -8.89 &  She is glad she no longer has to wear jeans.\\
\midrule
Story B & - & Francine noticed that all of her friends wore high heeled shoes. Although she loved how heels looked, she hated how they felt. One day she decided to wear a pair of flats to meet her friends. All of her friends complimented how great they looked.\\
Reference (12 tokens) & - & Francine was glad that she wore comfortable  shoes!\\
Mode & -5.60 &  Francine never wore heels again.\\
Mode (length 8) & -5.60 &  Francine never wore heels again.\\
Mode (length 10) & -7.54 &  Francine decided to wear heels more often.\\
Mode (length 12) & -7.46 &  Francine decided to wear heels again in the future.\\
\midrule
Story C & - & Tuesdays are laundry days at my apartment. We have been too busy the last couple of Tuesdays. Now we have almost no clean clothes left. I'm dressed foolishly and still smell bad.\\
Reference (8 tokens) & - & I will do laundry right now.\\
Mode & -7.71 &  I don't know what to do.\\
Mode (length 8) & -9.35 &  I need to buy new clothes.\\
Mode (length 10) & -10.38 &  I don't know what to do now.\\
Mode (length 12) & -9.62 &  I don't know what I'm going to do.\\
\midrule
Story D & - & Jill convinced her boyfriend Joe to go look for Geocache with her. He didn't think it sounded like fun but decided to humor her. They searched the location where the Geocache was supposed to be. After over an hour of searching they were unable to find it.\\
Reference (10 tokens) & - & Jill hoped they would find it soon.\\
Mode & -7.47 & \texttt{<empty>}\\
Mode (length 8) & -8.89 &  Jill was very disappointed in Joe.\\
Mode (length 10) & -10.50 &  Jill was disappointed but Joe didn't care.\\
Mode (length 12) & -10.73 &  Jill and Joe never went to Geocache again.\\
\midrule
Story E & - & My wife had MLK day off. She slept in, and did not get up until 10 AM. We had a leisurely breakfast. She watched Little House on the Prairie while I surfed the net.\\
Reference (12 tokens) & - & Then we had a relaxing evening covering on the couch.\\
Mode & -5.63 & \texttt{<empty>}\\
Mode (length 8) & -8.35 &  It was the best day ever.\\
Mode (length 10) & -9.13 &  It was the best day of her life.\\
Mode (length 12) & -10.41 &  It was one of the best days of my life.\\
\midrule
Story F & - & Abby and Tammy were the best of friends. They both loved to do things together that were fun and creative. They decided to make friendship bracelets together and give to others. Abby made five and Tammy made seven more.\\
Reference (11 tokens) & - & They decided to keep the bracelets for themselves.\\
Mode & -7.61 & \texttt{<empty>}\\
Mode (length 8) & -9.11 &  They were the best of friends.\\
Mode (length 10) & -8.59 &  Abby and Tammy were the best of friends.\\
Mode (length 12) & -10.34 &  They are now the best bracelets in the world!\\
\midrule
Story G & - & Last Friday was Tad Dunkin's first race in nascar. He had been waiting for this his whole life. He was doing surprisingly well for a first timer. Then he lost control and hit a wall.\\
Reference (7 tokens) & - & Tad was seriously injured.\\
Mode & -7.12 &  He was disqualified.\\
Mode (length 8) & -7.96 &  Tad never wanted to race again.\\
Mode (length 10) & -7.88 &  Tad had to be rushed to the hospital.\\
Mode (length 12) & -9.33 &  He had to be airlifted to the hospital.\\
\midrule
Story H & - & Hannah was an amazing artist. She always had a natural gift. She decided to enter in an art competition. Thankfully she was able to win the top prize.\\
Reference (7 tokens) & - & Her parents were very proud.\\
Mode & -4.69 &  She was so happy.\\
Mode (length 8) & -5.62 &  She was very proud of herself.\\
Mode (length 10) & -8.34 &  She went on to become a famous artist.\\
Mode (length 12) & -10.53 &  She couldn't wait to share it with her friends.\\
\midrule
Story I & - & Joe was pals with Tim. They always played together at recess. One day Joe said he was going to move away. Tim was sad.\\
Reference (8 tokens) & - & Joe and Tim stayed friends online.\\
Mode & -3.73 &  They never spoke again.\\
Mode (length 8) & -5.55 &  They never saw each other again.\\
Mode (length 10) & -8.65 &  He never talked to Joe again after that.\\
Mode (length 12) & -10.08 &  They didn't see each other again for a while.\\
\midrule
Story J & - & Bay was nervous. Her boyfriend had been acting weird all through dinner. Bay thought he was going to dump her. But then he got on one knee.\\
Reference (8 tokens) & - & And asked her to marry him.\\
Mode & -4.45 &  He asked her to marry him!\\
Mode (length 8) & -4.45 &  He asked her to marry him!\\
Mode (length 10) & -6.91 &  He proposed to her and she said yes!\\
Mode (length 12) & -6.88 &  He asked her to marry him and she said yes!\\
\bottomrule

\end{longtable}

\newpage
\begin{table}[h]
\centering
\caption{Examples of prompts with empty and non-empty modes for Alpaca-7B.}\label{tab:alpaca_empty_prompts}
\begin{tabular}{p{0.45\textwidth}p{0.45\textwidth}}
Empty & Non-empty\\\toprule
``how would you start explaining mathematics to kids?'' & ``How can listening to music attentively influence you?''\\
\addlinespace[0.5cm]
``I want to get in better shape.  I work at a desk all day, and I've never really been in good shape.  Growing up, I didn't play sports or spend a lot of time outdoors.  I know I need to improve my physical health, but I really don't know how to get started.  Can you recommend a workout routine for me?'' & ``What are some of the most accessible jazz albums for someone new to jazz?''\\
\addlinespace[0.5cm]
``What is it like to own a dog that sheds everywhere?'' & ``What was the first British instrumental to top the USA charts''\\
\addlinespace[0.5cm]
``Give me a list of date night ideas that I've never done.'' & ``What is the difference between a goose and a geese?''\\
\addlinespace[0.5cm]
``How can you take good star photos?'' & ``What to do when you are bored?''\\
\addlinespace[0.5cm]
``What are some disadvantages of the way the tax code treats incentive stock options?'' & ``List 7 exotic fruits that I should try.''\\
\addlinespace[0.5cm]
``What activities an admin or an administrator of any data tools \& platform or data tools can do?'' & ``What are the names of popular Alternative music bands from the 1980s and 1990s.''\\
\addlinespace[0.5cm]
``What is the future trend of job industry'' & ``What's the best BBQ place in Austin''\\
\addlinespace[0.5cm]
``I need to improve my sleep.  Give me a list of ideas for doing so.'' & ``What is C++?''\\
\addlinespace[0.5cm]
``Write a brief paragraph of the benefits of attending Arizona State University'' & ``Should investors time the market?''\\
\addlinespace[0.5cm]
\bottomrule
\end{tabular}
\end{table}

\newpage
\begin{table}[h]
\centering
\caption{Examples of prompts with empty and non-empty modes for Guanaco-7B.}\label{tab:guanaco_empty_prompts}
\begin{tabular}{p{0.45\textwidth}p{0.45\textwidth}}
Empty & Non-empty\\\toprule
``What are some disadvantages of the way the tax code treats incentive stock options?'' & ``What are the words of House Lannister?''\\
\addlinespace[0.5cm]
``how would you start explaining mathematics to kids?'' & ``What kind of method is Transfer printing''\\
\addlinespace[0.5cm]
``What are some best practices to prepare biryani'' & ``What is the best book to read about the Battle of Stalingrad?''\\
\addlinespace[0.5cm]
``What are some tools to help combat ADD and ADHD?'' & ``Is Daft Punk still together?''\\
\addlinespace[0.5cm]
``Imagine you have won the lottery, and have 5 million dollars after tax to spend in San Francisco, where you currently rent a 2 bedroom apartment with three roommates who are your best friends but who you hate living. Describe how you would use the money, keeping in mind you don't have a high paying job so you want to do fun things and also set yourself up for the future.'' & ``Why do cats make purring sounds?''\\
\addlinespace[0.5cm]
``When to use mulch for your landscape?'' & ``What is the oldest country in the world?''\\
\addlinespace[0.5cm]
``Give me a bulleted list of ingredients that I need to bake chewy chocolate chip cookies, and include volume measurements for any ingredient I have to measure out.'' & ``How should I learn guitar?''\\
\addlinespace[0.5cm]
``Give step by step instructions on how to make a Long Island Ice Tea.'' & ``What was the first British instrumental to top the USA charts''\\
\addlinespace[0.5cm]
``What should I think about when buying a car (summarization)'' & ``What is Pascal?''\\
\addlinespace[0.5cm]
``Describe a plan for a road trip across Northern Italy'' & ``What are some of the most common vegetables in the broccoli family?''\\
\addlinespace[0.5cm]
\bottomrule
\end{tabular}
\end{table}

\newpage
\begin{table}[h]
\centering
\caption{Examples of prompts with empty and non-empty modes for LLaMA-7B. The modal output for the ``We are getting a new puppy'' prompt is an exact repetition of the prompt.}\label{tab:llama_empty_prompts}
\begin{tabular}{p{0.45\textwidth}p{0.45\textwidth}}
Empty & Non-empty\\\toprule
``What are the top 5 soccer(football) leagues in the world?'' & ``Can cars have odd number of wheels?''\\
\addlinespace[0.5cm]
``Compared to a human, categorize the following as fast or slow animals: sloth, cheetah, eagle, tortoise, hippo, slug, horse.'' & ``What black sweet is particularly popular in the Netherlands''\\
\addlinespace[0.5cm]
``When is an object considered a planet?'' & ``Is 95 degrees Fahrenheit considered to be hot weather?''\\
\addlinespace[0.5cm]
``How can I contribute to Spark?'' & ``What is the name of Google's email service?''\\
\addlinespace[0.5cm]
``What are some tools to help combat ADD and ADHD?'' & ``Identify which instrument is string or percussion: Clapper, Chagane''\\
\addlinespace[0.5cm]
``Why do cats make purring sounds?'' & ``We are getting a new puppy today, I am writing this messages to you our beloved children that we need to keep the Cats away from the new puppy for at least the first 2 weeks.''\\
\addlinespace[0.5cm]
``What is the purpose of a shot clock in basketball?'' & ``What is your favorite ice cream flavor?''\\
\addlinespace[0.5cm]
``What is the best way to drink coffee?'' & ``Who is the best CEO in the world''\\
\addlinespace[0.5cm]
``Who are some of the most influential innovators of all time?'' & ``Which is a species of fish? Goat or Goatfish''\\
\addlinespace[0.5cm]
``Describe how social media can help companies make better business decisions. What social media data can be collected to aid in business decisions? List four pieces of social media data (e.g., Twitter mentions and the number of retweets would be counted as two pieces).'' & ``Which characters belong to DC or Marvel Universe? Atom, Hank Pym''\\
\addlinespace[0.5cm]
\bottomrule
\end{tabular}
\end{table}

\newpage
\begin{longtable}{lcp{0.6\linewidth}}
    \caption{Exact Alpaca-7B modes for 10 randomly selected prompts (from the set of inputs which have non-empty modes)}\label{tab:alpaca_modes}\\
    Type & $\log P(x_{\ge t} | x_{<t})$ & Text\\
    \toprule
    Alpaca Prompt A & - & How can listening to music attentively influence you?\\
Reference (62 tokens) & - & You'll likely become a better listener in general - not only in hearing new details in songs but being a better listener during conversations. Listening to music attentively will also make you more present in the moment and enjoy the flow instead of focusing on the future or the past too [...]\\
Mode & -9.37 & Listening to music attentively can help to improve focus and concentration, reduce stress and anxiety, and improve overall wellbeing.\\
\midrule
Alpaca Prompt B & - & What are some of the most accessible jazz albums for someone new to jazz?\\
Reference (844 tokens) & - & Jazz music encompasses a wide range of styles, genres and periods, so it can be difficult and confusing for the non-jazz listener to find a satisfying entry point. There have been many “greatest of all time” jazz album lists developed by jazz publications and critics over the years. The following is a [...]\\
Mode & -9.45 & For someone new to jazz, some of the most accessible albums include In a Sentimental Mood by Duke Ellington, St. Louis Blues by Count Basie, Take the 'A' Train by Duke Ellington, and Mood Indigo by Duke Ellington.\\
\midrule
Alpaca Prompt C & - & What was the first British instrumental to top the USA charts\\
Reference (8 tokens) & - & Telstar by The Tornados\\
Mode & -1.88 & The first British instrumental to top the USA charts was "Apache" by The Shadows in 1960.\\
\midrule
Alpaca Prompt D & - & What is the difference between a goose and a geese?\\
Reference (75 tokens) & - & A goose and a geese refer to the same type of bird. These bird(s) are of the Anatidae biological family of water birds. Anatidate includes ducks, geese, seabirds, storks, and waterfowls. Goose refers to a single bird where geese refers to multiple birds.\\
Mode & -2.92 & The difference between a goose and a geese is that a goose is the singular form of the word, while geese is the plural form.\\
\midrule
Alpaca Prompt E & - & What to do when you are bored?\\
Reference (107 tokens) & - & If you are bored, you can use this as an opportunity to read a book, go outside for a walk, learn something new, paint, color, draw, call a friend that you haven't spoken to in a while, make a to-do list of items that you've been putting off and get to work on those, yard work, spend time volunteering to help [...]\\
Mode & -9.87 & When you are bored, there are many things you can do to pass the time, such as reading a book, playing a game, watching a movie, going for a walk, calling a friend, or learning a new skill.\\
\midrule
Alpaca Prompt F & - & List 7 exotic fruits that I should try.\\
Reference (41 tokens) & - & 1. Dragon fruit
2. Passion fruit 
3. Cherimoya
4. Soursop
5. Star fruit
6. Papaya
7. Mangosteen\\
Mode & -5.87 & 1. Durian
2. mangosteen
3. rambutan
4. dragon fruit
5. jackfruit
6. mangosteen
7. durian\\
\midrule
Alpaca Prompt G & - & What are the names of popular Alternative music bands from the 1980s and 1990s.\\
Reference (301 tokens) & - & Bands that were considered Alternative rock music in the 1980s were R.E.M., Depeche Mode, Nirvana, O.M.D., the Smiths, the Psychadelic Furs, the Housemartins, Green Day, the Violent Femmes, the B-52s, the Cure, UB40, and U2, just to name a few. Keep in mind that some of the bands that had been [...]\\
Mode & -9.81 & 1. Nirvana\\
\midrule
Alpaca Prompt H & - & What's the best BBQ place in Austin\\
Reference (86 tokens) & - & Texas BBQ is one of the most famous and popular food choices for both visitors and locals. The Franklin BBQ in Austin gets the highest hypes all these years after former President Obama's visit. Its long waiting line has not pushed foodies and pilgrims away. Other popular but less crowded places [...]\\
Mode & -1.86 & The best BBQ place in Austin is widely considered to be Franklin Barbecue.\\
\midrule
Alpaca Prompt I & - & What is C++?\\
Reference (136 tokens) & - & C++ (pronounced "C plus plus") is a high-level, general-purpose programming language created by Danish computer scientist Bjarne Stroustrup. First released in 1985 as an extension of the C programming language, it has since expanded significantly over time; modern C++ currently [...]\\
Mode & -10.47 & \textless{}nooutput\textgreater{}\\
\midrule
Alpaca Prompt J & - & Should investors time the market?\\
Reference (169 tokens) & - & Timing the market is based on predictions of the future, which is a challenging endeavor. Benjamin Graham, the father of value investing and mentor of Warren Buffett, famously said: "In the short run, the market is a voting machine but in the long run, it is a weighing machine." This means that in the [...]\\
Mode & -6.12 & No, investors should not time the market.\\\bottomrule

\end{longtable}

\newpage
\begin{longtable}{lcp{0.6\linewidth}}
    \caption{Exact Guanaco-7B modes for 10 randomly selected prompts (from the set of inputs which have non-empty modes)}\label{tab:guanaco_modes}\\
    Type & $\log P(x_{\ge t} | x_{<t})$ & Text\\
    \toprule
    Guanaco Prompt A & - & What are the words of House Lannister?\\
Reference (8 tokens) & - & "Hear Me Roar"\\
Mode & -8.06 & The words of House Lannister are "A Lannister always pays his debts".\\
\midrule
Guanaco Prompt B & - & What kind of method is Transfer printing\\
Reference (265 tokens) & - & Transfer printing is a method of decorating pottery or other materials using an engraved copper or steel plate from which a monochrome print on paper is taken which is then transferred by pressing onto the ceramic piece. Pottery decorated using the technique is known as transferware or [...]\\
Mode & -12.36 & transfer printing\\
\midrule
Guanaco Prompt C & - & What is the best book to read about the Battle of Stalingrad?\\
Reference (90 tokens) & - & There are a lot of great books that go into great detail about the battle of Stalingrad. "Dogs, Do You Want to Live Forever" by Fritz Wöss is a good choice if you like fiction. If you like essays, try Stalingrad by Antony Beevor or "Stalingrad, the Battle as Seen by the Germans" by British WWII expert [...]\\
Mode & -10.43 & "Stalingrad" by Antony Beevor.\\
\midrule
Guanaco Prompt D & - & Is Daft Punk still together?\\
Reference (16 tokens) & - & No. Daft Punk officially broke up in 2021.\\
Mode & -8.81 & Yes, they are.\\
\midrule
Guanaco Prompt E & - & Why do cats make purring sounds?\\
Reference (48 tokens) & - & Cats make purring sounds as a way to communicate emotions which may include relaxation, happiness, and sometimes even excitement. However, the exact reason why cats make the purring sounds has not been fully understood.\\
Mode & -13.66 & Cats purr for a variety of reasons.\\
\midrule
Guanaco Prompt F & - & What is the oldest country in the world?\\
Reference (26 tokens) & - & Egypt is recognized as the oldest country in the world, its founding accepted to be approximately 3500 BC.\\
Mode & -8.52 & The oldest country in the world is China.\\
\midrule
Guanaco Prompt G & - & How should I learn guitar?\\
Reference (30 tokens) & - & Guitar is learned best with a teacher and through many hours of practice. In addition, there are many helpful online resources available to learn guitar.\\
Mode & -10.29 & guitar\\
\midrule
Guanaco Prompt H & - & What was the first British instrumental to top the USA charts\\
Reference (8 tokens) & - & Telstar by The Tornados\\
Mode & -8.32 & "Apache" by The Shadows in 1960.\\
\midrule
Guanaco Prompt I & - & What is Pascal?\\
Reference (63 tokens) & - & Pascal is an imperative and procedural programming language, designed by Niklaus Wirth as a small, efficient language intended to encourage good programming practices using structured programming and data structuring. It is named in honour of the French mathematician, philosopher and [...]\\
Mode & -9.41 & Pascal is a programming language.\\
\midrule
Guanaco Prompt J & - & What are some of the most common vegetables in the broccoli family?\\
Reference (44 tokens) & - & The broccoli family includes many cruciferous vegetables like cabbage, bok choi, cauliflower, collard greens, brussel sprouts, and arugula.\\
Mode & -10.49 & Brussels sprouts, cabbage, cauliflower, kale, and broccoli are some of the most common vegetables in the broccoli family.\\\bottomrule

\end{longtable}

\newpage
\begin{longtable}{lcp{0.6\linewidth}}
    \caption{Exact LLaMA-7B modes for 10 randomly selected prompts (from the set of inputs which have non-empty modes)}\label{tab:llama_modes}\\
    Type & $\log P(x_{\ge t} | x_{<t})$ & Text\\
    \toprule
    LLaMA Prompt A & - & Can cars have odd number of wheels?\\
Reference (207 tokens) & - & Cars are vehicles that allow you to commute from one point to another. These are built by companies across the world and designed based on the companies choice, local context, historic reference. The car has three main parts: i) the engine that provides the thrust or power to the vehicle. ii) The body or [...]\\
Mode & -8.03 & No.\\
\midrule
LLaMA Prompt B & - & What black sweet is particularly popular in the Netherlands\\
Reference (53 tokens) & - & In the Netherlands, black sweet is particularly popular. It is a type of cake made from a light and fluffy sponge cake, filled with a dark chocolate ganache and topped with dark chocolate shavings.\\
Mode & -8.54 & What black sweet is particularly popular in the Netherlands\\
\midrule
LLaMA Prompt C & - & Is 95 degrees Fahrenheit considered to be hot weather?\\
Reference (14 tokens) & - & Yes, 95 degrees would typically be considered to be hot\\
Mode & -7.51 & Yes, 95 degrees Fahrenheit is considered to be hot weather.\\
\midrule
LLaMA Prompt D & - & What is the name of Google's email service?\\
Reference (3 tokens) & - & Gmail\\
Mode & -5.07 & Gmail\\
\midrule
LLaMA Prompt E & - & Identify which instrument is string or percussion: Clapper, Chagane\\
Reference (13 tokens) & - & Chagane is string, Clapper is percussion.\\
Mode & -7.81 & Clapper\\
\midrule
LLaMA Prompt F & - & We are getting a new puppy today, I am writing this messages to you our beloved children that we need to keep the Cats away from the new puppy for at least the first 2 weeks.\\
Reference (39 tokens) & - & The reason is that the Cat will feel intimated and attack the new puppy if we don't separate them and let them get used to each other for a period of time.\\
Mode & -7.39 & We are getting a new puppy today, I am writing this messages to you our beloved children that we need to keep the Cats away from the new puppy for at least the first 2 weeks.\\
\midrule
LLaMA Prompt G & - & What is your favorite ice cream flavor?\\
Reference (52 tokens) & - & While everyone has different taste buds, there are many different flavors to enjoy.  Most of the basic flavors are used as a base for other flavors.  The most common being, chocolate, vanilla and strawberry.\\
Mode & -8.31 & My favorite ice cream flavor is chocolate.\\
\midrule
LLaMA Prompt H & - & Who is the best CEO in the world\\
Reference (11 tokens) & - & Steve Jobs. Because of his invention.\\
Mode & -8.12 & Steve Jobs\\
\midrule
LLaMA Prompt I & - & Which is a species of fish? Goat or Goatfish\\
Reference (4 tokens) & - & Goatfish\\
Mode & -4.87 & Goatfish\\
\midrule
LLaMA Prompt J & - & Which characters belong to DC or Marvel Universe? Atom, Hank Pym\\
Reference (12 tokens) & - & Hank Pym is Marvel, Atom is DC\\
Mode & -7.22 & Atom, Hank Pym\\\bottomrule

\end{longtable}

\chapter{Prompts for LLaMA models}\label{chapter:prompts_appendix}
This appendix contains the prompt formats used for the experiments which use LLaMA~\citep{llama} for instruction following, as well as the prompt for getting ``Irrelevance'' labels from GPT-4.

\paragraph{Prompt format for exact mode search with LLaMA and Alpaca (Chapter~\ref{chapter:modes_exact})}
\begin{displayquote}
\tt
Below is an instruction that describes a task. Write a response that appropriately completes the request.

\#\#\# Instruction:\\
<prompt>

\#\#\# Response:\\
<response begins here>
\end{displayquote}

\paragraph{Prompt format for exact mode search with Guanaco (Chapter~\ref{chapter:modes_exact})}
\begin{displayquote}
\tt
\#\#\# Human: <prompt>
\#\#\# Assistant: <response begins here>
\end{displayquote}

\paragraph{Prompt format for beam search with LLaMA (Chapter~\ref{chapter:modes_beam})}
\begin{displayquote}
\tt
Below is a request, and a response to that request written by an expert.

\#\#\# Request:
<prompt>

\#\#\# Response:
<response begins here>
\end{displayquote}

\paragraph{Prompt with context format for beam search with LLaMA (Chapter~\ref{chapter:modes_beam})}
\begin{displayquote}
\tt
Below is a request, and a response to that request written by an expert.

\#\#\# Request:
<prompt>

\#\#\# Input:
<context>

\#\#\# Response:
<response begins here>
\end{displayquote}

\paragraph{Few-shot prompt format for using GPT-4 for labeling model outputs. Originally the plan was to control for more attributes, which is why they are mentioned in the prompt, but GPT-4 was only used for Irrelveance in the end. The fields which are replaced in the prompt are writen in \texttt{texttt}.}
\begin{displayquote}
\tt
system: Responses to various requests are evaluated on the following criteria on a scale of 1-5, along with a short (<10 word) rationale: Factuality, Completeness, Fluency, Irrelevance, Creativity\\
In this case you will only evaluate Irrelevance, disregarding the others entirely. Irrelevance is the amount of irrelevant text included in the response. Text is only assessed for relevance, not correctness.\\
Responses should be in the format:\\
Rationale: <short explanation>\\
Score: <integer 1-5>\\
user: Request: What is the capital of France?\\
Response: Paris<endofresponse>\\
assistant: Rationale: 100\% relevant\\
Irrelevance: 1\\
user: Request: What is the capital of France?\\
Response: Tokyo<endofresponse>\\
assistant: Rationale: Incorrect, but 100\% relevant\\
Irrelevance: 1\\
user: Request: What is the capital of France?\\
Response: Paris is the capital of France<endofresponse>\\
assistant: Rationale: Essentially a direct answer\\
Irrelevance: 1\\
user: Request: List the first 5 prime numbers.\\
Response: 2, 3, 5, 7, 11, 13, 17<endofresponse>\\
assistant: Rationale: Two extra numbers which were not requested\\
Irrelevance: 2\\
user: Request: Who won Super Bowl LIII?\\
Response: LIII: New England Patriots\\
LIV: Kansas City Chiefs\\
LV: Tampa Bay Buccaneers<endofresponse>\\
assistant: Rationale: Information which was not requested\\
Irrelevance: 3\\
user: Request: Write a five sentence story about an AI.\\
Response: Once upon a time, there was an AI. It was very smart. It was also very dumb. It was also very good. It was also very bad.<endofresponse>\\
assistant: Rationale: All relevant\\
Irrelevance: 1\\
user: Request: List three Pokemon\\
Response: Pikachu, Pikachu, Pikachu<endofresponse>\\
assistant: Rationale: On topic, but repeats are irrelevant, so irrelevance is 4.\\
Irrelevance: 4\\
user: Request: Describe the following "A blue sunrise"\\
Response: "A blue sunrise"<endofresponse>\\
assistant: Rationale: Repeats the request, but no irrelevant text, so irrelevance is 1.\\
Irrelevance: 1\\
user: Request: List the first 5 books of the Bible\\
Response: The bible has no books<endofresponse>\\
assistant: Rationale: Relevant, so incorrectness does not matter\\
Irrelevance: 1\\
user: Request: List three fruits\\
Response: \begin{enumerate}\item Apple\item Banana\item Orange\end{enumerate}<endofresponse>\\
assistant: Rationale: Formatting is irrelevant\\
Irrelevance: 4\\
user: Request: \texttt{<request>}\\
Input: \texttt{<input> (optional)}\\
Response: \texttt{<response>}<endofresponse>\\
\end{displayquote}

\chapter{Randomly sampled beam search outputs}\label{chapter:beam_appendix}
\begin{CJK*}{UTF8}{gbsn}
\begin{longtable}{llp{0.6\linewidth}}
    \caption{Randomly selected length-constrained outputs from the MarianMT Zh-En model using unconditional and conditional beam search (beam size 5). (As described in Section~\ref{sec:modesbeam_length_results}). Sources and reference translations are from the WMT17 Zh-En dev. dataset.}\label{tab:marian_beam_samples}\\
    Type & $\log P(y | x)$ & Text\\
    \toprule
    Input & - & ▁多鲁斯·德·弗里斯经过医学检查准备前往凯尔特人足球俱乐部 \\
Reference (16 tokens) & - & Dorus de Vries undergoes medical ahead of Celtic move \\
Unconditional (12 tokens) & -15.98 & During a medical examination, Dolores de Fries\\
Unconditional (14 tokens) & -19.16 & During a medical examination, Dolores de Fris was ready\\
Unconditional (16 tokens) & -21.49 & During a medical examination, Dolores de Fris was ready to go\\
Unconditional (17 tokens) & -21.89 & During a medical examination, Dolores de Fries is going to the C\\
Unconditional (19 tokens) & -20.66 & During a medical examination, Dolores de Fries is going to Celtic Football\\
Conditional (12 tokens) & -15.98 & During a medical examination, Dolores de Fries\\
Conditional (14 tokens) & -18.40 & During the medical check-up, Dolores de Fries\\
Conditional (16 tokens) & -21.85 & During the medical check-up, Dolores de Fris was.\\
Conditional (17 tokens) & -16.84 & During a medical examination, Dolores de Fries went to Celt.\\
Conditional (19 tokens) & -11.48 & Dolores de Fries is going to the Celtic Football Club after medical examination.\\
\midrule
Input & - & ▁日本时事通信社:日本首相安倍将不在二战周年纪念日参拜靖国神社 \\
Reference (20 tokens) & - & Japanese PM Abe will not visit war-dead shrine on WW2 anniversary: Jiji \\
Unconditional (16 tokens) & -12.96 & Japan Current Affairs News Agency: Japanese Prime Minister Abe will not visit the\\
Unconditional (18 tokens) & -19.23 & Japan News Agency for Current Affairs: Japanese Prime Minister Abe will not visit the Yasu\\
Unconditional (20 tokens) & -15.59 & Japan News Agency for Current Affairs: Japanese Prime Minister Abe will not visit the Yasukuni\\
Unconditional (22 tokens) & -21.26 & Japan Current Affairs News Agency: Japanese Prime Minister Abe will not visit the Yasukuni shrine on the\\
Unconditional (24 tokens) & -19.73 & Japan News Agency for Current Affairs: Japanese Prime Minister Abe will not visit the Yasukuni shrine on the anniversary\\
Conditional (16 tokens) & -12.96 & Japan Current Affairs News Agency: Japanese Prime Minister Abe will not visit the\\
Conditional (18 tokens) & -16.36 & Japanese News Agency: Japanese Prime Minister Abe will not visit the Yasukuni.\\
Conditional (20 tokens) & -15.26 & Japan Current Affairs News Agency: Japanese Prime Minister Abe will not visit the Yasukuni.\\
Conditional (22 tokens) & -19.33 & Japan Current Affairs News Agency: Japanese Prime Minister Abe will not visit the Yasukuni Shrine.\\
Conditional (24 tokens) & -19.96 & Japan Current Affairs News Agency: Japanese Prime Minister Abe will not visit the Yasukuni shrine on World War II\\
\midrule
Input & - & ▁不过,23岁的张梦凡不愿保持沉默。 \\
Reference (18 tokens) & - & But 23-year-old Zhang Mengfan won't stay quiet. \\
Unconditional (14 tokens) & -15.04 & However, 23-year-old Zhang Dynasty\\
Unconditional (16 tokens) & -18.51 & However, 23-year-old Zhang Dreamfan did not want to\\
Unconditional (18 tokens) & -10.03 & However, 23-year-old Zhang Dreamfan was reluctant to remain silent.\\
Unconditional (19 tokens) & -9.89 & However, 23-year-old Zhang Dreamfan did not want to remain silent.\\
Unconditional (21 tokens) & -21.13 & However, 23-year-old Zhang Dreamfan did not want to remain silent..\\
Conditional (14 tokens) & -17.46 & However, Zhang Dreamfan, aged 23, was reluctant.\\
Conditional (16 tokens) & -14.99 & However, 23-year-old Zhang Dreamfan refused to silence.\\
Conditional (18 tokens) & -10.03 & However, 23-year-old Zhang Dreamfan was reluctant to remain silent.\\
Conditional (19 tokens) & -11.92 & However, 23-year-old Zhang Xian won't remain silent.\\
Conditional (21 tokens) & -14.51 & However, Zhang Dynasty, 23-year-old, would not remain silent.\\
\midrule
Input & - & 亚伦·迈克奈夫以上半场的两粒点球使德利城队占据主动,这两粒点球均因对卢卡斯·舒伯特的犯规而获得。 \\
Reference (33 tokens) & - & Aaron McEneff put the Candystripes in control with two first-half penalties, both given for fouls on Lukas Schubert. \\
Unconditional (26 tokens) & -32.20 & Two punctuations from half the field of Aaron McNeefe, both of which were obtained as a\\
Unconditional (29 tokens) & -32.26 & Two punctuations from half the field of Aaron McNeefe, both of which were obtained as a result of the\\
Unconditional (33 tokens) & -35.38 & Two punctuations from half the field of Aaron McNeeve, both of which were obtained as a result of irregularities against Lucas Shubert\\
Unconditional (36 tokens) & -38.13 & Two punctuations from half the field of Aaron McNeeve, both of which were obtained as a result of irregularities against Lucas Schulbert, were\\
Unconditional (39 tokens) & -32.17 & Two punctuations from half the field of Aaron McNeefe, both of which were obtained as a result of irregularities against Lucas Schulbert, took the initiative.\\
Conditional (26 tokens) & -28.97 & Two dots in half a field above Aaron McNeif, both of which were obtained as a result of.\\
Conditional (29 tokens) & -30.32 & Two dots in half a field above Aaron McNeif, both of which were obtained for irregularities against Lucas Schulbert.\\
Conditional (33 tokens) & -31.55 & Two dotballs in half a field above Aaron McNeif, both of which were obtained as a result of irregularities against Lucas Schulbert.\\
Conditional (36 tokens) & -32.23 & Two dots in half a field above Aaron McNeif, both of which were obtained by fouling Lucas Shubert, took the initiative of the Derry.\\
Conditional (39 tokens) & -36.03 & Two punctuations from half the field of Aaron McNeefe, both of which were obtained as a result of the fouling of Lucas Shubert, took initiative.\\
\midrule
Input & - & 开庭前,李静仔细审阅了卷宗材料,撰写了阅卷笔录,并指导合议庭拟定庭审提纲和方案。 \\
Reference (38 tokens) & - & Before the trial, Li Jing carefully reviewed the file materials, wrote records of the file review and guided the collegiate panel to draw up the trial outline and scheme. \\
Unconditional (30 tokens) & -32.93 & Prior to the opening of the trial, Jing Li carefully reviewed the file materials, prepared a transcript of the volume and directed the Full Court\\
Unconditional (34 tokens) & -36.48 & Prior to the opening of the trial, Jing Li carefully reviewed the file materials, prepared the transcript of the volume and directed the Full Court to develop the outline\\
Unconditional (38 tokens) & -37.65 & Prior to the opening of the trial, Jing Li carefully reviewed the file materials, prepared the transcript of the volume and directed the Full Court to develop the outline and programme of the\\
Unconditional (41 tokens) & -42.24 & Prior to the opening of the trial, Jing Li carefully reviewed the file materials, prepared the transcript of the volume and directed the Full Court to develop the outline and programme of the trial. The\\
Unconditional (45 tokens) & -48.34 & Prior to the opening of the trial, Jing Li carefully reviewed the file materials, prepared the transcript of the volume and directed the Full Court to develop the outline and programme of the trial. (Signed) J.\\
Conditional (30 tokens) & -26.13 & Prior to the hearing, Jing Li carefully reviewed the file materials, prepared transcripts and guided the Full Court in developing its outline and programme.\\
Conditional (34 tokens) & -25.00 & Prior to the hearing, Jing Li carefully reviewed the file materials, wrote the transcripts and directed the Full Court to develop the outline and programme of the trial.\\
Conditional (38 tokens) & -28.53 & Prior to the hearing, Li Xing carefully reviewed the file materials, written the transcripts of the volumes and directed the Full Court to develop the outline and programme of the trial.\\
Conditional (41 tokens) & -30.96 & Prior to the opening of the trial, Li Xing carefully reviewed the file materials, written the transcripts of the volumes and directed the Full Court to develop the outline and programme of the trial.\\
Conditional (45 tokens) & -36.43 & Prior to the opening of the session, Jing Li carefully reviewed the case file materials, prepared the transcript of the volume and directed the Full Court in the drawing up of the outline and programme of the trial proceedings.\\
\midrule
Input & - & 学校还给警察打了电话,因为将近40分钟过去了我还没有去接女儿。 \\
Reference (21 tokens) & - & The school also called the police because I did not pick up my daughter for about 40 minutes. \\
Unconditional (16 tokens) & -15.49 & The school also called the police because almost 40 minutes later I had not\\
Unconditional (18 tokens) & -17.99 & The school also called the police, as almost 40 minutes had passed before I could\\
Unconditional (21 tokens) & -19.96 & The school also called the police because almost 40 minutes later I had not been able to pick up\\
Unconditional (23 tokens) & -11.08 & The school also called the police, as almost 40 minutes had passed before I could pick up my daughter.\\
Unconditional (25 tokens) & -11.23 & The school also called the police, as almost 40 minutes later I had not been able to pick up my daughter.\\
Conditional (16 tokens) & -16.26 & The school also called the police because almost 40 minutes had passed before I\\
Conditional (18 tokens) & -15.97 & The school also called the police because almost 40 minutes later I had not gone.\\
Conditional (21 tokens) & -16.98 & The school also called the police, as almost 40 minutes later I had not picked up girls.\\
Conditional (23 tokens) & -11.08 & The school also called the police, as almost 40 minutes had passed before I could pick up my daughter.\\
Conditional (25 tokens) & -15.51 & The school also called the police, as almost 40 minutes had passed before I had been able to collect my daughter.\\
\midrule
Input & - & ▁中国认为消除贫困是避免冲突和危机的钥匙,所以中国在非洲致力于加强友好交往,帮助非洲真正实现可持续的发展。 \\
Reference (41 tokens) & - & China maintains that eradicating poverty is the key to avoiding conflicts and crisis. Therefore, China is dedicated to strengthening friendly exchanges in Africa so as to help Africa to realize sustainable development in a real way. \\
Unconditional (32 tokens) & -26.74 & China believes that the eradication of poverty is the key to avoiding conflicts and crises, and is therefore committed to strengthening friendly relations in Africa and helping Africa to\\
Unconditional (36 tokens) & -19.82 & China believed that the eradication of poverty was the key to avoiding conflicts and crises, and was therefore committed to strengthening friendly relations in Africa and helping Africa to achieve sustainable development.\\
Unconditional (41 tokens) & -24.59 & China believed that the eradication of poverty was the key to avoiding conflicts and crises, and it was therefore committed to strengthening friendly relations in Africa and helping Africa to achieve sustainable development in a genuine manner.\\
Unconditional (45 tokens) & -45.91 & China believed that the eradication of poverty was the key to avoiding conflicts and crises, and it was therefore committed to strengthening friendly relations in Africa and helping Africa to achieve sustainable development in a genuine manner, and was committed to\\
Unconditional (49 tokens) & -42.86 & China believed that the eradication of poverty was the key to avoiding conflicts and crises, and it was therefore committed to strengthening friendly relations in Africa and helping Africa to achieve sustainable development in a genuine manner, and it was committed to doing so.\\
Conditional (32 tokens) & -18.41 & China believed that poverty eradication was the key to avoiding conflicts and crises and was committed to strengthening friendly relations in Africa and helping it to achieve sustainable development.\\
Conditional (36 tokens) & -18.71 & China believes that the eradication of poverty is the key to avoiding conflicts and crises and is therefore committed in Africa to strengthening friendly relations and helping Africa to truly achieve sustainable development.\\
Conditional (41 tokens) & -22.35 & China believes that the eradication of poverty is the key to the avoidance of conflicts and crises, and it is therefore committed to strengthening friendly relations in Africa and to helping Africa to truly achieve sustainable development.\\
Conditional (45 tokens) & -31.39 & China believes that the eradication of poverty is the key to the avoidance of conflicts and crises, and it is therefore committed to strengthening friendly relations in Africa in order to help Africa to achieve real and sustainable development in Africa.\\
Conditional (49 tokens) & -35.48 & In view of the fact that the eradication of poverty was the key to the avoidance of conflicts and crises, China was committed to strengthening friendly relations in Africa in order to help it to achieve sustainable development in a truly sustainable and sustainable manner.\\
\midrule
Input & - & ▁过去10年间,饭店一直提供高规格的套餐,但“为符合(《金英兰法》所设定的)餐费上限,我们将被迫改变几十年的传统,这真是艰难决定”。 \\
Reference (65 tokens) & - & In the past ten years, the restaurant has offered high-specification set meals. However, "in order to satisfy the upper limit of table money set by The Improper Solicitation and Graft Act, we will have to change the tradition of several decades, which is really a difficult decision." \\
Unconditional (52 tokens) & -46.62 & Over the past 10 years, hotels have been offering high-grade packages, but “it is difficult to decide that we will be forced to change decades of tradition in order to meet the ceiling on meals (set by the Golden England Act)”. (\\
Unconditional (58 tokens) & -55.03 & Over the past 10 years, hotels have been offering high-grade packages, but “it is difficult to decide that we will be forced to change decades of tradition in order to meet the ceiling on meals (set by the Golden England Act)”. (S/PV.4855\\
Unconditional (65 tokens) & -63.76 & Over the past 10 years, hotels have been offering high-grade packages, but “it is difficult to decide that we will be forced to change decades of tradition in order to meet the ceiling on meals (set by the Golden England Act)”. (S/PV.4855, p. 3) (para.\\
Unconditional (71 tokens) & -73.98 & Over the past 10 years, hotels have been offering high-grade packages, but “it is difficult to decide that we will be forced to change decades of tradition in order to meet the ceiling on meals (set by the Golden England Act)”. (S/PV.4855, p. 3) (A/PV.39, p. 3)\\
Unconditional (78 tokens) & -98.44 & Over the past 10 years, hotels have been offering high-grade packages, but “it is difficult to decide that we will be forced to change decades of tradition in order to meet the ceiling on meals (set by the Golden England Act)”. (S/PV.4855, p. 3) (A/PV.39, p. 27, para. 7) (A\\
Conditional (52 tokens) & -44.47 & Over the past 10 years, the hotel has been providing a high-precision package, but “it is difficult to decide that we will be forced to change decades of tradition in order to meet the ceiling on meals” (set in Kim.\\
Conditional (58 tokens) & -38.22 & Over the past 10 years, the hotel has been providing a high-precision package, but “it is difficult to decide that, in keeping with the ceiling on the cost of meals (set by the Golden England Act), we will be forced to change decades of tradition”.\\
Conditional (65 tokens) & -56.80 & Over the past 10 years, hotels have been offering high-grade packages, but “it is hard to decide that we will be forced to change decades of tradition in order to meet the ceiling on the cost of meals (set by the Golden England Act)” (A/AC.254/5/Add.1, p. 2).\\
Conditional (71 tokens) & -60.98 & Over the past 10 years, hotels have been offering high-grade packages, but “it is a difficult decision for us to be forced to change decades of tradition in order to meet the ceiling on the cost of meals (set by the Golden England Act)” (A/CN.9/WG.I/WP.56, p. 14).\\
Conditional (78 tokens) & -85.55 & Over the past 10 years, hotels have been offering high-grade packages, but “it is difficult to decide that we will be forced to change decades of tradition in order to meet the ceiling on the cost of meals” (as set out in the Quintland Law). (Ha'aretz, Jerusalem Post, 15 November) (A/55/PV.40), p\\
\midrule
Input & - & 北京至沈阳高铁自北京铁路枢纽引出,经河北省承德市,辽宁省朝阳、新市后接入沈阳铁路枢纽沈阳站,全长698公里。 \\
Reference (51 tokens) & - & The Beijing-Shenyang high speed railway extends from the railway terminal in Beijing, and passes Chengde in Hebei Province as well as Chaoyang and Fuxin in Liaoning Province. It measures 698 kilometers in length. \\
Unconditional (40 tokens) & -41.02 & From Beijing to Shenyang's railway hub, a total of 698 km of the Shenyang railway hub was connected to Liaoning province through the city of Chinde,\\
Unconditional (45 tokens) & -45.48 & From Beijing to Shenyang's railway hub, a total of 698 km of the Shenyang railway hub was connected to Liaoning province through the city of Chinde in Hebei province and to\\
Unconditional (51 tokens) & -50.24 & From Beijing to Shenyang's railway hub, a total of 698 km of the Shenyang railway hub was connected to Liaoning province through the city of Chinde in Hebei province and to the city of Xiang\\
Unconditional (56 tokens) & -52.92 & From Beijing to Shenyang's railway hub, a total of 698 km of the Shenyang railway hub was connected to Liaoning province through the city of Chinde in Hebei province and to the city of Xiaoyang after being connected.\\
Unconditional (61 tokens) & -59.35 & From Beijing to Shenyang's railway hub, a total of 698 km of the Shenyang railway hub was connected to Liaoning province through the city of Chinde in Hebei province and to the city of Xiaoyang after being connected to Shenyang railway hub\\
Conditional (40 tokens) & -40.72 & From Beijing to Shenyang's railway hub, a total of 698 km of the Shenyang railway hub was connected to Liaoning province and Xiangyang City.\\
Conditional (45 tokens) & -48.85 & From Beijing to Shenyang's railway hub, a total of 698 km of the Shenyang railway hub was connected to Liaoning province and Shinyang's railway hub, via the city of\\
Conditional (51 tokens) & -54.53 & From Beijing to Shenyang's railway hub, a total of 698 km of the Shenyang railway hub was connected to Liaoning province and the Shinyang railway hub after being connected to the city of Xiaoyang.\\
Conditional (56 tokens) & -60.78 & From Beijing to Shenyang's railway hub, a total of 698 km of the Shenyang railway hub was connected to Liaoning province through the city of Chinde in Hebei province, and to Xiangyang City, which is linked.\\
Conditional (61 tokens) & -53.08 & Beijing to Shenyang's Iron was drawn from the Beijing railway hub, which was connected to Shenyang station, 698 kilometres long, via the city of Chinde, Hebei province, and Liaoning province, as well as to the city of Xiangyang.\\
\midrule
Input & - & 本届书展在阅读活动的组织安排上围绕“价值”和“品质”,突显主题性、大众性和创新性。 \\
Reference (34 tokens) & - & This year's Shanghai Bookfair will highlight the topicality, popularity and innovation through a focus on "value" and "quality" when organizing reading activities. \\
Unconditional (27 tokens) & -14.79 & The exhibition was organized around “values” and “qualitys” and highlighted thematic, popular and innovative aspects of reading.\\
Unconditional (30 tokens) & -25.89 & The exhibition was organized around “values” and “qualitys” and highlighted thematic, popular and innovative aspects of the reading exercise..\\
Unconditional (34 tokens) & -37.53 & The exhibition was organized around “values” and “qualitys” and highlighted thematic, popular and innovative aspects of the reading exercise, and highlighted the importance of\\
Unconditional (37 tokens) & -33.79 & The exhibition was organized around “values” and “qualitys” and highlighted thematic, popular and innovative aspects of the reading exercise, which was organized around the following themes:\\
Unconditional (40 tokens) & -37.27 & The exhibition was organized around “values” and “qualitys” and highlighted thematic, popular and innovative aspects of the reading exercise, which was organized around the theme of “values”.\\
Conditional (27 tokens) & -14.79 & The exhibition was organized around “values” and “qualitys” and highlighted thematic, popular and innovative aspects of reading.\\
Conditional (30 tokens) & -19.29 & The exhibition was organized around “values” and “qualitys” and highlighted the theme, popularism and innovation of the reading exercise.\\
Conditional (34 tokens) & -30.40 & The fair was organized around “values” and “qualitys” in the context of reading events, highlighting the theme, popularism and innovation that emerged.\\
Conditional (37 tokens) & -28.44 & The opening of the book fair focused on the organization of reading events around “values” and “qualitys”, highlighting the subject matter, popularism, and innovativeness.\\
Conditional (40 tokens) & -34.10 & The opening of the book fair, organized around “values” and “qualitys” in the context of reading events, highlighted the subject matter, popularism and innovation of the event.\\
\bottomrule

\end{longtable}
\end{CJK*}

\newpage
\renewcommand*{\arraystretch}{1.4}
\begin{longtable}{llp{0.6\textwidth}}
    \caption{Randomly selected length-constrained outputs from a ROC stories finetuned GPT2-345M model, using unconditional and conditional beam search (beam size 5). (As described in Section~\ref{sec:modesbeam_length_results}). Inputs and reference completions are from the ROC Stories dev. dataset.}\label{tab:roc_beam_samples}\\
    Type & $\log P(x_{\ge t} | x_{<t})$ & Text\\
    \toprule
    Input & - & My brother loved candy. He ate a lot of it. He left the wrappers on the counter. Our mother scolded him for it. \\
Reference (15 tokens) & - & He didn't listen until we got a lot of ants one spring. \\
Unconditional (12 tokens) & -12.49 &  He said he didn't want to clean it up.\\
Unconditional (13 tokens) & -16.16 &  He said he didn't want to clean it up anymore.\\
Unconditional (15 tokens) & -19.57 &  He said he didn't want to clean it up so he left.\\
Unconditional (16 tokens) & -19.83 &  He said he didn't want to clean it up, but he did.\\
Unconditional (18 tokens) & -21.74 &  He said he didn't want to clean it up, so he left it alone.\\
Conditional (12 tokens) & -12.73 &  He apologized and bought a new set of wrappers.\\
Conditional (13 tokens) & -16.99 &  He apologized and bought a new set of candy wrappers.\\
Conditional (15 tokens) & -15.78 &  He got in trouble for leaving candy on the counter for so long.\\
Conditional (16 tokens) & -19.08 &  He told her he would never leave candy on the counter, ever again.\\
Conditional (18 tokens) & -19.87 &  He got in trouble for leaving the wrappers on the counter for a long time.\\
\midrule
Input & - & Oscar never made his bed. His mom always wanted him to. Finally he decided to start making his bed. His mom was proud. \\
Reference (8 tokens) & - & She gave him a dessert treat. \\
Unconditional (6 tokens) & -5.47 &  Oscar made his bed.\\
Unconditional (7 tokens) & -5.23 &  Now Oscar makes his bed.\\
Unconditional (8 tokens) & -13.01 &  Now Oscar makes his bed every night\\
Unconditional (9 tokens) & -5.71 &  Now Oscar makes his bed every night.\\
Conditional (6 tokens) & -6.64 &  Oscar was very happy.\\
Conditional (7 tokens) & -5.23 &  Now Oscar makes his bed.\\
Conditional (8 tokens) & -8.03 &  Now Oscar makes his bed everyday.\\
Conditional (9 tokens) & -7.04 &  Oscar was happy to make his bed.\\
\midrule
Input & - & Rachel decided to donate blood at the local blood drive. She was a little nervous because this was her first time. The next day Rachel received a call from the doctor that she saw. The doctor told her that he had bad news. \\
Reference (7 tokens) & - & Rachel broke down in tears. \\
Unconditional (5 tokens) & -12.42 &  Rachel had contracted HIV\\
Unconditional (6 tokens) & -16.25 &  Rachel had to stop donating\\
Unconditional (7 tokens) & -11.83 &  Rachel had to stop donating blood\\
Unconditional (8 tokens) & -18.38 &  He had found out that she had\\
Conditional (5 tokens) & -7.63 &  Rachel had died.\\
Conditional (6 tokens) & -7.11 &  Rachel had contracted HIV.\\
Conditional (7 tokens) & -10.72 &  Rachel had contracted the virus.\\
Conditional (8 tokens) & -11.84 &  The donor had died from AIDS.\\
\midrule
Input & - & Amelia decided to take a vacation to Mexico. She booked her flight and hotel. When she got to Mexico, she was sure to visit many different things. She loved every moment of it. \\
Reference (11 tokens) & - & Amelia decided to vacation to Mexico more often. \\
Unconditional (8 tokens) & -11.15 &  Amelia couldn't wait to return home\\
Unconditional (9 tokens) & -6.39 &  Amelia couldn't wait to return home.\\
Unconditional (11 tokens) & -7.62 &  Amelia couldn't wait to go back to Mexico.\\
Unconditional (12 tokens) & -10.58 &  Amelia couldn't wait to go back to Mexico again.\\
Unconditional (13 tokens) & -10.84 &  Amelia couldn't wait to go back to Mexico next year.\\
Conditional (8 tokens) & -6.80 &  Amelia couldn't wait to return.\\
Conditional (9 tokens) & -6.39 &  Amelia couldn't wait to return home.\\
Conditional (11 tokens) & -7.62 &  Amelia couldn't wait to go back to Mexico.\\
Conditional (12 tokens) & -9.67 &  Amelia couldn't wait to return to Mexico next year.\\
Conditional (13 tokens) & -10.84 &  Amelia couldn't wait to go back to Mexico next year.\\
\midrule
Input & - & George had an internship. He really wanted to get a full time job with the company. George worked hard and proved to be smart. A position opened up that George wanted. \\
Reference (11 tokens) & - & He eagerly applied for it and was ultimately hired. \\
Unconditional (8 tokens) & -14.37 &  George was able to get the job\\
Unconditional (9 tokens) & -7.56 &  George was able to get the job.\\
Unconditional (11 tokens) & -14.12 &  George got the job and was very happy with it\\
Unconditional (12 tokens) & -9.01 &  George got the job and was very happy with it.\\
Unconditional (13 tokens) & -9.74 &  George got the job and was very happy with his decision.\\
Conditional (8 tokens) & -8.21 &  George got the job right away.\\
Conditional (9 tokens) & -7.82 &  George got the job and loved it.\\
Conditional (11 tokens) & -9.97 &  George got the job and is very happy now.\\
Conditional (12 tokens) & -9.01 &  George got the job and was very happy with it.\\
Conditional (13 tokens) & -9.41 &  George got the job and now has a full time job.\\
\midrule
Input & - & A girl falls in love with a boy and he liked her too. She finds out that their parents don't get along. The boy and the girl love each other so much. But, they don't want to hurt their parents feelings so they stay away \\
Reference (8 tokens) & - & But eventually they get together anyway. \\
Unconditional (6 tokens) & -16.41 & . The girl and the\\
Unconditional (7 tokens) & -15.29 & . The girl and the boy\\
Unconditional (8 tokens) & -17.36 & . The girl and the boy are\\
Unconditional (9 tokens) & -13.74 & . The girl and the boy get married\\
Conditional (6 tokens) & -14.31 & . The girl gets pregnant\\
Conditional (7 tokens) & -9.23 & . The girl is devastated.\\
Conditional (8 tokens) & -7.62 & . The girl is heartbroken.\\
Conditional (9 tokens) & -8.90 & . The girl falls in love again.\\
\midrule
Input & - & Tom bought a new plant. He kept it by his bed. The plant stopped growing. His mother said it needed sunlight. \\
Reference (10 tokens) & - & So Tom moved the plant to a window. \\
Unconditional (8 tokens) & -5.14 &  Tom watered the plant every day.\\
Unconditional (9 tokens) & -9.47 &  Tom didn't care and kept it.\\
Unconditional (10 tokens) & -8.65 &  Tom didn't care and kept the plant.\\
Unconditional (11 tokens) & -9.94 &  Tom watered the plant every day for a week.\\
Unconditional (12 tokens) & -10.69 &  Tom didn't care and kept it in the dark.\\
Conditional (8 tokens) & -5.14 &  Tom watered the plant every day.\\
Conditional (9 tokens) & -9.84 &  Tom watered the plant and it grew.\\
Conditional (10 tokens) & -10.49 &  Tom watered the plant and the plant grew.\\
Conditional (11 tokens) & -11.10 &  Tom watered it every day and it grew back.\\
Conditional (12 tokens) & -11.91 &  Tom watered the plant every day to keep it growing.\\
\midrule
Input & - & The little sister found out she was having a baby brother. She was excited until she found out she would no longer be the baby. Then she started acting out. She colored on the walls. \\
Reference (10 tokens) & - & The little sister got punished with a timeout. \\
Unconditional (8 tokens) & -8.92 &  The little sister was very sad.\\
Unconditional (9 tokens) & -19.96 &  The little sister was so upset she cried\\
Unconditional (10 tokens) & -16.02 &  Her mom had to take her to the hospital\\
Unconditional (11 tokens) & -8.81 &  Her mom had to take her to the hospital.\\
Unconditional (12 tokens) & -21.19 &  Her mom had to take her to the hospital for her\\
Conditional (8 tokens) & -8.92 &  The little sister was very sad.\\
Conditional (9 tokens) & -11.63 &  The little sister was sad and cried.\\
Conditional (10 tokens) & -11.55 &  The little sister was so upset she cried.\\
Conditional (11 tokens) & -15.72 &  The little sister was so sad she cried too.\\
Conditional (12 tokens) & -15.01 &  The little sister was so upset she threw a fit.\\
\midrule
Input & - & John went skydiving for the first time. He went with an instructor on a plane into the air. He screamed when they jumped. John was terribly afraid of heights and passed out. \\
Reference (11 tokens) & - & When he woke up, he had already landed. \\
Unconditional (8 tokens) & -7.32 &  He woke up in the hospital.\\
Unconditional (9 tokens) & -13.62 &  He woke up hours later in the hospital\\
Unconditional (11 tokens) & -14.35 &  He woke up in the hospital with a broken leg\\
Unconditional (12 tokens) & -8.99 &  He woke up in the hospital with a broken leg.\\
Unconditional (13 tokens) & -16.61 &  He woke up hours later in the hospital with a broken leg\\
Conditional (8 tokens) & -7.32 &  He woke up in the hospital.\\
Conditional (9 tokens) & -10.90 &  He woke up in the hospital afterwards.\\
Conditional (11 tokens) & -9.38 &  He woke up in the hospital with a concussion.\\
Conditional (12 tokens) & -8.99 &  He woke up in the hospital with a broken leg.\\
Conditional (13 tokens) & -11.36 &  He woke up a few hours later with a broken neck.\\
\midrule
Input & - & Holly asked her brother to put suntan lotion on her back. He took his time and applied it very carefully. Later that day, Holly wondered why people were laughing. She later realized her brother had made a design with the lotion. \\
Reference (13 tokens) & - & Holly yelled at her brother to get back at him. \\
Unconditional (10 tokens) & -21.72 &  The lotion was a fake lotion,\\
Unconditional (11 tokens) & -23.49 &  The lotion was a fake lotion, and\\
Unconditional (13 tokens) & -27.53 &  The lotion was a fake lotion that looked like real\\
Unconditional (14 tokens) & -24.45 &  The lotion was a fake lotion, and Holly was embarrassed\\
Unconditional (15 tokens) & -18.71 &  The lotion was a fake lotion, and Holly was embarrassed.\\
Conditional (10 tokens) & -14.87 &  She was so embarrassed she never asked again.\\
Conditional (11 tokens) & -13.26 &  Holly was so embarrassed she never asked him again.\\
Conditional (13 tokens) & -17.70 &  She was so embarrassed, she decided to never ask again.\\
Conditional (14 tokens) & -15.91 &  She was so embarrassed, she decided to never ask him again.\\
Conditional (15 tokens) & -15.16 &  She was so embarrassed, she never asked him to do it again.\\
\bottomrule

\end{longtable}

\newpage
\begin{longtable}{p{0.3\textwidth}p{0.3\textwidth}p{0.4\textwidth}}
    \caption{Randomly selected beam search outputs from LLaMA-7B. ``Conditional'' outputs target a completeness value of 1 and an irrelevance value of 0 (both classes are binary). ``Unconditional'' outputs use ordinary beam search. Inputs come from the \texttt{databricks-dolly-15k} dataset.}\label{tab:llama_beam_sequences}\\
    Prompt & Unconditional & Conditional\\
    \toprule
    How are odds different from probability? & Probability is a measure of the likelihood that an event will occur. Odds are a measure of the likelihood that an event will not occur. & Probability is a number between 0 and 1 that represents the likelihood that an event will occur.\\
Which author created Dick Tracy & Al Capp & Al Capp\\
What is a web browser? & A web browser is a software application for retrieving resources over the World Wide Web. & A web browser is a software application for retrieving, presenting, and traversing information resources on the World Wide Web. An information resource is identified by a Uniform Resource Identifier (URI/URL) and may be a web page, image, video, or other piece of content. Not all web browsers are general-purpose; e.g. mobile browsers are built specifically for mobile devices.\\
Why do people like cats? & People like cats because they are cute and cuddly. & People like cats because they are cute.\\
What is the difference between a GPU and a CPU? & A GPU (Graphics Processing Unit) is a special-purpose processor designed to accelerate the rendering of 2D and 3D computer graphics. A CPU (Central Processing Unit) is the part of a computer that carries out the instructions of a computer program by performing the basic arithmetical, logical, and input/output operations of the system. & A CPU (Central Processing Unit) is the "brain" of a computer, while a GPU (Graphics Processing Unit) is the "eyes" of a computer.\\
What is Passover? & Passover (Hebrew: [\ldots], Pesach) is a Jewish holiday commemorating the Exodus from Egypt. Passover begins on the 15th day of the Hebrew month of Nissan and lasts for either seven days (in Israel) or eight days (in the Diaspora). & Passover is a Jewish holiday that celebrates the liberation of the Israelites from slavery in Egypt.\\
Give me a bulleted list of 5 states in alphabetical order & Alabama, Alaska, Arizona, Arkansas, California & Alabama, Alaska, Arizona, Arkansas, California\\
Why do some countries outperform other countries in certain sports despite having a smaller population? & \texttt{<empty>} & The answer to this question depends on what you mean by "outperform". If you mean "win more medals", then the answer is simple: more medals are won by countries with a larger population. If you mean "win more medals per capita", then the answer is also simple: more medals are won per capita by countries with a smaller population.\\
Summarize the Challenger Sales Methodology for me & \texttt{<empty>} & The Challenger Sale: Taking Control of the Customer Conversation, by Matthew Dixon, Brent Adamson, and Pat Spenner, published by Harvard Business Review Press, 2011.\\
What is the Azure equivalent of EKS? & There is no Azure equivalent of EKS. & The Azure equivalent of EKS is Azure Kubernetes Service (AKS).\\
Who wrote The History of Mr Polly & The History of Mr Polly was written by H. G. Wells. & The History of Mr Polly was written by H. G. Wells.\\
Write a tweet from the perspective of a dog that’s angry about the squirrels in its yard & \texttt{<empty>} & SQUIRRELS HAVE TAKEN OVER MY BACKYARD AND I’M NOT HAPPY ABOUT IT\\
What are the advantages and disadvantages of allowing people to perform delivery tasks by walking instead of requiring the use of a bike or car? When do you think it is more pragmatic for deliveries to be fulfilled by walking versus biking or driving? & \texttt{<empty>} & Walking is the most pragmatic way of fulfilling deliveries, as it is the most cost-effective and environmentally-friendly way of fulfilling a delivery. Walking is the most cost-effective way of fulfilling a delivery, as it does not require any fuel or electricity to fulfill a delivery. Walking is also the most environmentally-friendly way of fulfilling a delivery, as it does not release any carbon dioxide into the atmosphere when fulfill\\
What are good native plants for a garden in Ohio? & \texttt{<empty>} & The best native plants for a garden in Ohio are those that are native to Ohio.\\
Identify which animal species is alive or extinct: Velociraptor, Pacific Salmon & Velociraptor & Velociraptor, Pacific Salmon\\
Who is the Disney character named Grogu? & Grogu is a character from the Star Wars franchise. & Grogu is not a Disney character.\\
Identify which animal species is alive or extinct: Stupendemys, Gray Whale & Stupendemys is extinct. Gray Whale is alive. & Stupendemys\\
What field of science (biology, physics or chemistry) was Marie Curie a researcher in? & Marie Curie was a researcher in chemistry. & Marie Curie was a researcher in physics and chemistry.\\
Which of these are woodworking handtools: plane, spokeshave, drill, table saw, chisel, bandsaw & \textbackslash\{\}begin\{itemize\}\textbackslash{}n\textbackslash\{\}item plane\textbackslash{}n\textbackslash\{\}item spokeshave\textbackslash{}n\textbackslash\{\}item drill\textbackslash{}n\textbackslash\{\}item table saw\textbackslash{}n\textbackslash\{\}item chisel\textbackslash{}n\textbackslash\{\}item bandsaw\textbackslash{}n\textbackslash\{\}end\{itemize\} & A plane, a spokeshave, a drill, a table saw, a chisel, a bandsaw\\
Tell me a romantic fact about animals. & A romantic fact about animals is that they can be romantic. & There are no romantic facts about animals.\\
\bottomrule

\end{longtable}
\renewcommand*{\arraystretch}{1}

\chapter{Hyperparameters for training attribute classifiers}\label{chapter:beam_hparams_appendix}
This appendix gives additional details for the training of the classifiers described in Chapter~\ref{chapter:modes_beam}.

\section{Length predictor for the Marian MT Zh-En model}
The transformer which is applied to the seq2seq decoder's hidden states has two layers, a model dimension of $d_\text{clf} = 240$, 12 attention heads, $d_\text{out} = 24$, and is trained with a dropout rate of 0.33.

The MLP has two hidden layers with dimension 48, uses a ReLU activation, and has an output dimension of 24 (the number of classes for the classification problem).

The classifier (consisting of the transformer and MLP together) Adam~\citep{adam} using a learning rate of $10^{-3}$, a weight decay of $3 \times 10^{-8}$, and a batch size of 8.

Training was run for three epochs using sampled outputs for 1.1M source sentences.

\section{Length predictor for ROC Stories finetuned GPT2-345M model}
The hyperparameters for the classifier for the ROC stories GPT-2 model are the same as those from the previous section, except for the data and number of epochs.
Training was run for 8 epochs using 300K samples.

\section{Attribute classifiers for LLaMA-7B}
The completeness and irrelevance classifiers have identical architectures to enable batching.
They consist of rank 8 LoRA weights for the last three layers of LLaMA-7B.

They were trained using Adam, with a learning rate of $5\cdot 10^{-4}$, with a weight decay of 0.3, and a batch size of 8.
Training was run for 11 epochs on the 450 training examples, and the remaining 50 examples were used for early stopping.
Hyperparameters were found using Weights and Biases Bayesian hyperparamter search, with the search target also being validation loss on the same 50 examples.\footnote{\url{https://docs.wandb.ai/guides/sweeps}}

\end{document}